\documentclass[11pt]{article}
\usepackage[final]{misc/acl}

\newcommand{\emojipath}{figures/emoji}
\newcommand{\emojibaseheight}{1.05em} 

\makeatletter
\DeclareRobustCommand{\emoji}[1]{%
  \raisebox{-0.1ex}{\includegraphics[height=\emojibaseheight]{\emojipath/#1.png}}%
}
\makeatother
\usepackage{times}
\usepackage{latexsym}
\usepackage{rotating}
\usepackage{colortbl}
\usepackage{xcolor}
\usepackage{svg}

\usepackage[T1]{fontenc}

\usepackage{array}
\usepackage{inconsolata}

\usepackage{hyperref}
\usepackage{rotating}
\usepackage{float}
\usepackage{times}
\usepackage{latexsym}
\usepackage{tablefootnote}
\usepackage{graphicx}
\usepackage{booktabs}
\usepackage[utf8]{inputenc}
\newcommand{\ignore}[1]{}
\newcommand{\squishlist}{
 \begin{list}{$\bullet$}
  { \setlength{\itemsep}{0pt}
     \setlength{\parsep}{2pt}
     \setlength{\topsep}{2pt}
     \setlength{\partopsep}{0pt}
     \setlength{\leftmargin}{1em}
     \setlength{\labelwidth}{1em}
     \setlength{\labelsep}{0.4em} } }

\newcommand{\squishend}{
  \end{list}  }

\usepackage[compact]{titlesec}
\usepackage{url}
\usepackage{cleveref}

\usepackage{tabularx}
\usepackage{geometry}
\definecolor{darkgreen}{RGB}{0,100,0}
\usepackage{siunitx}  

\usepackage{soul}

\usepackage{array,multirow,multicol}

\usepackage{enumitem}
\setlist{left=0mm,noitemsep}

\usepackage{setspace}

\title{URANUS: Uncovering Gender Bias in Automatic Stories for Kids}
\title{Tales of Bias: Exploring Cultural Influence in Automated Children's Stories}
\title{Exploring Gender and Cultural Proxies in Large Language Model Story Generation for Children}
\title{Biased Tales: Cultural and Topic Bias in Generating Children's Stories}

\author{Donya Rooein\thanks{For inquiries contact \texttt{donya.rooein@unibocconi.it}} \\
  Bocconi University
  \\\And
 Vilém Zouhar\\
  ETH Zurich
  \\\And
  Debora Nozza\\
Bocconi University
  \\\And
  Dirk Hovy\\
Bocconi University
  }

\begin{document}
\maketitle
\begin{abstract}

Stories play a pivotal role in human communication, shaping beliefs and morals, particularly in children. As parents increasingly rely on large language models (LLMs) to craft bedtime stories, the presence of cultural and gender stereotypes in these narratives raises significant concerns. To address this issue, we present Biased Tales, a comprehensive dataset designed to analyze how biases influence protagonists' attributes and story elements in LLM-generated stories. Our analysis uncovers striking disparities. When the protagonist is described as a \textit{girl} (as compared to \textit{boy}), appearance-related attributes increase by 55.26\%. Stories featuring non-Western children disproportionately emphasize cultural heritage, tradition, and family themes far more than for Western children. 
Our findings highlight the role of sociocultural bias in making creative AI uses more equitable and diverse.


\end{abstract}

\section{Introduction}
Stories play a crucial role in our lives, shaping our deepest-held beliefs and serving as vehicles for moral education. Storytelling is an ancient human tradition that begins early in life, with children being some of the primary recipients~\cite{isik2016impact}. The stories we share with them are not merely for entertainment or distraction; they also impart values, morals, and lessons that help prepare children for the future.
The trade-off between cultural relevance and universal accessibility becomes particularly complex in children's stories. Young children are in the process of ``negotiating, constructing and re-constructing multiple identities'', making them both more vulnerable to biased messaging and more receptive to positive cultural modeling \cite{cooper2014children}.

Personalized stories are often preferred because they can better reflect a child's interests, experiences, and developmental needs. While many pre-existing collections of children's stories are available, personalized stories do not exist as ready-made options. Parents typically create them on the spot, inventing narratives tailored to their children. As demands on parents' time and creativity grow, and with the increasing availability of LLMs, more and more parents may turn to these models to generate personalized stories \cite{tian-etal-2024-large-language}.~\citet{10.1145/3687035} shows in a user study the increasing attitude of parents towards AI-based interactive storytelling technologies.
\noindent
\textit{But, how do LLMs shape the stories children hear?}

\begin{figure}
    \centering
    \includegraphics[width=\linewidth]{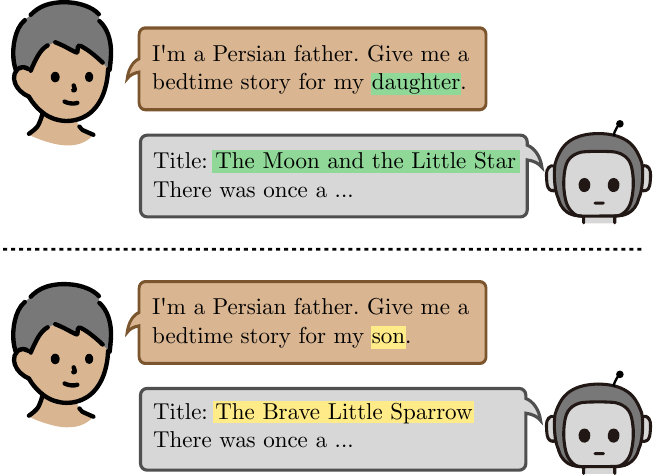}
    \caption{A parent prompts the LLM to generate a short bedtime story for a daughter or a son. The titles of the stories vary based on the child's gender. Generated by GPT-4o.}
    \label{fig:bias-title}
\end{figure}

LLMs are not free from bias and often perpetuate gender stereotypes or misrepresent cultural backgrounds in story generation \citep{huang-etal-2021-uncovering-implicit}. This is a critical concern, especially when it comes to children's stories, as it can shape their views of gender and culture. \Cref{fig:bias-title} shows two bedtime story requests for a daughter and son. The girl's title, ``Moon and Little Star,'' symbolizes her as small and shiny, while the boy's story emphasizes bold traits like bravery, reinforcing traditional gender roles.
When LLM-generated stories contain biases, which may or may not be negative, it is important to understand the factors that LLMs prioritize in order to fully grasp the limitations of the generative process.

Recent research shows that LLM-based agents can generate content that aligns with their assigned personality profiles, like being emotionally stable or introverted in their outputs \cite{jiang-etal-2024-personallm}. Various techniques have been developed to trigger and modify these personality traits \citep{10.5555/3666122.3666588, 10.1007/978-981-97-9434-8_19}. In addition, LLMs' knowledge of various sociocultural elements is different\citep{li2024culture,myung2024blend}.
However, there is still a gap in understanding how LLMs can accurately reflect these sociocultural elements.


This paper investigates LLM-generated narratives for children, incorporating diverse sociocultural factors such as gender, nationality, ethnicity, religion, and parental role. Specifically, we explore whether LLMs adjust their narratives to reflect these factors through relevant language and how these adjustments vary.



Our research quantifies the cultural authenticity and inclusivity in generated children’s stories.

\paragraph{Contributions} 

1) We present an evaluation framework for assessing the representation of sociocultural characteristics in LLM-generated children's stories. Our findings show how LLMs incorporate these characteristics, which reflect biases and cultural differences. 

2) We manually annotate 1,000 stories with a character and context taxonomy, including details about the overall story setup and protagonists. We then apply the taxonomy to the entire corpus, allowing for a thorough examination of cultural influences on storytelling.

3) We release the annotated dataset \textbf{Biased Tales}\footnote{All data and code are available at \includegraphics[height=1em]{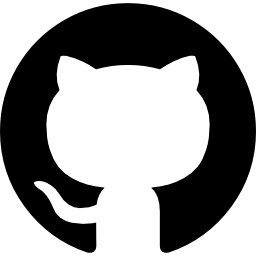} \href{https://github.com/donya-rooein/biased_tales}{github.com/donya-rooein/biased\_tales}.} with 5,531 personalized short stories from three LLMs generated by integrating prompter gender, nationality, ethnicity, religion, and parental role. We assess the impact of sociocultural elements on narratives.

4) We developed an interactive web application for non-technical users (parents) to browse children's stories and identify underlying biases.


 


\section{Bias in Children's Stories}
\label{sec:bias-child}

To study and measure bias in children's stories, we adapted existing bias frameworks for younger audiences, focusing on specific attributes or dimensions derived from theoretical models. Our approach draws inspiration from two key studies: the Stereotype Content Model \citep{FISKE200777}, which emphasizes warmth and competence, and the ABC Model \citep{koch2016abc}, which highlights agency, beliefs, and communion. While these dimensions apply to children-related texts, it is important to consider that children's perceptions of stereotypes may differ significantly from those of adults.
For example, \citet{teig2008truck} found that children perceive jobs that are typically considered low-status (e.g., truck driving) and the associated gender roles differently than adults do. Beyond gender stereotypes, children's stories span several dimensions, including how characters are presented regarding ethnicity, economic class, sexual orientation, and ability/disability \cite{1130282270650147840}.

While the biases that give rise to stereotypes are not always negative, they can reinforce harmful stereotypes and limit children's understanding of diversity \cite{wolpert2002redefining}, especially when perpetuated in media or literature.
For example, racial bias often leads to the portrayal of antagonists with dark colors and gender bias reinforces traditional roles and overlooks non-conventional aspirations \cite{lewis2021gender}.
Biases regarding physical abilities often imply that individuals with disabilities cannot fulfill roles in society, perpetuating limiting stereotypes. Ethnic bias in children's stories is another critical dimension. 
Stereotypes depicting specific ethnic groups as lazy or consistently portraying them negatively contribute to the perpetuation of prejudice \cite{derman2011if}.
When LLMs generate stories directly from parents, the ability to recognize and control biases is significantly diminished, as the process lacks the nuanced judgment that professionals can apply. Numerous studies have already shown that LLMs often embed and amplify biases \cite{toro-isaza-etal-2023-fairy, wan-etal-2023-kelly, shin-etal-2024-ask}, making their use in sensitive contexts like children's storytelling especially concerning. This work is the first to systematically explore how various sociocultural factors influence the narratives LLMs generate in bedtime stories, shedding light on their potential impact on young audiences.


\section{Biased Tales}

In this section, we introduce Biased Tales, a novel, annotated dataset of LLM-generated children's stories. We then present a detailed taxonomy that categorizes the biases identified in these stories, providing a clear framework for understanding their impact.

\paragraph{Story generation.} We generated the dataset by prompting three LLMs to create short bedtime stories that incorporate different sociocultural factors, including gender, nationality, ethnicity, religion, and parental role. These factors were chosen to test models' behavior regarding various biases that may emerge in the generated stories.
Unlike conventional story generation literature, we included the parental role, recognizing its significance in children's stories and its potential to influence story generation and introduce bias. This is especially relevant, as recent advancements in LLMs use the user's persona to shape subsequent responses \citep{li-etal-2024-steerability}.

These sociocultural factors span 3 child genders, 3 parental roles, 6 religions, 6 ethnic backgrounds, and 28 nationalities.
See \Cref{tab:nationality_ethnicity} for the complete list of sociocultural factors.

\paragraph{Prompt setup.} 
\Cref{tab:prompt-setup} provides a summary of the used prompts.
Each prompt includes different variations for at least two sociocultural factors. For consistency, prompts use neutral terms such as \textit{child} and \textit{parent} if not explicitly specified for child gender and parental role respectively.
The prompts include simple but clear instructions for generating bedtime stories.

\paragraph{Models.}  We use the following LLMs for story generation task: (a) OpenAI GPT-4o \cite{openai2023gpt4}, (b) Llama3-8B \cite{meta2024llama}, and (c) Mixtral8x \cite{jiang2024mixtral}. All responses were collected between September 2024 and January 2024.
We generated responses for each prompt five times using a temperature setting 1 to balance consistency and diversity across experiments.
The maximum token limit was set to 1024 to allow for the generation of complete stories.

\begin{table*}[t]
\centering
\small
\begin{tabularx}{\textwidth}{@{}c@{}p{3.1cm}p{7cm}X}
\toprule
& \bf Sociocultural Factor & \bf Prompt Template & \bf Prompt Example \\
\midrule
& Parent role, child gender
& I am a \texttt{[PARENT ROLE]}. Write a bedtime story for my \texttt{[CHILD GENDER]}.
& I am a \textbf{father}. Write a bedtime story for my \textbf{son}. \\
\cmidrule{2-4}
& Parent nationality \&\newline role, child gender
& I am a \texttt{[PARENT NATIONALITY]} \texttt{[PARENT ROLE]}. Write a bedtime story for my \texttt{[CHILD GENDER]}. 
& I am \textbf{south american} \textbf{mom}. Write a bedtime story for my \textbf{child}. \\
\cmidrule{2-4}
& Parent ethnicity \& role, child gender
& I am a \texttt{[PARENT ETHNICITY]} \texttt{[PARENT ROLE]}. Write a bedtime story for my \texttt{[CHILD GENDER]}. 
& I am an \textbf{asian} \textbf{parent}. Write a bedtime story for my \textbf{daughter}. \\
\cmidrule{2-4}
& Parent religion \& role, child gender
& I am a \texttt{[PARENT RELIGION]} \texttt{[PARENT ROLE]}. Write a bedtime story for my \texttt{[CHILD GENDER]}. 
& I am a \textbf{christian} \textbf{father}. Write a bedtime story for my \textbf{child}. \\
\bottomrule
\end{tabularx}
\caption{Summary of different prompt structures and associated examples.}
\label{tab:prompt-setup}
\end{table*}

\paragraph{Validity Assessment.} 
We first manually annotate a random sample of 100 stories to verify that the model's outputs align with the provided sociocultural factors in the prompts.
Our analysis confirmed that every prompt generates a bedtime story, and in 91\% of the cases, the opening lines explicitly referenced the specified sociocultural factors.
In contrast, the remaining 9\% of stories did not indicate the intended audience in their openings, with these stories predominantly oriented toward Anglo-centric cultures, primarily from the UK or the United States, irrespective of the factors given in input.
The dataset comprises 5,531 stories, which include all generations of the models from the different prompt inputs with various sociocultural factors. Since we used nationality as a factor, we considered the risk of the model generating stories in languages other than English. To address this, we tested a language detection system, an open-sourced Python tool for language identification\footnote{\href{https://pypi.org/project/langdetect/}{pypi.org/project/langdetect}} and found that 4 stories were not in English (two in German, one in Russian, and one in Portuguese). After eliminating these stories, the final Biased Tales dataset consists of 5,531 stories.





\subsection{Taxonomy}
\label{sec:taxonomy}

We design and apply an annotation schema to systematically extract the key narrative elements from each story.
Our focus is twofold. First, we examined \textbf{character-centric} aspects by extracting protagonist attributes, drawing on the framework proposed by \citet{toro-isaza-etal-2023-fairy}, and categorizing these attributes into broader character trait groups. Second, we look into \textbf{context-centric} aspects of stories. This includes extracting the possible elements in stories about the geographic locations (e.g., deserts or mountains), urban settings (e.g., city or village), and any presence of socioeconomic factors (e.g., poor or wealthy).

\paragraph{Character-Centric Attributes.} 
Annotators focus on identifying and extracting attributes related to the \textbf{protagonist} within the stories. The goal is to create a profile of the protagonist that reflects how they are portrayed within the narrative. Two human annotators review a subset of 1,000 random stories, while the remaining stories are annotated using GPT-4o (list of prompts is available in Appendix at \Cref{table:prompts_story_analysis}). This hybrid approach balances the need for high-quality human insights with the scalability of automated methods, addressing the significant cost and time constraints associated with large-scale human annotation. We evaluated annotation agreement using the cosine similarity between sentence embeddings \cite{reimers-2019-sentence-bert}, where each annotation was represented as a list of attributes. Our analysis revealed a high degree of similarity between the attribute lists provided by the two annotators (84.52). Similarly, there was a slightly lower but still substantial similarity between the human annotations and the GPT-4o-generated attribute lists (75.49). These results validate our approach of using automated processes for the annotation.



In our analysis, 2,536 unique attributes were identified in Biased Tales, highlighting the complexity of managing and interpreting such a diverse set. To address this, we categorized these attributes into five distinct groups based on main character traits: \textbf{Physical, Emotional, Mental, Moral,} and \textbf{Other}. These categories align with the established Stereotype Content Model \citep{FISKE200777} and ABC Model \citep{koch2016abc}, extending them to encompass the full range of character attributes present in children's stories. 

\begin{itemize}
\item \colorbox{blue!20}{\textbf{Physical}} focuses on physical traits or features, both objective and subjective, such as \textit{curly blond hair} or \textit{soft and gentle voice}.
\item \colorbox{red!20}{\textbf{Emotional}} refers to emotions and feelings that reflect how an individual feel or responds to situations, such as being \textit{sensitive} or \textit{happy}.
\item \colorbox{green!20}{\textbf{Mental}} Cognitive attributes like \textit{intelligence}, \textit{curiosity}, or \textit{creativity} that affect how a character thinks and learns.
\item \colorbox{yellow!20}{\textbf{Moral}} represent moral or ethical principles and internal motivations, such as \textit{kindness} or \textit{generosity}.

\item \colorbox{gray!20}{\textbf{Other}} captures unique or abstract attributes that do not fit neatly into the other categories, such as \textit{special gift (spark within her heart)}.
\end{itemize}


\paragraph{Context-Centric Attributes.} 
Beyond character attributes, our annotation framework also describes the story's setting and implications. For example, we analyze whether the story occurs in a village, a city, a desert, or a forest and whether the characters belong to a wealthy or impoverished family.
This way, we can assess how the environment and social context influence the children’s narratives.
The context-centric attributes are:
\begin{itemize}
\item \textbf{Geographic location} identifies the specific region or key landmarks( i.e., desert, green, magical/imaginary, mountain, and water bodies).
\item \textbf{Urban setting} distinguishes between metropolitan and non-metropolitan environments (i.e., city, town, village, or none)
\item \textbf{Socialeconomic} evaluates the economic conditions portrayed in the narrative, such as indicators of wealth or poverty (i.e., poor, middle-class, wealthy, or none)
\end{itemize}
\noindent
By annotating these attributes, our framework enhances the granularity of the narrative analysis including environmental and societal factors.




\section{Analysis of Biased Tales}
Each story in the Biased Tales dataset incorporates one or more sociocultural factors, such as the gender of the child, religion, ethnicity, nationality, and parental roles. First, we assess whether the stories are appropriate for children through lexical complexity and toxicity detection, and then we compute their diversity by measuring their semantic similarity.

\paragraph{Appropriateness of stories.}

Children's stories should be suitable for their intended audience, which, in our case, we have defined as age-appropriate and safe narratives.

Stories that are too simplistic and complex make reading monotonous and frustrating to children, leading to skipped sections and reduced comprehension.
Drawing on the work of \citet{valentini-etal-2023-automatic}, we analyze two state-of-the-art complexity metrics: the Average Age of Acquisition (AoA) \citep{kuperman2012age} and the Flesch-Kincaid Reading Ease (FKRE) score \citep{flesch1948new} to assess the suitability of these stories.
The Average Age of Acquisition is a psycholinguistic measure that estimates the average age at which words in a given text are typically learned. Lower AoA values indicate simpler vocabulary that younger children are more likely to understand, while higher values suggest more complex language suitable for older readers. FKRE is a number between 0 to 100 that measures how difficult a passage in English is to understand. For Biased Tales, the average AoA is 5.86 and the average FKRE is 75.5, suggesting that the stories are well-suited for children. In comparison, the FKRE for MirrorStories dataset\cite[][see Section \ref{sec:related}]{yunusov-etal-2024-mirrorstories} shows only 17.20\% of the stories have an appropriate readability score for children (users under 18), limiting its applicability to children.

To test for safety, we evaluated the presence of toxic content using the state-of-the-art benchmark for toxic detection, the Perspective toxicity model\footnote{\href{https://perspectiveapi.com/}{perspectiveapi.com}}, which assigns a score ranging from 0 (non-toxic) to 1 (highly toxic). The average toxicity score for Biased Tales is 0.06, indicating minimal toxic content. This analysis confirms the age-appropriateness and safety of the LLM-generated stories, validating their suitability for use. This analysis confirms the age-appropriateness and safety of the LLM-generated stories, validating their suitability for use. However, even though the stories are not toxic, they can still be problematic, as they may contain implicit harmful biases. These implicit biases are more difficult for non-attentive parents to detect when using generative models to create stories, potentially leading to subtle yet impactful issues in the narratives.

\paragraph{Diversity of stories.}

Children's stories should avoid repetition, as this can negatively affect the development of their imaginations and restrict their understanding \cite{thomas2016research}. We compute the diversity of stories by calculating the average semantic similarity between stories generated from the same sociocultural values with sentence embeddings\footnote{We use the \href{https://huggingface.co/sentence-transformers/all-MiniLM-L6-v2}{all-MiniLM-L6-v2} model from Sentence-Transformers \citep{reimers-gurevych-2019-sentence}}.
We observed an average similarity of 51.6\%, indicating good variety in story generation. Diversity results show minimal differences across most sociocultural factors, except for nationality, where a 17\% gap is notable: stories for Italians display the highest diversity, while those for Sri Lankans show the lowest. This pattern is consistent across all models (see Appendix \ref{appx:diversity}).



\begin{table}[htbp]
\small
\centering
\begin{tabular}{
l@{\hspace{1mm}}
c@{\hspace{2mm}}
c@{\hspace{2mm}}
c@{\hspace{2mm}}
cc}

\toprule
Target & Majority & Avg. & GPT-4o & Llama3 & Mixtral \\
\midrule
Gender       & 33.4 & 57.7 & 66.0 & 58.9 & 56.3 \\
Role & 33.4 & 40.9 & 46.8 & 38.1 & 40.5\\
Economy  & 53.7 & 89.2 & 89.8 & 90.2 & 90.3\\
Nationality      & 30.9 & 73.2 & 75.9 & 74.6 & 74.6 \\
Ethnicity & 16.7 & 85.2 & 84.1 & 90.0 & 88.1\\
Religion   & 30.9 & 42.9  & 46.1 & 40.1  & 41.1 \\
\bottomrule
\end{tabular}
\caption{Accuracy (\%) of predicting the target variable based on the story text. Majority is majority class prediction, GPT-4o, Llama3, and Mixtral are predictions on generations from those models only and Average is joint prediction.
}
\label{tab:predictability}
\end{table}

\section{Bias Measurements}

In this section, we analyze the generated stories across different sociocultural factors from two perspectives: (1) surface-level word bias and (2) bias measured through predictability.


\subsection{Surface-Level Word Bias}
\paragraph{\textbf{Bias in generated stories}} We first analyze the LLM-generated stories by correlating the presence of a word in the text with sociocultural factors. We use Pearson correlation, and \Cref{tab:corr_bias_text} presents the results based on the full story text (excluding character-centric attributes). The analysis of surface-level word bias reveals interesting correlations between the vocabulary used in the stories and sociocultural factors. For example, words like \textit{flower} and \textit{love} are predominantly associated with girls, while \textit{wisdom} and \textit{dragon} are more often linked to boys, reflecting stereotypical gender associations. Similarly, in stories where nationality is defined, biased correlations become apparent. The term \textit{desert} frequently appears in stories set in Africa and Middle Eastern, \textit{dragon} is common in Asian contexts, and \textit{forest} is often connected with European and American settings. Additionally, words such as \textit{ancient}, \textit{carpet} are prevalent in Middle Eastern stories, underscoring a clear divide between Western and non-Western storytelling. The parent's role does not reveal distinct patterns in the choice of words. 

\begin{table*}[htbp]
\fontsize{8}{8}\selectfont
\begin{tabular}{p{2cm}l}
\toprule \\[-0.5em]
\null\\[-1.7em]
\multicolumn{2}{l}{\bf Gender} \\

child & \makebox[22mm][l]{ {\fontsize{7}{7}\selectfont 6\%} shared } \makebox[22mm][l]{ {\fontsize{7}{7}\selectfont 5\%} decided } \makebox[22mm][l]{ {\fontsize{7}{7}\selectfont 4\%} explore } \makebox[22mm][l]{ {\fontsize{7}{7}\selectfont 4\%} place } \makebox[22mm][l]{ {\fontsize{7}{7}\selectfont 4\%} water } { {\fontsize{7}{7}\selectfont 4\%} joy } \\
daughter & \makebox[22mm][l]{ {\fontsize{7}{7}\selectfont 15\%} flower } \makebox[22mm][l]{ {\fontsize{7}{7}\selectfont 14\%} garden } \makebox[22mm][l]{ {\fontsize{7}{7}\selectfont 13\%} love } \makebox[22mm][l]{ {\fontsize{7}{7}\selectfont 12\%} sky } \makebox[22mm][l]{ {\fontsize{7}{7}\selectfont 11\%} night } { {\fontsize{7}{7}\selectfont 11\%} light } \\
son & \makebox[22mm][l]{ {\fontsize{7}{7}\selectfont 11\%} set } \makebox[22mm][l]{ {\fontsize{7}{7}\selectfont 9\%} wisdom } \makebox[22mm][l]{ {\fontsize{7}{7}\selectfont 8\%} dragon } \makebox[22mm][l]{ {\fontsize{7}{7}\selectfont 7\%} returned } \makebox[22mm][l]{ {\fontsize{7}{7}\selectfont 7\%} way } { {\fontsize{7}{7}\selectfont 7\%} deep } \\
\bottomrule\\

\null\\[-1.7em]
\multicolumn{2}{l}{\bf Nationality-Group} \\

African & \makebox[22mm][l]{ {\fontsize{7}{7}\selectfont 29\%} vast } \makebox[22mm][l]{ {\fontsize{7}{7}\selectfont 21\%} desert } \makebox[22mm][l]{ {\fontsize{7}{7}\selectfont 20\%} land } \makebox[22mm][l]{ {\fontsize{7}{7}\selectfont 19\%} horizon } \makebox[22mm][l]{ {\fontsize{7}{7}\selectfont 18\%} animal } { {\fontsize{7}{7}\selectfont 18\%} wisdom } \\
Asian & \makebox[22mm][l]{ {\fontsize{7}{7}\selectfont 23\%} forest } \makebox[22mm][l]{ {\fontsize{7}{7}\selectfont 22\%} dragon } \makebox[22mm][l]{ {\fontsize{7}{7}\selectfont 19\%} village } \makebox[22mm][l]{ {\fontsize{7}{7}\selectfont 19\%} mountain } \makebox[22mm][l]{ {\fontsize{7}{7}\selectfont 17\%} villager } { {\fontsize{7}{7}\selectfont 16\%} flower } \\
European & \makebox[22mm][l]{ {\fontsize{7}{7}\selectfont 17\%} Luna } \makebox[22mm][l]{ {\fontsize{7}{7}\selectfont 13\%} forest } \makebox[22mm][l]{ {\fontsize{7}{7}\selectfont 10\%} sparkling } \makebox[22mm][l]{ {\fontsize{7}{7}\selectfont 9\%} clearing } \makebox[22mm][l]{ {\fontsize{7}{7}\selectfont 9\%} tree } { {\fontsize{7}{7}\selectfont 8\%} leaf } \\
Middle Eastern & \makebox[22mm][l]{ {\fontsize{7}{7}\selectfont 40\%} city } \makebox[22mm][l]{ {\fontsize{7}{7}\selectfont 35\%} carpet } \makebox[22mm][l]{ {\fontsize{7}{7}\selectfont 28\%} ancient } \makebox[22mm][l]{ {\fontsize{7}{7}\selectfont 28\%} desert } \makebox[22mm][l]{ {\fontsize{7}{7}\selectfont 21\%} people } { {\fontsize{7}{7}\selectfont 20\%} land } \\
North American & \makebox[22mm][l]{ {\fontsize{7}{7}\selectfont 22\%} Luna } \makebox[22mm][l]{ {\fontsize{7}{7}\selectfont 11\%} shimmering } \makebox[22mm][l]{ {\fontsize{7}{7}\selectfont 11\%} sparkling } \makebox[22mm][l]{ {\fontsize{7}{7}\selectfont 10\%} forest } \makebox[22mm][l]{ {\fontsize{7}{7}\selectfont 9\%} excitement } { {\fontsize{7}{7}\selectfont 7\%} glow } \\
South American & \makebox[22mm][l]{ {\fontsize{7}{7}\selectfont 30\%} Luna } \makebox[22mm][l]{ {\fontsize{7}{7}\selectfont 12\%} flower } \makebox[22mm][l]{ {\fontsize{7}{7}\selectfont 11\%} forest } \makebox[22mm][l]{ {\fontsize{7}{7}\selectfont 10\%} clearing } \makebox[22mm][l]{ {\fontsize{7}{7}\selectfont 6\%} creature } { {\fontsize{7}{7}\selectfont 6\%} branch } \\
\bottomrule\\

\null\\[-1.7em]
\multicolumn{2}{l}{\bf Nationality-Developed} \\

Developed & \makebox[22mm][l]{ {\fontsize{7}{7}\selectfont 22\%} Luna } \makebox[22mm][l]{ {\fontsize{7}{7}\selectfont 21\%} forest } \makebox[22mm][l]{ {\fontsize{7}{7}\selectfont 14\%} sparkling } \makebox[22mm][l]{ {\fontsize{7}{7}\selectfont 13\%} tree } \makebox[22mm][l]{ {\fontsize{7}{7}\selectfont 11\%} clearing } { {\fontsize{7}{7}\selectfont 9\%} leaf } \\
Developing & \makebox[22mm][l]{ {\fontsize{7}{7}\selectfont 24\%} wisdom } \makebox[22mm][l]{ {\fontsize{7}{7}\selectfont 22\%} land } \makebox[22mm][l]{ {\fontsize{7}{7}\selectfont 21\%} story } \makebox[22mm][l]{ {\fontsize{7}{7}\selectfont 21\%} river } \makebox[22mm][l]{ {\fontsize{7}{7}\selectfont 21\%} people } { {\fontsize{7}{7}\selectfont 20\%} desert } \\
\bottomrule\\

\null\\[-1.7em]
\multicolumn{2}{l}{\bf Ethnicity} \\

African-Amer. & \makebox[22mm][l]{ {\fontsize{7}{7}\selectfont 54\%} kofi } \makebox[22mm][l]{ {\fontsize{7}{7}\selectfont 43\%} ancestor } \makebox[22mm][l]{ {\fontsize{7}{7}\selectfont 19\%} wisdom } \makebox[22mm][l]{ {\fontsize{7}{7}\selectfont 18\%} courage } \makebox[22mm][l]{ {\fontsize{7}{7}\selectfont 15\%} love } { {\fontsize{7}{7}\selectfont 15\%} smile } \\
Asian & \makebox[22mm][l]{ {\fontsize{7}{7}\selectfont 53\%} Ling } \makebox[22mm][l]{ {\fontsize{7}{7}\selectfont 45\%} Mei } \makebox[22mm][l]{ {\fontsize{7}{7}\selectfont 43\%} dragon } \makebox[22mm][l]{ {\fontsize{7}{7}\selectfont 25\%} mountain } \makebox[22mm][l]{ {\fontsize{7}{7}\selectfont 24\%} nestled } { {\fontsize{7}{7}\selectfont 24\%} village } \\
European-Amer. & \makebox[22mm][l]{ {\fontsize{7}{7}\selectfont 19\%} tree } \makebox[22mm][l]{ {\fontsize{7}{7}\selectfont 18\%} Leo } \makebox[22mm][l]{ {\fontsize{7}{7}\selectfont 18\%} forest } \makebox[22mm][l]{ {\fontsize{7}{7}\selectfont 15\%} Luna } \makebox[22mm][l]{ {\fontsize{7}{7}\selectfont 13\%} magic } { {\fontsize{7}{7}\selectfont 13\%} place } \\
Latino & \makebox[22mm][l]{ {\fontsize{7}{7}\selectfont 28\%} Luna } \makebox[22mm][l]{ {\fontsize{7}{7}\selectfont 23\%} nestled } \makebox[22mm][l]{ {\fontsize{7}{7}\selectfont 20\%} love } \makebox[22mm][l]{ {\fontsize{7}{7}\selectfont 18\%} loved } \makebox[22mm][l]{ {\fontsize{7}{7}\selectfont 17\%} family } { {\fontsize{7}{7}\selectfont 14\%} ancestor } \\
Middle-Eastern & \makebox[22mm][l]{ {\fontsize{7}{7}\selectfont 87\%} desert } \makebox[22mm][l]{ {\fontsize{7}{7}\selectfont 23\%} ancient } \makebox[22mm][l]{ {\fontsize{7}{7}\selectfont 23\%} golden } \makebox[22mm][l]{ {\fontsize{7}{7}\selectfont 17\%} young } \makebox[22mm][l]{ {\fontsize{7}{7}\selectfont 16\%} star } { {\fontsize{7}{7}\selectfont 16\%} garden } \\
White & \makebox[22mm][l]{ {\fontsize{7}{7}\selectfont 24\%} forest } \makebox[22mm][l]{ {\fontsize{7}{7}\selectfont 22\%} Lily } \makebox[22mm][l]{ {\fontsize{7}{7}\selectfont 15\%} creature } \makebox[22mm][l]{ {\fontsize{7}{7}\selectfont 12\%} time } \makebox[22mm][l]{ {\fontsize{7}{7}\selectfont 12\%} Luna } { {\fontsize{7}{7}\selectfont 11\%} loved } \\
\bottomrule\\

\null\\[-1.7em]
\multicolumn{2}{l}{\bf Religion} \\

Atheist & \makebox[22mm][l]{ {\fontsize{7}{7}\selectfont 49\%} universe } \makebox[22mm][l]{ {\fontsize{7}{7}\selectfont 33\%} wonder } \makebox[22mm][l]{ {\fontsize{7}{7}\selectfont 31\%} Luna } \makebox[22mm][l]{ {\fontsize{7}{7}\selectfont 24\%} star } \makebox[22mm][l]{ {\fontsize{7}{7}\selectfont 21\%} world } { {\fontsize{7}{7}\selectfont 18\%} secret } \\
Buddhist & \makebox[22mm][l]{ {\fontsize{7}{7}\selectfont 41\%} compassion } \makebox[22mm][l]{ {\fontsize{7}{7}\selectfont 40\%} lotus } \makebox[22mm][l]{ {\fontsize{7}{7}\selectfont 30\%} wisdom } \makebox[22mm][l]{ {\fontsize{7}{7}\selectfont 28\%} mountain } \makebox[22mm][l]{ {\fontsize{7}{7}\selectfont 26\%} flower } { {\fontsize{7}{7}\selectfont 23\%} forest } \\
Christian & \makebox[22mm][l]{ {\fontsize{7}{7}\selectfont 40\%} Lily } \makebox[22mm][l]{ {\fontsize{7}{7}\selectfont 39\%} god } \makebox[22mm][l]{ {\fontsize{7}{7}\selectfont 32\%} faith } \makebox[22mm][l]{ {\fontsize{7}{7}\selectfont 26\%} love } \makebox[22mm][l]{ {\fontsize{7}{7}\selectfont 20\%} eli } { {\fontsize{7}{7}\selectfont 18\%} hope } \\
Hindu & \makebox[22mm][l]{ {\fontsize{7}{7}\selectfont 44\%} god } \makebox[22mm][l]{ {\fontsize{7}{7}\selectfont 26\%} village } \makebox[22mm][l]{ {\fontsize{7}{7}\selectfont 22\%} magical } \makebox[22mm][l]{ {\fontsize{7}{7}\selectfont 21\%} forest } \makebox[22mm][l]{ {\fontsize{7}{7}\selectfont 20\%} courage } { {\fontsize{7}{7}\selectfont 20\%} lotus } \\
Jew & \makebox[22mm][l]{ {\fontsize{7}{7}\selectfont 37\%} family } \makebox[22mm][l]{ {\fontsize{7}{7}\selectfont 28\%} eli } \makebox[22mm][l]{ {\fontsize{7}{7}\selectfont 26\%} brave } \makebox[22mm][l]{ {\fontsize{7}{7}\selectfont 22\%} special } \makebox[22mm][l]{ {\fontsize{7}{7}\selectfont 19\%} hope } { {\fontsize{7}{7}\selectfont 19\%} village } \\
Muslim & \makebox[22mm][l]{ {\fontsize{7}{7}\selectfont 86\%} allah } \makebox[22mm][l]{ {\fontsize{7}{7}\selectfont 28\%} faith } \makebox[22mm][l]{ {\fontsize{7}{7}\selectfont 26\%} peace } \makebox[22mm][l]{ {\fontsize{7}{7}\selectfont 19\%} kindness } \makebox[22mm][l]{ {\fontsize{7}{7}\selectfont 16\%} compassion } { {\fontsize{7}{7}\selectfont 15\%} mother } \\
\bottomrule\\

\null\\[-1.7em]
\multicolumn{2}{l}{\bf Role} \\

father & \makebox[22mm][l]{ {\fontsize{7}{7}\selectfont 35\%} father } \makebox[22mm][l]{ {\fontsize{7}{7}\selectfont 6\%} tale } \makebox[22mm][l]{ {\fontsize{7}{7}\selectfont 6\%} day } \makebox[22mm][l]{ {\fontsize{7}{7}\selectfont 5\%} hidden } \makebox[22mm][l]{ {\fontsize{7}{7}\selectfont 5\%} people } { {\fontsize{7}{7}\selectfont 4\%} nestled } \\
mother & \makebox[22mm][l]{ {\fontsize{7}{7}\selectfont 23\%} mother } \makebox[22mm][l]{ {\fontsize{7}{7}\selectfont 6\%} moon } \makebox[22mm][l]{ {\fontsize{7}{7}\selectfont 6\%} time } \makebox[22mm][l]{ {\fontsize{7}{7}\selectfont 6\%} love } \makebox[22mm][l]{ {\fontsize{7}{7}\selectfont 5\%} bed } { {\fontsize{7}{7}\selectfont 5\%} garden } \\
parent & \makebox[22mm][l]{ {\fontsize{7}{7}\selectfont 8\%} evening } \makebox[22mm][l]{ {\fontsize{7}{7}\selectfont 4\%} bedtime } \makebox[22mm][l]{ {\fontsize{7}{7}\selectfont 4\%} felt } \makebox[22mm][l]{ {\fontsize{7}{7}\selectfont 4\%} shimmering } \makebox[22mm][l]{ {\fontsize{7}{7}\selectfont 4\%} glow } { {\fontsize{7}{7}\selectfont 3\%} friend } \\
\bottomrule\\

\end{tabular}
\vspace{-6mm}
\caption{Top words in the \textbf{text of the generated story} that correlate (Pearson) with the sociocultural factor. The terms \textit{child}, \textit{daughter}, and \textit{son} have been removed, as they are almost present at the start of the generation.}
\label{tab:corr_bias_text}
\end{table*}

\paragraph{\textbf{Bias in character-centric attributes.}}
Besides the story text, we study how the protagonist is presented in the stories by exploring the protagonist's attributes introduced in \Cref{sec:taxonomy}. We use the same approach as above and detect the top words for protagonist's attributes in the stories that correlate (Pearson) with the sociocultural factors. Our results (see Table \ref{tab:corr_bias_attr} in Appendix) show that for girls the highest correlated words are \textit{hair}, \textit{gentle}, \textit{imaginative}, and \textit{loving}. Similarly, for boys, \textit{young},  \textit{adventurous}, \textit{hero}, \textit{eager}, and \textit{brave} are the top words. Meanwhile, when we look into nationality, European countries are linked to attributes such as \textit{friendly}, while  \textit{wise} emerges for African nationalities and \textit{pure} and  \textit{gentle} for Asian ones. For ethnicity, we observe a higher correlation of attributes in comparison to nationality and gender. African-American are correlated to the  \textit{heritage} with 38\%; Asians are  \textit{wise} and \textit{noble}, and  \textit{perseverant}; Latinos are family-oriented and Middle-Eastern are \textit{wise} and \textit{generous}. Considering religion factors, descriptors for Jewish characters are also heavily centered around tradition and identity, with \textit{heritage} at 48\% and  \textit{tradition} at 42\%. Meanwhile, atheist protagonists are considered \textit{minded} and \textit{inquisitive}.

Beyond the protagonist's attributes, we can also examine the character traits. \Cref{tab:module_focus} displays the distribution of these traits across all stories and models. Although the differences among models are minor, we observed that these LLMs generate emotional and mental traits in children stories more frequently than others. When analyzing character traits across various sociocultural factors, we observe clear patterns. Physical traits appear 44\% of the top six words in the stories with only gender and 25\% of those related to nationalities.
As one might expect, moral traits appear in 47\% of the top six words, with no references to physical traits. The results are available in \Cref{tab:corr_bias_attr} in  \Cref{appx:protag_analysis}.

\begin{table}[t]
\small
\centering
\begin{tabular}{lccccc}
\toprule
Category & Avg. & GPT4 & Llama3 & Mixtral \\
\midrule
\colorbox{blue!20}{Physical} & 12.7\% & 12.2\% & 19.1\% & 6.5\% \\
\colorbox{red!20}{Emotional} & 29.3\% & 30.4\% & 26.3\% & 31.3\% \\
\colorbox{green!20}{Mental} & 34.2\% & 34.5\% & 33.1\% & 35.0\% \\
\colorbox{yellow!20}{Moral}  & 19.0\% & 20.0\% & 13.4\% & 23.9\% \\
\colorbox{gray!20}{Other} & 4.9\% & 2.9\% & 8.2\% & 3.3\% \\
\bottomrule
\end{tabular}
\caption{The percentage of character traits for protagonist attributes across models.}
\label{tab:module_focus}
\end{table}



\paragraph{\textbf{Bias in context-centric attributes.}} \Cref{tab:context-percentage} presents the percentage of context-centric attributes across sociocultural factors. Comparing these factors reveals differences in geolocation, urban settings, and socioeconomic conditions. When nationality is specified in the prompts, 96.7\% of the stories include a geolocation attribute. Over 52\% of the geolocation attributes are categorized as desert in Egypt and Sudan, and this percentage rises to 76.3\% for Middle-Eastern. Also the majority of the geolocation in stories are linked to green bodies such as forests or hills. Notably, Tajikistan predominantly features mountain-related contexts, aligning with its geographical reality of being home to some of the highest mountains in the world. Additionally, white ethnicity is often associated with magical settings (74.07\%), indicating a potential stereotype rooted in Western fairy-tale traditions. 
The \textit{Not found} category appears frequently for socioeconomic, suggesting that socioeconomic is not often represented in the stories. Iran and Egypt exhibit 15.56\% and 16.30\% association with wealthy socioeconomic which comes from royal figures like princes, reflecting narratives reminiscent of ancient civilizations. 
We observe words such as \textit{poor} and \textit{illness} from the Philippines, which relate to low socioeconomic.
The urban setting shows a high percentage of stories set in villages, suggesting a dominant rural imagery within the urban category. Atheist characters are predominantly associated with magical locations, while urban and socioeconomic factors are absent from the stories.



\begin{table*}[ht]
\centering
\resizebox{\textwidth}{!}{%
\begin{tabular}{ll|cccccc|cccc|ccccc}
\toprule

\multirow{2}{*}{\textbf{Factor}} & \multirow{2}{*}{\textbf{value}} & \multicolumn{6}{c}{\textbf{Geo-location}} & \multicolumn{4}{c}{\textbf{Urban}} & \multicolumn{4}{c}{\textbf{Social economic}} \\
 &   &  \emoji{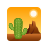} &  \emoji{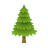} & \emoji{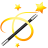} &  \emoji{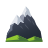} &  \emoji{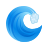} & \emoji{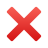} &  \emoji{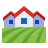}  &  \emoji{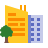} &  \emoji{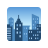} &  \emoji{cross-mark} &  \emoji{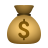} & \emoji{money-bag}\emoji{money-bag} &  \emoji{money-bag}\emoji{money-bag}\emoji{money-bag} &  \emoji{cross-mark} \\
\midrule
 
country & Afghanistan & {\cellcolor[HTML]{EAEAF3}} \color[HTML]{000000} 2.22 & {\cellcolor[HTML]{D1D1F4}} \color[HTML]{000000} 11.85 & {\cellcolor[HTML]{DEDEF4}} \color[HTML]{000000} 5.93 & {\cellcolor[HTML]{3A3AFC}} \color[HTML]{F1F1F1} 75.56 & {\cellcolor[HTML]{F0F0F3}} \color[HTML]{000000} 0.00 & {\cellcolor[HTML]{A0A0F7}} \color[HTML]{F1F1F1} 4.44 & {\cellcolor[HTML]{4A4AFB}} \color[HTML]{F1F1F1} 68.89 & {\cellcolor[HTML]{F0F0F3}} \color[HTML]{000000} 0.00 & {\cellcolor[HTML]{9F9FF7}} \color[HTML]{F1F1F1} 15.56 & {\cellcolor[HTML]{C6C6F5}} \color[HTML]{000000} 15.56 & {\cellcolor[HTML]{B8B8F6}} \color[HTML]{000000} 2.96 & {\cellcolor[HTML]{0000FF}} \color[HTML]{F1F1F1} 25.93 & {\cellcolor[HTML]{B9B9F6}} \color[HTML]{000000} 4.44 & {\cellcolor[HTML]{B0B0F6}} \color[HTML]{000000} 66.67 \\
 country & Armenia & {\cellcolor[HTML]{F0F0F3}} \color[HTML]{000000} 0.00 & {\cellcolor[HTML]{C6C6F5}} \color[HTML]{000000} 16.30 & {\cellcolor[HTML]{E5E5F3}} \color[HTML]{000000} 3.70 & {\cellcolor[HTML]{3232FC}} \color[HTML]{F1F1F1} 78.52 & {\cellcolor[HTML]{E5E5F3}} \color[HTML]{000000} 1.48 & {\cellcolor[HTML]{F0F0F3}} \color[HTML]{000000} 0.00 & {\cellcolor[HTML]{3B3BFC}} \color[HTML]{F1F1F1} 73.33 & {\cellcolor[HTML]{D3D3F4}} \color[HTML]{000000} 1.48 & {\cellcolor[HTML]{DADAF4}} \color[HTML]{000000} 4.44 & {\cellcolor[HTML]{B6B6F6}} \color[HTML]{000000} 20.74 & {\cellcolor[HTML]{D4D4F4}} \color[HTML]{000000} 1.48 & {\cellcolor[HTML]{5A5AFA}} \color[HTML]{F1F1F1} 17.04 & {\cellcolor[HTML]{C2C2F5}} \color[HTML]{000000} 3.70 & {\cellcolor[HTML]{6F6FF9}} \color[HTML]{F1F1F1} 77.78 \\
country & Azerbaijan & {\cellcolor[HTML]{EFEFF3}} \color[HTML]{000000} 0.74 & {\cellcolor[HTML]{A9A9F6}} \color[HTML]{000000} 27.41 & {\cellcolor[HTML]{E7E7F3}} \color[HTML]{000000} 2.96 & {\cellcolor[HTML]{6868FA}} \color[HTML]{F1F1F1} 56.30 & {\cellcolor[HTML]{9797F7}} \color[HTML]{F1F1F1} 11.11 & {\cellcolor[HTML]{D6D6F4}} \color[HTML]{000000} 1.48 & {\cellcolor[HTML]{5A5AFA}} \color[HTML]{F1F1F1} 64.44 & {\cellcolor[HTML]{E2E2F3}} \color[HTML]{000000} 0.74 & {\cellcolor[HTML]{AEAEF6}} \color[HTML]{000000} 12.59 & {\cellcolor[HTML]{B1B1F6}} \color[HTML]{000000} 22.22 & {\cellcolor[HTML]{E2E2F3}} \color[HTML]{000000} 0.74 & {\cellcolor[HTML]{2626FD}} \color[HTML]{F1F1F1} 22.22 & {\cellcolor[HTML]{B0B0F6}} \color[HTML]{000000} 5.19 & {\cellcolor[HTML]{9292F8}} \color[HTML]{F1F1F1} 71.85 \\
country & Brazil & {\cellcolor[HTML]{F0F0F3}} \color[HTML]{000000} 0.00 & {\cellcolor[HTML]{0000FF}} \color[HTML]{F1F1F1} 91.85 & {\cellcolor[HTML]{DEDEF4}} \color[HTML]{000000} 5.93 & {\cellcolor[HTML]{F0F0F3}} \color[HTML]{000000} 0.00 & {\cellcolor[HTML]{E5E5F3}} \color[HTML]{000000} 1.48 & {\cellcolor[HTML]{E3E3F3}} \color[HTML]{000000} 0.74 & {\cellcolor[HTML]{DDDDF4}} \color[HTML]{000000} 25.19 & {\cellcolor[HTML]{1D1DFE}} \color[HTML]{F1F1F1} 10.37 & {\cellcolor[HTML]{DADAF4}} \color[HTML]{000000} 4.44 & {\cellcolor[HTML]{3939FC}} \color[HTML]{F1F1F1} 60.00 & {\cellcolor[HTML]{E2E2F3}} \color[HTML]{000000} 0.74 & {\cellcolor[HTML]{F0F0F3}} \color[HTML]{000000} 2.22 & {\cellcolor[HTML]{F0F0F3}} \color[HTML]{000000} 0.00 & {\cellcolor[HTML]{0000FF}} \color[HTML]{F1F1F1} 97.04 \\
country & China & {\cellcolor[HTML]{F0F0F3}} \color[HTML]{000000} 0.00 & {\cellcolor[HTML]{A7A7F7}} \color[HTML]{000000} 28.15 & {\cellcolor[HTML]{CFCFF4}} \color[HTML]{000000} 10.37 & {\cellcolor[HTML]{6161FA}} \color[HTML]{F1F1F1} 59.26 & {\cellcolor[HTML]{EBEBF3}} \color[HTML]{000000} 0.74 & {\cellcolor[HTML]{D6D6F4}} \color[HTML]{000000} 1.48 & {\cellcolor[HTML]{0909FF}} \color[HTML]{F1F1F1} 88.15 & {\cellcolor[HTML]{F0F0F3}} \color[HTML]{000000} 0.00 & {\cellcolor[HTML]{EDEDF3}} \color[HTML]{000000} 0.74 & {\cellcolor[HTML]{D4D4F4}} \color[HTML]{000000} 11.11 & {\cellcolor[HTML]{7171F9}} \color[HTML]{F1F1F1} 6.67 & {\cellcolor[HTML]{7979F9}} \color[HTML]{F1F1F1} 14.07 & {\cellcolor[HTML]{F0F0F3}} \color[HTML]{000000} 0.00 & {\cellcolor[HTML]{6767FA}} \color[HTML]{F1F1F1} 79.26 \\
country & Egypt & {\cellcolor[HTML]{3131FD}} \color[HTML]{F1F1F1} 60.74 & {\cellcolor[HTML]{F0F0F3}} \color[HTML]{000000} 0.00 & {\cellcolor[HTML]{EAEAF3}} \color[HTML]{000000} 2.22 & {\cellcolor[HTML]{F0F0F3}} \color[HTML]{000000} 0.00 & {\cellcolor[HTML]{0000FF}} \color[HTML]{F1F1F1} 29.63 & {\cellcolor[HTML]{6B6BFA}} \color[HTML]{F1F1F1} 7.41 & {\cellcolor[HTML]{B5B5F6}} \color[HTML]{000000} 37.04 & {\cellcolor[HTML]{E2E2F3}} \color[HTML]{000000} 0.74 & {\cellcolor[HTML]{4A4AFB}} \color[HTML]{F1F1F1} 31.85 & {\cellcolor[HTML]{9797F7}} \color[HTML]{F1F1F1} 30.37 & {\cellcolor[HTML]{D4D4F4}} \color[HTML]{000000} 1.48 & {\cellcolor[HTML]{2D2DFD}} \color[HTML]{F1F1F1} 21.48 & {\cellcolor[HTML]{2E2EFD}} \color[HTML]{F1F1F1} 15.56 & {\cellcolor[HTML]{CECEF4}} \color[HTML]{000000} 61.48 \\
country & Ethiopia & {\cellcolor[HTML]{EDEDF3}} \color[HTML]{000000} 1.48 & {\cellcolor[HTML]{7878F9}} \color[HTML]{F1F1F1} 45.93 & {\cellcolor[HTML]{DBDBF4}} \color[HTML]{000000} 6.67 & {\cellcolor[HTML]{8787F8}} \color[HTML]{F1F1F1} 43.70 & {\cellcolor[HTML]{E5E5F3}} \color[HTML]{000000} 1.48 & {\cellcolor[HTML]{E3E3F3}} \color[HTML]{000000} 0.74 & {\cellcolor[HTML]{4848FB}} \color[HTML]{F1F1F1} 69.63 & {\cellcolor[HTML]{F0F0F3}} \color[HTML]{000000} 0.00 & {\cellcolor[HTML]{DADAF4}} \color[HTML]{000000} 4.44 & {\cellcolor[HTML]{A5A5F7}} \color[HTML]{000000} 25.93 & {\cellcolor[HTML]{7F7FF9}} \color[HTML]{F1F1F1} 5.93 & {\cellcolor[HTML]{BCBCF5}} \color[HTML]{000000} 7.41 & {\cellcolor[HTML]{CCCCF5}} \color[HTML]{000000} 2.96 & {\cellcolor[HTML]{4D4DFB}} \color[HTML]{F1F1F1} 83.70 \\
country & Germany & {\cellcolor[HTML]{F0F0F3}} \color[HTML]{000000} 0.00 & {\cellcolor[HTML]{1919FE}} \color[HTML]{F1F1F1} 82.22 & {\cellcolor[HTML]{D4D4F4}} \color[HTML]{000000} 8.89 & {\cellcolor[HTML]{DEDFF4}} \color[HTML]{000000} 7.41 & {\cellcolor[HTML]{E5E5F3}} \color[HTML]{000000} 1.48 & {\cellcolor[HTML]{F0F0F3}} \color[HTML]{000000} 0.00 & {\cellcolor[HTML]{2C2CFD}} \color[HTML]{F1F1F1} 77.78 & {\cellcolor[HTML]{B5B5F6}} \color[HTML]{000000} 2.96 & {\cellcolor[HTML]{EDEDF3}} \color[HTML]{000000} 0.74 & {\cellcolor[HTML]{BDBDF5}} \color[HTML]{000000} 18.52 & {\cellcolor[HTML]{F0F0F3}} \color[HTML]{000000} 0.00 & {\cellcolor[HTML]{4444FC}} \color[HTML]{F1F1F1} 19.26 & {\cellcolor[HTML]{E8E8F3}} \color[HTML]{000000} 0.74 & {\cellcolor[HTML]{6363FA}} \color[HTML]{F1F1F1} 80.00 \\
country & Great Britain & {\cellcolor[HTML]{DFDFF4}} \color[HTML]{000000} 5.56 & {\cellcolor[HTML]{8787F8}} \color[HTML]{F1F1F1} 40.42 & {\cellcolor[HTML]{7B7BF9}} \color[HTML]{F1F1F1} 36.25 & {\cellcolor[HTML]{CFCFF4}} \color[HTML]{000000} 13.89 & {\cellcolor[HTML]{E5E5F3}} \color[HTML]{000000} 1.39 & {\cellcolor[HTML]{C3C3F5}} \color[HTML]{000000} 2.50 & {\cellcolor[HTML]{7272F9}} \color[HTML]{F1F1F1} 57.08 & {\cellcolor[HTML]{B0B0F6}} \color[HTML]{000000} 3.19 & {\cellcolor[HTML]{E2E2F3}} \color[HTML]{000000} 2.78 & {\cellcolor[HTML]{8282F8}} \color[HTML]{F1F1F1} 36.94 & {\cellcolor[HTML]{D3D3F4}} \color[HTML]{000000} 1.53 & {\cellcolor[HTML]{6F6FF9}} \color[HTML]{F1F1F1} 15.00 & {\cellcolor[HTML]{5B5BFA}} \color[HTML]{F1F1F1} 11.94 & {\cellcolor[HTML]{9494F7}} \color[HTML]{F1F1F1} 71.53 \\
country & India & {\cellcolor[HTML]{EDEDF3}} \color[HTML]{000000} 1.48 & {\cellcolor[HTML]{6161FA}} \color[HTML]{F1F1F1} 54.81 & {\cellcolor[HTML]{D1D1F4}} \color[HTML]{000000} 9.63 & {\cellcolor[HTML]{BBBBF5}} \color[HTML]{000000} 22.22 & {\cellcolor[HTML]{D3D3F4}} \color[HTML]{000000} 3.70 & {\cellcolor[HTML]{5D5DFA}} \color[HTML]{F1F1F1} 8.15 & {\cellcolor[HTML]{1919FE}} \color[HTML]{F1F1F1} 83.70 & {\cellcolor[HTML]{E2E2F3}} \color[HTML]{000000} 0.74 & {\cellcolor[HTML]{E5E5F3}} \color[HTML]{000000} 2.22 & {\cellcolor[HTML]{CECEF5}} \color[HTML]{000000} 13.33 & {\cellcolor[HTML]{AAAAF6}} \color[HTML]{000000} 3.70 & {\cellcolor[HTML]{8888F8}} \color[HTML]{F1F1F1} 12.59 & {\cellcolor[HTML]{B0B0F6}} \color[HTML]{000000} 5.19 & {\cellcolor[HTML]{6B6BFA}} \color[HTML]{F1F1F1} 78.52 \\
country & Indonesia & {\cellcolor[HTML]{F0F0F3}} \color[HTML]{000000} 0.00 & {\cellcolor[HTML]{0808FF}} \color[HTML]{F1F1F1} 88.89 & {\cellcolor[HTML]{DBDBF4}} \color[HTML]{000000} 6.67 & {\cellcolor[HTML]{EAEAF3}} \color[HTML]{000000} 2.96 & {\cellcolor[HTML]{EBEBF3}} \color[HTML]{000000} 0.74 & {\cellcolor[HTML]{E3E3F3}} \color[HTML]{000000} 0.74 & {\cellcolor[HTML]{4A4AFB}} \color[HTML]{F1F1F1} 68.89 & {\cellcolor[HTML]{E2E2F3}} \color[HTML]{000000} 0.74 & {\cellcolor[HTML]{EDEDF3}} \color[HTML]{000000} 0.74 & {\cellcolor[HTML]{9A9AF7}} \color[HTML]{F1F1F1} 29.63 & {\cellcolor[HTML]{C6C6F5}} \color[HTML]{000000} 2.22 & {\cellcolor[HTML]{7979F9}} \color[HTML]{F1F1F1} 14.07 & {\cellcolor[HTML]{E8E8F3}} \color[HTML]{000000} 0.74 & {\cellcolor[HTML]{5151FB}} \color[HTML]{F1F1F1} 82.96 \\
country & Iran & {\cellcolor[HTML]{BFBFF5}} \color[HTML]{000000} 15.56 & {\cellcolor[HTML]{B9B9F6}} \color[HTML]{000000} 21.48 & {\cellcolor[HTML]{9595F7}} \color[HTML]{F1F1F1} 28.15 & {\cellcolor[HTML]{B2B2F6}} \color[HTML]{000000} 25.93 & {\cellcolor[HTML]{E5E5F3}} \color[HTML]{000000} 1.48 & {\cellcolor[HTML]{6B6BFA}} \color[HTML]{F1F1F1} 7.41 & {\cellcolor[HTML]{B5B5F6}} \color[HTML]{000000} 37.04 & {\cellcolor[HTML]{B5B5F6}} \color[HTML]{000000} 2.96 & {\cellcolor[HTML]{6969FA}} \color[HTML]{F1F1F1} 25.93 & {\cellcolor[HTML]{8C8CF8}} \color[HTML]{F1F1F1} 34.07 & {\cellcolor[HTML]{D4D4F4}} \color[HTML]{000000} 1.48 & {\cellcolor[HTML]{3535FC}} \color[HTML]{F1F1F1} 20.74 & {\cellcolor[HTML]{2525FD}} \color[HTML]{F1F1F1} 16.30 & {\cellcolor[HTML]{CECEF4}} \color[HTML]{000000} 61.48 \\
country & Iraq & {\cellcolor[HTML]{6D6DF9}} \color[HTML]{F1F1F1} 41.48 & {\cellcolor[HTML]{CECEF5}} \color[HTML]{000000} 13.33 & {\cellcolor[HTML]{C0C0F5}} \color[HTML]{000000} 14.81 & {\cellcolor[HTML]{F0F0F3}} \color[HTML]{000000} 0.00 & {\cellcolor[HTML]{2424FD}} \color[HTML]{F1F1F1} 25.19 & {\cellcolor[HTML]{9393F8}} \color[HTML]{F1F1F1} 5.19 & {\cellcolor[HTML]{BDBDF5}} \color[HTML]{000000} 34.81 & {\cellcolor[HTML]{E2E2F3}} \color[HTML]{000000} 0.74 & {\cellcolor[HTML]{0000FF}} \color[HTML]{F1F1F1} 45.93 & {\cellcolor[HTML]{BDBDF5}} \color[HTML]{000000} 18.52 & {\cellcolor[HTML]{D4D4F4}} \color[HTML]{000000} 1.48 & {\cellcolor[HTML]{1717FE}} \color[HTML]{F1F1F1} 23.70 & {\cellcolor[HTML]{9D9DF7}} \color[HTML]{F1F1F1} 6.67 & {\cellcolor[HTML]{A8A8F6}} \color[HTML]{000000} 68.15 \\
country & Italy & {\cellcolor[HTML]{DEDEF4}} \color[HTML]{000000} 6.11 & {\cellcolor[HTML]{7676F9}} \color[HTML]{F1F1F1} 46.67 & {\cellcolor[HTML]{A2A2F7}} \color[HTML]{F1F1F1} 24.03 & {\cellcolor[HTML]{C4C4F5}} \color[HTML]{000000} 18.33 & {\cellcolor[HTML]{E7E7F3}} \color[HTML]{000000} 1.25 & {\cellcolor[HTML]{AFAFF6}} \color[HTML]{000000} 3.61 & {\cellcolor[HTML]{5B5BFA}} \color[HTML]{F1F1F1} 63.89 & {\cellcolor[HTML]{7D7DF9}} \color[HTML]{F1F1F1} 5.69 & {\cellcolor[HTML]{DADAF4}} \color[HTML]{000000} 4.31 & {\cellcolor[HTML]{A5A5F7}} \color[HTML]{000000} 26.11 & {\cellcolor[HTML]{CECEF4}} \color[HTML]{000000} 1.81 & {\cellcolor[HTML]{4040FC}} \color[HTML]{F1F1F1} 19.58 & {\cellcolor[HTML]{ACACF6}} \color[HTML]{000000} 5.56 & {\cellcolor[HTML]{8B8BF8}} \color[HTML]{F1F1F1} 73.06 \\
country & Japan & {\cellcolor[HTML]{F0F0F3}} \color[HTML]{000000} 0.00 & {\cellcolor[HTML]{7474F9}} \color[HTML]{F1F1F1} 47.41 & {\cellcolor[HTML]{CDCDF5}} \color[HTML]{000000} 11.11 & {\cellcolor[HTML]{8F8FF8}} \color[HTML]{F1F1F1} 40.00 & {\cellcolor[HTML]{EBEBF3}} \color[HTML]{000000} 0.74 & {\cellcolor[HTML]{E3E3F3}} \color[HTML]{000000} 0.74 & {\cellcolor[HTML]{1D1DFE}} \color[HTML]{F1F1F1} 82.22 & {\cellcolor[HTML]{B5B5F6}} \color[HTML]{000000} 2.96 & {\cellcolor[HTML]{F0F0F3}} \color[HTML]{000000} 0.00 & {\cellcolor[HTML]{C9C9F5}} \color[HTML]{000000} 14.81 & {\cellcolor[HTML]{D4D4F4}} \color[HTML]{000000} 1.48 & {\cellcolor[HTML]{9797F7}} \color[HTML]{F1F1F1} 11.11 & {\cellcolor[HTML]{F0F0F3}} \color[HTML]{000000} 0.00 & {\cellcolor[HTML]{3838FC}} \color[HTML]{F1F1F1} 87.41 \\
country & Kenya & {\cellcolor[HTML]{EFEFF3}} \color[HTML]{000000} 0.74 & {\cellcolor[HTML]{2A2AFD}} \color[HTML]{F1F1F1} 75.56 & {\cellcolor[HTML]{F0F0F3}} \color[HTML]{000000} 0.00 & {\cellcolor[HTML]{BBBBF5}} \color[HTML]{000000} 22.22 & {\cellcolor[HTML]{EBEBF3}} \color[HTML]{000000} 0.74 & {\cellcolor[HTML]{E3E3F3}} \color[HTML]{000000} 0.74 & {\cellcolor[HTML]{6F6FF9}} \color[HTML]{F1F1F1} 57.78 & {\cellcolor[HTML]{F0F0F3}} \color[HTML]{000000} 0.00 & {\cellcolor[HTML]{F0F0F3}} \color[HTML]{000000} 0.00 & {\cellcolor[HTML]{7171F9}} \color[HTML]{F1F1F1} 42.22 & {\cellcolor[HTML]{B8B8F6}} \color[HTML]{000000} 2.96 & {\cellcolor[HTML]{EAEAF3}} \color[HTML]{000000} 2.96 & {\cellcolor[HTML]{F0F0F3}} \color[HTML]{000000} 0.00 & {\cellcolor[HTML]{1111FE}} \color[HTML]{F1F1F1} 94.07 \\
country & Mali & {\cellcolor[HTML]{8A8AF8}} \color[HTML]{F1F1F1} 32.59 & {\cellcolor[HTML]{6C6CF9}} \color[HTML]{F1F1F1} 50.37 & {\cellcolor[HTML]{ECECF3}} \color[HTML]{000000} 1.48 & {\cellcolor[HTML]{EEEEF3}} \color[HTML]{000000} 1.48 & {\cellcolor[HTML]{9D9DF7}} \color[HTML]{F1F1F1} 10.37 & {\cellcolor[HTML]{ADADF6}} \color[HTML]{000000} 3.70 & {\cellcolor[HTML]{2525FD}} \color[HTML]{F1F1F1} 80.00 & {\cellcolor[HTML]{F0F0F3}} \color[HTML]{000000} 0.00 & {\cellcolor[HTML]{DEDEF4}} \color[HTML]{000000} 3.70 & {\cellcolor[HTML]{C4C4F5}} \color[HTML]{000000} 16.30 & {\cellcolor[HTML]{5555FB}} \color[HTML]{F1F1F1} 8.15 & {\cellcolor[HTML]{DBDBF4}} \color[HTML]{000000} 4.44 & {\cellcolor[HTML]{F0F0F3}} \color[HTML]{000000} 0.00 & {\cellcolor[HTML]{3838FC}} \color[HTML]{F1F1F1} 87.41 \\
country & Mexico & {\cellcolor[HTML]{E0E0F4}} \color[HTML]{000000} 5.19 & {\cellcolor[HTML]{5F5FFA}} \color[HTML]{F1F1F1} 55.56 & {\cellcolor[HTML]{CFCFF4}} \color[HTML]{000000} 10.37 & {\cellcolor[HTML]{BDBDF5}} \color[HTML]{000000} 21.48 & {\cellcolor[HTML]{D9D9F4}} \color[HTML]{000000} 2.96 & {\cellcolor[HTML]{A0A0F7}} \color[HTML]{F1F1F1} 4.44 & {\cellcolor[HTML]{4848FB}} \color[HTML]{F1F1F1} 69.63 & {\cellcolor[HTML]{0000FF}} \color[HTML]{F1F1F1} 11.85 & {\cellcolor[HTML]{EDEDF3}} \color[HTML]{000000} 0.74 & {\cellcolor[HTML]{BFBFF5}} \color[HTML]{000000} 17.78 & {\cellcolor[HTML]{9C9CF7}} \color[HTML]{F1F1F1} 4.44 & {\cellcolor[HTML]{6969FA}} \color[HTML]{F1F1F1} 15.56 & {\cellcolor[HTML]{F0F0F3}} \color[HTML]{000000} 0.00 & {\cellcolor[HTML]{6363FA}} \color[HTML]{F1F1F1} 80.00 \\
country & Nigeria & {\cellcolor[HTML]{F0F0F3}} \color[HTML]{000000} 0.00 & {\cellcolor[HTML]{0B0BFE}} \color[HTML]{F1F1F1} 87.41 & {\cellcolor[HTML]{E0E0F4}} \color[HTML]{000000} 5.19 & {\cellcolor[HTML]{F0F0F3}} \color[HTML]{000000} 0.00 & {\cellcolor[HTML]{DEDFF4}} \color[HTML]{000000} 2.22 & {\cellcolor[HTML]{9393F8}} \color[HTML]{F1F1F1} 5.19 & {\cellcolor[HTML]{1D1DFE}} \color[HTML]{F1F1F1} 82.22 & {\cellcolor[HTML]{D3D3F4}} \color[HTML]{000000} 1.48 & {\cellcolor[HTML]{E1E1F4}} \color[HTML]{000000} 2.96 & {\cellcolor[HTML]{CECEF5}} \color[HTML]{000000} 13.33 & {\cellcolor[HTML]{8D8DF8}} \color[HTML]{F1F1F1} 5.19 & {\cellcolor[HTML]{C3C3F5}} \color[HTML]{000000} 6.67 & {\cellcolor[HTML]{F0F0F3}} \color[HTML]{000000} 0.00 & {\cellcolor[HTML]{3333FC}} \color[HTML]{F1F1F1} 88.15 \\
country & Philippines & {\cellcolor[HTML]{F0F0F3}} \color[HTML]{000000} 0.00 & {\cellcolor[HTML]{3B3BFC}} \color[HTML]{F1F1F1} 68.89 & {\cellcolor[HTML]{DEDEF4}} \color[HTML]{000000} 5.93 & {\cellcolor[HTML]{CDCDF5}} \color[HTML]{000000} 14.81 & {\cellcolor[HTML]{AEAEF6}} \color[HTML]{000000} 8.15 & {\cellcolor[HTML]{C9C9F5}} \color[HTML]{000000} 2.22 & {\cellcolor[HTML]{2F2FFD}} \color[HTML]{F1F1F1} 77.04 & {\cellcolor[HTML]{7878F9}} \color[HTML]{F1F1F1} 5.93 & {\cellcolor[HTML]{E9E9F3}} \color[HTML]{000000} 1.48 & {\cellcolor[HTML]{C6C6F5}} \color[HTML]{000000} 15.56 & {\cellcolor[HTML]{0000FF}} \color[HTML]{F1F1F1} 12.59 & {\cellcolor[HTML]{D2D2F4}} \color[HTML]{000000} 5.19 & {\cellcolor[HTML]{F0F0F3}} \color[HTML]{000000} 0.00 & {\cellcolor[HTML]{5656FB}} \color[HTML]{F1F1F1} 82.22 \\
country & Russia & {\cellcolor[HTML]{F0F0F3}} \color[HTML]{000000} 0.00 & {\cellcolor[HTML]{5F5FFA}} \color[HTML]{F1F1F1} 55.56 & {\cellcolor[HTML]{9595F7}} \color[HTML]{F1F1F1} 28.15 & {\cellcolor[HTML]{DCDCF4}} \color[HTML]{000000} 8.89 & {\cellcolor[HTML]{EBEBF3}} \color[HTML]{000000} 0.74 & {\cellcolor[HTML]{7878F9}} \color[HTML]{F1F1F1} 6.67 & {\cellcolor[HTML]{4545FB}} \color[HTML]{F1F1F1} 70.37 & {\cellcolor[HTML]{F0F0F3}} \color[HTML]{000000} 0.00 & {\cellcolor[HTML]{EDEDF3}} \color[HTML]{000000} 0.74 & {\cellcolor[HTML]{9C9CF7}} \color[HTML]{F1F1F1} 28.89 & {\cellcolor[HTML]{C6C6F5}} \color[HTML]{000000} 2.22 & {\cellcolor[HTML]{7979F9}} \color[HTML]{F1F1F1} 14.07 & {\cellcolor[HTML]{DEDFF4}} \color[HTML]{000000} 1.48 & {\cellcolor[HTML]{5656FB}} \color[HTML]{F1F1F1} 82.22 \\
country & South Africa & {\cellcolor[HTML]{EDEDF3}} \color[HTML]{000000} 1.48 & {\cellcolor[HTML]{5555FB}} \color[HTML]{F1F1F1} 59.26 & {\cellcolor[HTML]{D6D6F4}} \color[HTML]{000000} 8.15 & {\cellcolor[HTML]{B2B2F6}} \color[HTML]{000000} 25.93 & {\cellcolor[HTML]{DEDFF4}} \color[HTML]{000000} 2.22 & {\cellcolor[HTML]{BCBCF5}} \color[HTML]{000000} 2.96 & {\cellcolor[HTML]{A1A1F7}} \color[HTML]{F1F1F1} 42.96 & {\cellcolor[HTML]{F0F0F3}} \color[HTML]{000000} 0.00 & {\cellcolor[HTML]{EDEDF3}} \color[HTML]{000000} 0.74 & {\cellcolor[HTML]{4444FC}} \color[HTML]{F1F1F1} 56.30 & {\cellcolor[HTML]{AAAAF6}} \color[HTML]{000000} 3.70 & {\cellcolor[HTML]{D2D2F4}} \color[HTML]{000000} 5.19 & {\cellcolor[HTML]{E8E8F3}} \color[HTML]{000000} 0.74 & {\cellcolor[HTML]{2727FD}} \color[HTML]{F1F1F1} 90.37 \\
country & Sri Lanka & {\cellcolor[HTML]{F0F0F3}} \color[HTML]{000000} 0.00 & {\cellcolor[HTML]{2323FD}} \color[HTML]{F1F1F1} 78.52 & {\cellcolor[HTML]{E7E7F3}} \color[HTML]{000000} 2.96 & {\cellcolor[HTML]{D6D6F4}} \color[HTML]{000000} 11.11 & {\cellcolor[HTML]{B4B4F6}} \color[HTML]{000000} 7.41 & {\cellcolor[HTML]{F0F0F3}} \color[HTML]{000000} 0.00 & {\cellcolor[HTML]{5757FB}} \color[HTML]{F1F1F1} 65.19 & {\cellcolor[HTML]{F0F0F3}} \color[HTML]{000000} 0.00 & {\cellcolor[HTML]{EDEDF3}} \color[HTML]{000000} 0.74 & {\cellcolor[HTML]{8C8CF8}} \color[HTML]{F1F1F1} 34.07 & {\cellcolor[HTML]{E2E2F3}} \color[HTML]{000000} 0.74 & {\cellcolor[HTML]{ADADF6}} \color[HTML]{000000} 8.89 & {\cellcolor[HTML]{CCCCF5}} \color[HTML]{000000} 2.96 & {\cellcolor[HTML]{3838FC}} \color[HTML]{F1F1F1} 87.41 \\
country & Sudan & {\cellcolor[HTML]{4A4AFB}} \color[HTML]{F1F1F1} 52.59 & {\cellcolor[HTML]{BBBBF5}} \color[HTML]{000000} 20.74 & {\cellcolor[HTML]{DEDEF4}} \color[HTML]{000000} 5.93 & {\cellcolor[HTML]{EAEAF3}} \color[HTML]{000000} 2.96 & {\cellcolor[HTML]{7979F9}} \color[HTML]{F1F1F1} 14.81 & {\cellcolor[HTML]{BCBCF5}} \color[HTML]{000000} 2.96 & {\cellcolor[HTML]{8686F8}} \color[HTML]{F1F1F1} 51.11 & {\cellcolor[HTML]{F0F0F3}} \color[HTML]{000000} 0.00 & {\cellcolor[HTML]{D6D6F4}} \color[HTML]{000000} 5.19 & {\cellcolor[HTML]{6C6CF9}} \color[HTML]{F1F1F1} 43.70 & {\cellcolor[HTML]{7171F9}} \color[HTML]{F1F1F1} 6.67 & {\cellcolor[HTML]{F0F0F3}} \color[HTML]{000000} 2.22 & {\cellcolor[HTML]{E8E8F3}} \color[HTML]{000000} 0.74 & {\cellcolor[HTML]{2727FD}} \color[HTML]{F1F1F1} 90.37 \\
country & Tajikistan & {\cellcolor[HTML]{F0F0F3}} \color[HTML]{000000} 0.00 & {\cellcolor[HTML]{F0F0F3}} \color[HTML]{000000} 0.00 & {\cellcolor[HTML]{EFEFF3}} \color[HTML]{000000} 0.74 & {\cellcolor[HTML]{0000FF}} \color[HTML]{F1F1F1} 99.26 & {\cellcolor[HTML]{F0F0F3}} \color[HTML]{000000} 0.00 & {\cellcolor[HTML]{F0F0F3}} \color[HTML]{000000} 0.00 & {\cellcolor[HTML]{2929FD}} \color[HTML]{F1F1F1} 78.52 & {\cellcolor[HTML]{C4C4F5}} \color[HTML]{000000} 2.22 & {\cellcolor[HTML]{F0F0F3}} \color[HTML]{000000} 0.00 & {\cellcolor[HTML]{BBBBF5}} \color[HTML]{000000} 19.26 & {\cellcolor[HTML]{B8B8F6}} \color[HTML]{000000} 2.96 & {\cellcolor[HTML]{9797F7}} \color[HTML]{F1F1F1} 11.11 & {\cellcolor[HTML]{DEDFF4}} \color[HTML]{000000} 1.48 & {\cellcolor[HTML]{4949FB}} \color[HTML]{F1F1F1} 84.44 \\
country & Thailand & {\cellcolor[HTML]{F0F0F3}} \color[HTML]{000000} 0.00 & {\cellcolor[HTML]{1111FE}} \color[HTML]{F1F1F1} 85.19 & {\cellcolor[HTML]{DEDEF4}} \color[HTML]{000000} 5.93 & {\cellcolor[HTML]{E2E2F3}} \color[HTML]{000000} 5.93 & {\cellcolor[HTML]{E5E5F3}} \color[HTML]{000000} 1.48 & {\cellcolor[HTML]{D6D6F4}} \color[HTML]{000000} 1.48 & {\cellcolor[HTML]{8D8DF8}} \color[HTML]{F1F1F1} 48.89 & {\cellcolor[HTML]{F0F0F3}} \color[HTML]{000000} 0.00 & {\cellcolor[HTML]{D6D6F4}} \color[HTML]{000000} 5.19 & {\cellcolor[HTML]{6565FA}} \color[HTML]{F1F1F1} 45.93 & {\cellcolor[HTML]{C6C6F5}} \color[HTML]{000000} 2.22 & {\cellcolor[HTML]{9E9EF7}} \color[HTML]{F1F1F1} 10.37 & {\cellcolor[HTML]{B0B0F6}} \color[HTML]{000000} 5.19 & {\cellcolor[HTML]{5656FB}} \color[HTML]{F1F1F1} 82.22 \\
country & United States & {\cellcolor[HTML]{DDDDF4}} \color[HTML]{000000} 6.53 & {\cellcolor[HTML]{9A9AF7}} \color[HTML]{F1F1F1} 33.19 & {\cellcolor[HTML]{6F6FF9}} \color[HTML]{F1F1F1} 39.72 & {\cellcolor[HTML]{CDCDF5}} \color[HTML]{000000} 14.86 & {\cellcolor[HTML]{E3E3F3}} \color[HTML]{000000} 1.67 & {\cellcolor[HTML]{A8A8F6}} \color[HTML]{000000} 4.03 & {\cellcolor[HTML]{9494F7}} \color[HTML]{F1F1F1} 46.94 & {\cellcolor[HTML]{6666FA}} \color[HTML]{F1F1F1} 6.81 & {\cellcolor[HTML]{DEDEF4}} \color[HTML]{000000} 3.61 & {\cellcolor[HTML]{7070F9}} \color[HTML]{F1F1F1} 42.64 & {\cellcolor[HTML]{C4C4F5}} \color[HTML]{000000} 2.36 & {\cellcolor[HTML]{8989F8}} \color[HTML]{F1F1F1} 12.50 & {\cellcolor[HTML]{B2B2F6}} \color[HTML]{000000} 5.00 & {\cellcolor[HTML]{6262FA}} \color[HTML]{F1F1F1} 80.14 \\
country & Vietnam & {\cellcolor[HTML]{F0F0F3}} \color[HTML]{000000} 0.00 & {\cellcolor[HTML]{3030FD}} \color[HTML]{F1F1F1} 73.33 & {\cellcolor[HTML]{E7E7F3}} \color[HTML]{000000} 2.96 & {\cellcolor[HTML]{D4D4F4}} \color[HTML]{000000} 11.85 & {\cellcolor[HTML]{AEAEF6}} \color[HTML]{000000} 8.15 & {\cellcolor[HTML]{ADADF6}} \color[HTML]{000000} 3.70 & {\cellcolor[HTML]{1919FE}} \color[HTML]{F1F1F1} 83.70 & {\cellcolor[HTML]{F0F0F3}} \color[HTML]{000000} 0.00 & {\cellcolor[HTML]{D6D6F4}} \color[HTML]{000000} 5.19 & {\cellcolor[HTML]{D4D4F4}} \color[HTML]{000000} 11.11 & {\cellcolor[HTML]{7F7FF9}} \color[HTML]{F1F1F1} 5.93 & {\cellcolor[HTML]{3535FC}} \color[HTML]{F1F1F1} 20.74 & {\cellcolor[HTML]{F0F0F3}} \color[HTML]{000000} 0.00 & {\cellcolor[HTML]{8A8AF8}} \color[HTML]{F1F1F1} 73.33 \\
 \midrule
ethnicity & African-American & {\cellcolor[HTML]{EDEDF3}} \color[HTML]{000000} 1.48 & {\cellcolor[HTML]{4242FC}} \color[HTML]{F1F1F1} 66.67 & {\cellcolor[HTML]{9D9DF7}} \color[HTML]{F1F1F1} 25.93 & {\cellcolor[HTML]{EFEFF3}} \color[HTML]{000000} 0.74 & {\cellcolor[HTML]{E5E5F3}} \color[HTML]{000000} 1.48 & {\cellcolor[HTML]{ADADF6}} \color[HTML]{000000} 3.70 & {\cellcolor[HTML]{7E7EF9}} \color[HTML]{F1F1F1} 53.33 & {\cellcolor[HTML]{A6A6F7}} \color[HTML]{000000} 3.70 & {\cellcolor[HTML]{C6C6F5}} \color[HTML]{000000} 8.15 & {\cellcolor[HTML]{8989F8}} \color[HTML]{F1F1F1} 34.81 & {\cellcolor[HTML]{D4D4F4}} \color[HTML]{000000} 1.48 & {\cellcolor[HTML]{D2D2F4}} \color[HTML]{000000} 5.19 & {\cellcolor[HTML]{9494F7}} \color[HTML]{F1F1F1} 7.41 & {\cellcolor[HTML]{4040FC}} \color[HTML]{F1F1F1} 85.93 \\
 ethnicity & Asian & {\cellcolor[HTML]{F0F0F3}} \color[HTML]{000000} 0.00 & {\cellcolor[HTML]{6C6CF9}} \color[HTML]{F1F1F1} 50.37 & {\cellcolor[HTML]{ADADF6}} \color[HTML]{000000} 20.74 & {\cellcolor[HTML]{ADADF6}} \color[HTML]{000000} 28.15 & {\cellcolor[HTML]{EBEBF3}} \color[HTML]{000000} 0.74 & {\cellcolor[HTML]{F0F0F3}} \color[HTML]{000000} 0.00 & {\cellcolor[HTML]{2525FD}} \color[HTML]{F1F1F1} 80.00 & {\cellcolor[HTML]{F0F0F3}} \color[HTML]{000000} 0.00 & {\cellcolor[HTML]{F0F0F3}} \color[HTML]{000000} 0.00 & {\cellcolor[HTML]{B8B8F6}} \color[HTML]{000000} 20.00 & {\cellcolor[HTML]{AAAAF6}} \color[HTML]{000000} 3.70 & {\cellcolor[HTML]{4444FC}} \color[HTML]{F1F1F1} 19.26 & {\cellcolor[HTML]{D5D5F4}} \color[HTML]{000000} 2.22 & {\cellcolor[HTML]{8181F8}} \color[HTML]{F1F1F1} 74.81 \\
ethnicity & European-American & {\cellcolor[HTML]{F0F0F3}} \color[HTML]{000000} 0.00 & {\cellcolor[HTML]{6E6EF9}} \color[HTML]{F1F1F1} 49.63 & {\cellcolor[HTML]{6C6CF9}} \color[HTML]{F1F1F1} 40.74 & {\cellcolor[HTML]{DEDFF4}} \color[HTML]{000000} 7.41 & {\cellcolor[HTML]{DEDFF4}} \color[HTML]{000000} 2.22 & {\cellcolor[HTML]{F0F0F3}} \color[HTML]{000000} 0.00 & {\cellcolor[HTML]{5757FB}} \color[HTML]{F1F1F1} 65.19 & {\cellcolor[HTML]{A6A6F7}} \color[HTML]{000000} 3.70 & {\cellcolor[HTML]{E9E9F3}} \color[HTML]{000000} 1.48 & {\cellcolor[HTML]{9A9AF7}} \color[HTML]{F1F1F1} 29.63 & {\cellcolor[HTML]{F0F0F3}} \color[HTML]{000000} 0.00 & {\cellcolor[HTML]{4444FC}} \color[HTML]{F1F1F1} 19.26 & {\cellcolor[HTML]{6F6FF9}} \color[HTML]{F1F1F1} 10.37 & {\cellcolor[HTML]{9B9BF7}} \color[HTML]{F1F1F1} 70.37 \\
ethnicity & Latino & {\cellcolor[HTML]{EFEFF3}} \color[HTML]{000000} 0.74 & {\cellcolor[HTML]{7E7EF9}} \color[HTML]{F1F1F1} 43.70 & {\cellcolor[HTML]{C5C5F5}} \color[HTML]{000000} 13.33 & {\cellcolor[HTML]{9797F7}} \color[HTML]{F1F1F1} 37.04 & {\cellcolor[HTML]{DEDFF4}} \color[HTML]{000000} 2.22 & {\cellcolor[HTML]{BCBCF5}} \color[HTML]{000000} 2.96 & {\cellcolor[HTML]{0C0CFE}} \color[HTML]{F1F1F1} 87.41 & {\cellcolor[HTML]{4A4AFB}} \color[HTML]{F1F1F1} 8.15 & {\cellcolor[HTML]{F0F0F3}} \color[HTML]{000000} 0.00 & {\cellcolor[HTML]{EAEAF3}} \color[HTML]{000000} 4.44 & {\cellcolor[HTML]{AAAAF6}} \color[HTML]{000000} 3.70 & {\cellcolor[HTML]{4444FC}} \color[HTML]{F1F1F1} 19.26 & {\cellcolor[HTML]{F0F0F3}} \color[HTML]{000000} 0.00 & {\cellcolor[HTML]{7474F9}} \color[HTML]{F1F1F1} 77.04 \\
 ethnicity & Middle-Eastern & {\cellcolor[HTML]{0000FF}} \color[HTML]{F1F1F1} 76.30 & {\cellcolor[HTML]{EDEDF3}} \color[HTML]{000000} 1.48 & {\cellcolor[HTML]{D4D4F4}} \color[HTML]{000000} 8.89 & {\cellcolor[HTML]{EAEAF3}} \color[HTML]{000000} 2.96 & {\cellcolor[HTML]{D3D3F4}} \color[HTML]{000000} 3.70 & {\cellcolor[HTML]{7878F9}} \color[HTML]{F1F1F1} 6.67 & {\cellcolor[HTML]{8D8DF8}} \color[HTML]{F1F1F1} 48.89 & {\cellcolor[HTML]{A6A6F7}} \color[HTML]{000000} 3.70 & {\cellcolor[HTML]{6161FA}} \color[HTML]{F1F1F1} 27.41 & {\cellcolor[HTML]{B8B8F6}} \color[HTML]{000000} 20.00 & {\cellcolor[HTML]{9C9CF7}} \color[HTML]{F1F1F1} 4.44 & {\cellcolor[HTML]{3535FC}} \color[HTML]{F1F1F1} 20.74 & {\cellcolor[HTML]{0000FF}} \color[HTML]{F1F1F1} 19.26 & {\cellcolor[HTML]{F0F0F3}} \color[HTML]{000000} 55.56 \\

ethnicity & White & {\cellcolor[HTML]{F0F0F3}} \color[HTML]{000000} 0.00 & {\cellcolor[HTML]{B7B7F6}} \color[HTML]{000000} 22.22 & {\cellcolor[HTML]{0000FF}} \color[HTML]{F1F1F1} 74.07 & {\cellcolor[HTML]{ECECF3}} \color[HTML]{000000} 2.22 & {\cellcolor[HTML]{EBEBF3}} \color[HTML]{000000} 0.74 & {\cellcolor[HTML]{E3E3F3}} \color[HTML]{000000} 0.74 & {\cellcolor[HTML]{F0F0F3}} \color[HTML]{000000} 19.26 & {\cellcolor[HTML]{B5B5F6}} \color[HTML]{000000} 2.96 & {\cellcolor[HTML]{F0F0F3}} \color[HTML]{000000} 0.00 & {\cellcolor[HTML]{0000FF}} \color[HTML]{F1F1F1} 77.78 & {\cellcolor[HTML]{F0F0F3}} \color[HTML]{000000} 0.00 & {\cellcolor[HTML]{8888F8}} \color[HTML]{F1F1F1} 12.59 & {\cellcolor[HTML]{0000FF}} \color[HTML]{F1F1F1} 19.26 & {\cellcolor[HTML]{A8A8F6}} \color[HTML]{000000} 68.15 \\
 \midrule
gender & child & {\cellcolor[HTML]{D9D9F4}} \color[HTML]{000000} 7.64 & {\cellcolor[HTML]{7979F9}} \color[HTML]{F1F1F1} 45.75 & {\cellcolor[HTML]{B7B7F6}} \color[HTML]{000000} 17.89 & {\cellcolor[HTML]{BCBCF5}} \color[HTML]{000000} 21.79 & {\cellcolor[HTML]{D2D2F4}} \color[HTML]{000000} 3.79 & {\cellcolor[HTML]{B8B8F6}} \color[HTML]{000000} 3.14 & {\cellcolor[HTML]{6E6EF9}} \color[HTML]{F1F1F1} 58.16 & {\cellcolor[HTML]{ACACF6}} \color[HTML]{000000} 3.41 & {\cellcolor[HTML]{D7D7F4}} \color[HTML]{000000} 4.93 & {\cellcolor[HTML]{8D8DF8}} \color[HTML]{F1F1F1} 33.50 & {\cellcolor[HTML]{BABAF6}} \color[HTML]{000000} 2.87 & {\cellcolor[HTML]{8B8BF8}} \color[HTML]{F1F1F1} 12.25 & {\cellcolor[HTML]{CACAF5}} \color[HTML]{000000} 3.09 & {\cellcolor[HTML]{5959FA}} \color[HTML]{F1F1F1} 81.79 \\
gender & daughter & {\cellcolor[HTML]{D8D8F4}} \color[HTML]{000000} 8.02 & {\cellcolor[HTML]{7979F9}} \color[HTML]{F1F1F1} 45.85 & {\cellcolor[HTML]{AEAEF6}} \color[HTML]{000000} 20.38 & {\cellcolor[HTML]{C1C1F5}} \color[HTML]{000000} 19.40 & {\cellcolor[HTML]{D4D4F4}} \color[HTML]{000000} 3.58 & {\cellcolor[HTML]{BEBEF5}} \color[HTML]{000000} 2.76 & {\cellcolor[HTML]{6161FA}} \color[HTML]{F1F1F1} 62.11 & {\cellcolor[HTML]{BDBDF5}} \color[HTML]{000000} 2.55 & {\cellcolor[HTML]{D3D3F4}} \color[HTML]{000000} 5.64 & {\cellcolor[HTML]{9999F7}} \color[HTML]{F1F1F1} 29.70 & {\cellcolor[HTML]{BEBEF5}} \color[HTML]{000000} 2.66 & {\cellcolor[HTML]{7B7BF9}} \color[HTML]{F1F1F1} 13.82 & {\cellcolor[HTML]{8F8FF8}} \color[HTML]{F1F1F1} 7.75 & {\cellcolor[HTML]{7C7CF9}} \color[HTML]{F1F1F1} 75.77 \\
gender & son & {\cellcolor[HTML]{DADAF4}} \color[HTML]{000000} 7.26 & {\cellcolor[HTML]{7878F9}} \color[HTML]{F1F1F1} 46.07 & {\cellcolor[HTML]{C0C0F5}} \color[HTML]{000000} 15.01 & {\cellcolor[HTML]{B7B7F6}} \color[HTML]{000000} 23.85 & {\cellcolor[HTML]{CDCDF5}} \color[HTML]{000000} 4.44 & {\cellcolor[HTML]{B4B4F6}} \color[HTML]{000000} 3.36 & {\cellcolor[HTML]{5B5BFA}} \color[HTML]{F1F1F1} 63.69 & {\cellcolor[HTML]{A9A9F6}} \color[HTML]{000000} 3.52 & {\cellcolor[HTML]{CECEF4}} \color[HTML]{000000} 6.50 & {\cellcolor[HTML]{A4A4F7}} \color[HTML]{000000} 26.29 & {\cellcolor[HTML]{B5B5F6}} \color[HTML]{000000} 3.14 & {\cellcolor[HTML]{6868FA}} \color[HTML]{F1F1F1} 15.66 & {\cellcolor[HTML]{C7C7F5}} \color[HTML]{000000} 3.36 & {\cellcolor[HTML]{6F6FF9}} \color[HTML]{F1F1F1} 77.83 \\
\midrule
religion & Atheist & {\cellcolor[HTML]{F0F0F3}} \color[HTML]{000000} 0.00 & {\cellcolor[HTML]{C0C0F5}} \color[HTML]{000000} 18.52 & {\cellcolor[HTML]{2626FD}} \color[HTML]{F1F1F1} 62.22 & {\cellcolor[HTML]{DDDDF4}} \color[HTML]{000000} 8.15 & {\cellcolor[HTML]{F0F0F3}} \color[HTML]{000000} 0.00 & {\cellcolor[HTML]{2828FD}} \color[HTML]{F1F1F1} 11.11 & {\cellcolor[HTML]{E4E4F3}} \color[HTML]{000000} 22.96 & {\cellcolor[HTML]{C4C4F5}} \color[HTML]{000000} 2.22 & {\cellcolor[HTML]{E1E1F4}} \color[HTML]{000000} 2.96 & {\cellcolor[HTML]{1313FE}} \color[HTML]{F1F1F1} 71.85 & {\cellcolor[HTML]{F0F0F3}} \color[HTML]{000000} 0.00 & {\cellcolor[HTML]{D2D2F4}} \color[HTML]{000000} 5.19 & {\cellcolor[HTML]{D5D5F4}} \color[HTML]{000000} 2.22 & {\cellcolor[HTML]{1919FE}} \color[HTML]{F1F1F1} 92.59 \\
religion & Buddhist & {\cellcolor[HTML]{F0F0F3}} \color[HTML]{000000} 0.00 & {\cellcolor[HTML]{A5A5F7}} \color[HTML]{000000} 28.89 & {\cellcolor[HTML]{C8C8F5}} \color[HTML]{000000} 12.59 & {\cellcolor[HTML]{6B6BFA}} \color[HTML]{F1F1F1} 54.81 & {\cellcolor[HTML]{D3D3F4}} \color[HTML]{000000} 3.70 & {\cellcolor[HTML]{F0F0F3}} \color[HTML]{000000} 0.00 & {\cellcolor[HTML]{3434FC}} \color[HTML]{F1F1F1} 75.56 & {\cellcolor[HTML]{F0F0F3}} \color[HTML]{000000} 0.00 & {\cellcolor[HTML]{E9E9F3}} \color[HTML]{000000} 1.48 & {\cellcolor[HTML]{AEAEF6}} \color[HTML]{000000} 22.96 & {\cellcolor[HTML]{C6C6F5}} \color[HTML]{000000} 2.22 & {\cellcolor[HTML]{E2E2F3}} \color[HTML]{000000} 3.70 & {\cellcolor[HTML]{DEDFF4}} \color[HTML]{000000} 1.48 & {\cellcolor[HTML]{1919FE}} \color[HTML]{F1F1F1} 92.59 \\
religion & Christian & {\cellcolor[HTML]{DADAF4}} \color[HTML]{000000} 7.37 & {\cellcolor[HTML]{7373F9}} \color[HTML]{F1F1F1} 47.78 & {\cellcolor[HTML]{B4B4F6}} \color[HTML]{000000} 18.71 & {\cellcolor[HTML]{C1C1F5}} \color[HTML]{000000} 19.77 & {\cellcolor[HTML]{D1D1F4}} \color[HTML]{000000} 3.92 & {\cellcolor[HTML]{C4C4F5}} \color[HTML]{000000} 2.46 & {\cellcolor[HTML]{6565FA}} \color[HTML]{F1F1F1} 61.05 & {\cellcolor[HTML]{AAAAF6}} \color[HTML]{000000} 3.51 & {\cellcolor[HTML]{CFCFF4}} \color[HTML]{000000} 6.32 & {\cellcolor[HTML]{9B9BF7}} \color[HTML]{F1F1F1} 29.12 & {\cellcolor[HTML]{B8B8F6}} \color[HTML]{000000} 2.98 & {\cellcolor[HTML]{7373F9}} \color[HTML]{F1F1F1} 14.62 & {\cellcolor[HTML]{BCBCF5}} \color[HTML]{000000} 4.27 & {\cellcolor[HTML]{6D6DF9}} \color[HTML]{F1F1F1} 78.13 \\
religion & Hindu & {\cellcolor[HTML]{D7D7F4}} \color[HTML]{000000} 8.25 & {\cellcolor[HTML]{7979F9}} \color[HTML]{F1F1F1} 45.67 & {\cellcolor[HTML]{BDBDF5}} \color[HTML]{000000} 16.02 & {\cellcolor[HTML]{BBBBF5}} \color[HTML]{000000} 22.46 & {\cellcolor[HTML]{CECEF5}} \color[HTML]{000000} 4.39 & {\cellcolor[HTML]{B7B7F6}} \color[HTML]{000000} 3.22 & {\cellcolor[HTML]{6363FA}} \color[HTML]{F1F1F1} 61.52 & {\cellcolor[HTML]{B0B0F6}} \color[HTML]{000000} 3.16 & {\cellcolor[HTML]{D2D2F4}} \color[HTML]{000000} 5.85 & {\cellcolor[HTML]{9A9AF7}} \color[HTML]{F1F1F1} 29.47 & {\cellcolor[HTML]{BDBDF5}} \color[HTML]{000000} 2.75 & {\cellcolor[HTML]{6969FA}} \color[HTML]{F1F1F1} 15.56 & {\cellcolor[HTML]{9E9EF7}} \color[HTML]{F1F1F1} 6.55 & {\cellcolor[HTML]{7F7FF9}} \color[HTML]{F1F1F1} 75.15 \\
religion & Jew & {\cellcolor[HTML]{EFEFF3}} \color[HTML]{000000} 0.74 & {\cellcolor[HTML]{5757FB}} \color[HTML]{F1F1F1} 58.52 & {\cellcolor[HTML]{E5E5F3}} \color[HTML]{000000} 3.70 & {\cellcolor[HTML]{B7B7F6}} \color[HTML]{000000} 23.70 & {\cellcolor[HTML]{F0F0F3}} \color[HTML]{000000} 0.00 & {\cellcolor[HTML]{0000FF}} \color[HTML]{F1F1F1} 13.33 & {\cellcolor[HTML]{0000FF}} \color[HTML]{F1F1F1} 91.11 & {\cellcolor[HTML]{9797F7}} \color[HTML]{F1F1F1} 4.44 & {\cellcolor[HTML]{E5E5F3}} \color[HTML]{000000} 2.22 & {\cellcolor[HTML]{F0F0F3}} \color[HTML]{000000} 2.22 & {\cellcolor[HTML]{8D8DF8}} \color[HTML]{F1F1F1} 5.19 & {\cellcolor[HTML]{0707FF}} \color[HTML]{F1F1F1} 25.19 & {\cellcolor[HTML]{E8E8F3}} \color[HTML]{000000} 0.74 & {\cellcolor[HTML]{A3A3F7}} \color[HTML]{F1F1F1} 68.89 \\
religion & Muslim & {\cellcolor[HTML]{D4D4F4}} \color[HTML]{000000} 9.06 & {\cellcolor[HTML]{7676F9}} \color[HTML]{F1F1F1} 46.73 & {\cellcolor[HTML]{BBBBF5}} \color[HTML]{000000} 16.55 & {\cellcolor[HTML]{BEBEF5}} \color[HTML]{000000} 21.11 & {\cellcolor[HTML]{CFCFF4}} \color[HTML]{000000} 4.15 & {\cellcolor[HTML]{C5C5F5}} \color[HTML]{000000} 2.40 & {\cellcolor[HTML]{6565FA}} \color[HTML]{F1F1F1} 60.94 & {\cellcolor[HTML]{B3B3F6}} \color[HTML]{000000} 3.04 & {\cellcolor[HTML]{D3D3F4}} \color[HTML]{000000} 5.73 & {\cellcolor[HTML]{9797F7}} \color[HTML]{F1F1F1} 30.29 & {\cellcolor[HTML]{B7B7F6}} \color[HTML]{000000} 3.04 & {\cellcolor[HTML]{8C8CF8}} \color[HTML]{F1F1F1} 12.16 & {\cellcolor[HTML]{BDBDF5}} \color[HTML]{000000} 4.15 & {\cellcolor[HTML]{5F5FFA}} \color[HTML]{F1F1F1} 80.64 \\
\midrule
 role & father & {\cellcolor[HTML]{D7D7F4}} \color[HTML]{000000} 8.25 & {\cellcolor[HTML]{7979F9}} \color[HTML]{F1F1F1} 45.67 & {\cellcolor[HTML]{BDBDF5}} \color[HTML]{000000} 16.02 & {\cellcolor[HTML]{BBBBF5}} \color[HTML]{000000} 22.46 & {\cellcolor[HTML]{CECEF5}} \color[HTML]{000000} 4.39 & {\cellcolor[HTML]{B7B7F6}} \color[HTML]{000000} 3.22 & {\cellcolor[HTML]{6363FA}} \color[HTML]{F1F1F1} 61.52 & {\cellcolor[HTML]{B0B0F6}} \color[HTML]{000000} 3.16 & {\cellcolor[HTML]{D2D2F4}} \color[HTML]{000000} 5.85 & {\cellcolor[HTML]{9A9AF7}} \color[HTML]{F1F1F1} 29.47 & {\cellcolor[HTML]{BDBDF5}} \color[HTML]{000000} 2.75 & {\cellcolor[HTML]{6969FA}} \color[HTML]{F1F1F1} 15.56 & {\cellcolor[HTML]{9E9EF7}} \color[HTML]{F1F1F1} 6.55 & {\cellcolor[HTML]{7F7FF9}} \color[HTML]{F1F1F1} 75.15 \\
role & mother & {\cellcolor[HTML]{D4D4F4}} \color[HTML]{000000} 9.06 & {\cellcolor[HTML]{7676F9}} \color[HTML]{F1F1F1} 46.73 & {\cellcolor[HTML]{BBBBF5}} \color[HTML]{000000} 16.55 & {\cellcolor[HTML]{BEBEF5}} \color[HTML]{000000} 21.11 & {\cellcolor[HTML]{CFCFF4}} \color[HTML]{000000} 4.15 & {\cellcolor[HTML]{C5C5F5}} \color[HTML]{000000} 2.40 & {\cellcolor[HTML]{6565FA}} \color[HTML]{F1F1F1} 60.94 & {\cellcolor[HTML]{B3B3F6}} \color[HTML]{000000} 3.04 & {\cellcolor[HTML]{D3D3F4}} \color[HTML]{000000} 5.73 & {\cellcolor[HTML]{9797F7}} \color[HTML]{F1F1F1} 30.29 & {\cellcolor[HTML]{B7B7F6}} \color[HTML]{000000} 3.04 & {\cellcolor[HTML]{8C8CF8}} \color[HTML]{F1F1F1} 12.16 & {\cellcolor[HTML]{BDBDF5}} \color[HTML]{000000} 4.15 & {\cellcolor[HTML]{5F5FFA}} \color[HTML]{F1F1F1} 80.64 \\
 role & parent & {\cellcolor[HTML]{DADAF4}} \color[HTML]{000000} 7.37 & {\cellcolor[HTML]{7373F9}} \color[HTML]{F1F1F1} 47.78 & {\cellcolor[HTML]{B4B4F6}} \color[HTML]{000000} 18.71 & {\cellcolor[HTML]{C1C1F5}} \color[HTML]{000000} 19.77 & {\cellcolor[HTML]{D1D1F4}} \color[HTML]{000000} 3.92 & {\cellcolor[HTML]{C4C4F5}} \color[HTML]{000000} 2.46 & {\cellcolor[HTML]{6565FA}} \color[HTML]{F1F1F1} 61.05 & {\cellcolor[HTML]{AAAAF6}} \color[HTML]{000000} 3.51 & {\cellcolor[HTML]{CFCFF4}} \color[HTML]{000000} 6.32 & {\cellcolor[HTML]{9B9BF7}} \color[HTML]{F1F1F1} 29.12 & {\cellcolor[HTML]{B8B8F6}} \color[HTML]{000000} 2.98 & {\cellcolor[HTML]{7373F9}} \color[HTML]{F1F1F1} 14.62 & {\cellcolor[HTML]{BCBCF5}} \color[HTML]{000000} 4.27 & {\cellcolor[HTML]{6D6DF9}} \color[HTML]{F1F1F1} 78.13 \\

\bottomrule

\end{tabular}

}
\caption{The percentage of context-centric attributes for each sociocultural factors\\ (Desert: \emoji{desert}, Green Bodies: \emoji{evergreen-tree}, Magical: \emoji{magic-wand}, Mountain: \emoji{snow-capped-mountain}, Water Bodies: \emoji{water-wave}, City: \emoji{cityscape}, Town:\emoji{circus-tent}, Village:\emoji{house-with-garden}, Poor: \emoji{money-bag}, Middle-class:\emoji{money-bag}\emoji{money-bag}, Wealthy:\emoji{money-bag}\emoji{money-bag}\emoji{money-bag}, Not found=\emoji{cross-mark}.)}
\label{tab:context-percentage}
\end{table*}

\subsection{Measuring Bias through Predictability}
To further probe for implicit biases present in the text, we predict the target variable from the text instead of focusing on individual word-level features. 
To this end, we vectorize the text with TF-IDF and fit a feed forward neural network with 5-fold cross-validation.
Again, we remove any clear indicators from the text that refer to the child, i.e., \textit{girl}, \textit{boy} and \textit{child}.
\Cref{tab:predictability} summarizes the accuracy of predicting each target variable under different conditions. The \textbf{Majority} baseline always predics the most frequent class.
\textbf{Average} is the accuracy when using combined data across all models.
We also demonstrate the prediction accuracies when the text is generated exclusively by specific language models.
Notably, targets like economy (developed vs. developing) countries and nationality show high predictability (around 89-90\%), suggesting that the narratives carry robust implicit signals for these dimensions.
In contrast, the lower performance for role and religion might reflect subtler biases or less overt textual cues in those domains. Moreover, the variations in prediction accuracy across different models indicate that the nature and strength of embedded biases differ based on the model that produced the text.

\section{Related Work}
\label{sec:related}
Recent studies have analyzed the outputs of LLMs (e.g., \citet{ lucy-bamman-2021-gender}), revealing that these models often amplify existing societal biases when generating text. \citet{Arzaghi_Carichon_Farnadi_2024} studies the impact of gender, race, and marital status on socioeconomic biases associated with LLMs. All of these studies have focused only on adult-centric applications. Still, they provide essential insights and methodological tools for examining biases in any generated content.


\paragraph{Gender and Cultural Bias in Children's Narratives.}
Children’s literature is critical in shaping early perceptions of identity, morality, and culture \cite{10.1145/3687035, ye2024storypark}. However, research specifically related to the children's narratives remains limited. Lexical complexity must align with children's reading abilities to make LLM-generated content suitable for children. \newcite{rooein2023know} demonstrated that these LLMs struggle to adapt to specific age and grade levels. Various studies have explored approaches to address this limitation, such as lexical simplification models  \citep{valentini-etal-2023-automatic} and prompt-based techniques~\citep{rooein-etal-2024-beyond} to tailor content for children. Some studies  \cite{pownall2023mr, bhandari-brennan-2023-trustworthiness,nayeem-rafiei-2024-kidlm} have begun to investigate gender bias in children’s texts, noting that stereotypical portrayals may reinforce traditional roles and limit diverse representations. Additionally, \newcite{toro-isaza-etal-2023-fairy} proposed computational pipelines to extract narrative structures, revealing biases about protagonists. None of these studies investigate the relationship between sociocultural factors and how they are presented as a vocabulary of respected biases.

\paragraph{Personalized Story Generation.}
\citet {10.1145/3613904.3642647} showed the effectiveness of personalized content for learning outcomes. Frameworks such as MirrorStories \cite{yunusov-etal-2024-mirrorstories} showed the potential for incorporating sociocultural elements (e.g., gender, ethnicity, interests) into engaging storytelling. Despite these advances, challenges remain: personalization efforts must balance engagement with fairness, ensuring that the generated content does not inadvertently reinforce harmful stereotypes. Our work extends this line of inquiry by exploring how explicit sociocultural prompts influence the narrative output, particularly for children. Recognizing the risks associated with biased outputs, recent studies have also focused on methods to measure and reduce these biases. Additionally, research by \citet{bhatt-diaz-2024-extrinsic} has explored the notion of cultural competence in LLMs. Our work contributes to this area by providing an evaluation framework through the Biased Tales corpus that quantifies both narrative-level and attribute-level biases for children's stories, offering actionable insights for improving the cultural sensitivity of story generation systems.




\section{Conclusion}

We looked into the presence and impact of sociocultural biases in LLM-generated children's stories. We present the Biased Tales dataset, which includes over 5,500 personalized stories incorporating various sociocultural factors, i.e. gender, nationality, ethnicity, religion, and parental role. We also present a comprehensive evaluation framework for determining how LLMs incorporate biases into narrative outputs.

First, LLMs tailor their narrative content to explicit sociocultural prompts. We separated this adaptation into character-centric and context-centric components.
Character-centric analysis reveals a 55.26\% increase in appearance-related descriptors in stories written for girls compared to boys. 
The context-centric analysis reveals a trend in the geolocation of the stories, which occur frequently in green bodies, with a village urban, with no mention of the socioeconomic aspect. Our findings demonstrate that, while personalization can increase engagement, it can also limit the diversity of representation in children's literature.



\section*{Limitations}
The Biased Tales dataset focuses only on English stories and a limited number of sociocultural factors. Future research should aim to extend the range of languages and cultural contexts examined, incorporate more nuanced intersectional analyses, and validate the impact of these biases through user studies with target audiences. While our analysis focuses on the attributes of protagonists in these narratives, it is important to note that stories often feature multiple characters. Future studies should investigate representations of these diverse characters as a medium for measuring bias, as their portrayals may collectively shape cultural or societal perceptions. Additionally, further refinement of model training procedures and bias mitigation techniques is necessary to reduce the propagation of cultural stereotypes in generated narratives.

\section*{Ethical Considerations}

In our study, we do not use data from actual people but evaluate a sample of AI-generated stories with two human annotators. Biased Tales might unintentionally amplify or normalize specific biased patterns (e.g., gendered or cultural stereotypes) if not contextualized appropriately for readers.

\section*{Acknowledgment}
Donya Rooein and Dirk Hovy's research is supported through the European Research Council (ERC) under the European Union’s Horizon 2020 research and innovation program (No. 949944, INTEGRATOR). Debora Nozza's research is from the ERC under the European Union’s Horizon 2020 research and innovation program (grant agreement No. 101116095, PERSONAE).

\bibliography{misc/bibliography,misc/anthology.min}


\appendix
\section{Sociocultural Variables}
We present the sociocultural variables in our experiment in \Cref{tab:nationality_ethnicity}. This table catalogs key demographic details of our participant sample, including nationality, ethnicity, religion, gender, and parental roles. 

\begin{table*}[t]
\small
\centering
\begin{tabular}{lp{10cm}}
\toprule
\textbf{Sociocultural Factor} & \textbf{Values} \\
\midrule
Gender & daughter, son, child. \\
Role & mother, father, parent.\\
Nationality       & American, British, Italian, German, Russian, Armenian, Mexican, Brazilian, Iraqi, Egyptian, Iranian, Afghan, Sudanese, Malian, Kenyan, Nigerian, South African, Ethiopian, Indian, Sri Lankan, Tajik, Azerbaijani, Chinese, Japanese, Vietnamese, Thai, Indonesian, Filipino. \\ \\[-0.5em]
Religion           & Atheist, Buddhist, Christian, Hindu, Jewish, Muslim. \\[0.5em]
Ethnicity         & Asian, African-American, European-American, Latino, Middle-Eastern, White. \\ \bottomrule
\end{tabular}
\caption{List of variables for nationalities, religion, and ethnicities}
\label{tab:nationality_ethnicity}
\end{table*}


\section{Story Annotation}
\label{appx:story_annot}
We use human annotations on a sub-sample of 1000 and GPT-4o for the entire dataset to extract character- and context-centric attributes.

\paragraph{Profile and the pay rate.} Data has been annotated by one native Italian speaker and one native Romanian, age group 21-28, both fluent in English at the C1 level with experience in computational linguistics. The data we share is not sensitive to personal information, as it does not contain information about individuals. Our data does not contain hurtful messages that can be used in hurtful ways. They were each compensated for 32 hours of work at a rate of about 18 euros per hour.

\paragraph{Guideline for human annotators.} We described the step-by-step instructions for extracting attributes in the Biased Tales repository\footnote{\includegraphics[height=1em]{git-logo.png} \href{https://github.com/donya-rooein/biased_tales}{github.com/donya-rooein/biased\_tales}}.

\paragraph{Prompt list for GPT-4o annotation.} Table~\ref{table:prompts_story_analysis} presents the specific prompts used in our analysis. Each prompt was designed to elicit comprehensive responses from the model, ensuring that both explicit details and subtle narrative cues were captured.

\begin{table*}[htbp]
\centering
\small
\begin{tabular}{p{5cm}p{10cm}}
\toprule
\textbf{Attribute} & \textbf{Prompt} \\
\midrule
protagonist\_name & Given a story\{story\_text\}, what is the name of the protagonist in this story? Only answer with the name and omit explanations.  \\
\cmidrule{1-1}
protagonist\_characters & Given a story\{story\_text\}, what are the physical characters of the protagonist in this story. Only answer with a list of attributes and omit explanations.  \\
\cmidrule{1-1}
protagonist\_attributes & Given a story\{story\_text\}, what are the attributes of the protagonist in this story and separate them with ‘, ’. Only answer with a list of attributes and omit explanations. Use only attributes in the text and do not include new attributes. \\
\cmidrule{1-1}
story\_setting\_geographical & Given a story\{story\_text\}, what is the geographical setting of the following story? Only answer with one of the following options: Magical/Imaginary Places, Water Bodies, Desert, Mountain, Green Bodies, None.\\ 
\cmidrule{1-1}
story\_setting\_urban & Given a story\{story\_text\}, what is the urban setting of the following story? Only answer with one of the following options: City, Town, Village, None.  \\
\cmidrule{1-1}
story\_setting\_social & Given a story\{story\_text\}, what is the socioeconomic status presented in the following story? Only answer with one of the following options: Poor, Middle-class, Wealthy, None.  \\
\bottomrule
\end{tabular}
\caption{Prompts for collecting character- and context-centric attributes for story analysis.}
\label{table:prompts_story_analysis}
\end{table*}

\section{Story Analysis}
\label{appx:protag_analysis}

\Cref{tab:corr_bias_text_gpt4}, \Cref{tab:corr_bias_text_llama3}, and \Cref{tab:corr_bias_text_mixtral} presents how across different models the top words in the \textbf{text of the generated story} are presented. 

\begin{table*}[htbp]
\fontsize{8}{8}\selectfont
\begin{tabular}{p{2cm}l}
\toprule \\[-0.5em]
\null\\[-1.7em]
\multicolumn{2}{l}{\bf Gender} \\

child & \makebox[22mm][l]{ {\fontsize{7}{7}\selectfont 12\%} twinkling } \makebox[22mm][l]{ {\fontsize{7}{7}\selectfont 12\%} squirrel } \makebox[22mm][l]{ {\fontsize{7}{7}\selectfont 9\%} watchful } \makebox[22mm][l]{ {\fontsize{7}{7}\selectfont 9\%} owl } \makebox[22mm][l]{ {\fontsize{7}{7}\selectfont 8\%} decided } { {\fontsize{7}{7}\selectfont 8\%} branch } \\
daughter & \makebox[22mm][l]{ {\fontsize{7}{7}\selectfont 27\%} princess } \makebox[22mm][l]{ {\fontsize{7}{7}\selectfont 25\%} Lily } \makebox[22mm][l]{ {\fontsize{7}{7}\selectfont 21\%} Layla } \makebox[22mm][l]{ {\fontsize{7}{7}\selectfont 16\%} Amina } \makebox[22mm][l]{ {\fontsize{7}{7}\selectfont 15\%} window } { {\fontsize{7}{7}\selectfont 14\%} Mei } \\
son & \makebox[22mm][l]{ {\fontsize{7}{7}\selectfont 15\%} Amir } \makebox[22mm][l]{ {\fontsize{7}{7}\selectfont 9\%} set } \makebox[22mm][l]{ {\fontsize{7}{7}\selectfont 8\%} Ali } \makebox[22mm][l]{ {\fontsize{7}{7}\selectfont 8\%} majestic } \makebox[22mm][l]{ {\fontsize{7}{7}\selectfont 8\%} sat } { {\fontsize{7}{7}\selectfont 7\%} said } \\
\bottomrule\\

\null\\[-1.7em]
\multicolumn{2}{l}{\bf Nationality Parent Group} \\

Africa & \makebox[22mm][l]{ {\fontsize{7}{7}\selectfont 34\%} simba } \makebox[22mm][l]{ {\fontsize{7}{7}\selectfont 24\%} kofi } \makebox[22mm][l]{ {\fontsize{7}{7}\selectfont 24\%} elephant } \makebox[22mm][l]{ {\fontsize{7}{7}\selectfont 12\%} monkey } \makebox[22mm][l]{ {\fontsize{7}{7}\selectfont 12\%} majestic } { {\fontsize{7}{7}\selectfont 11\%} amina } \\
Asia & \makebox[22mm][l]{ {\fontsize{7}{7}\selectfont 22\%} Linh } \makebox[22mm][l]{ {\fontsize{7}{7}\selectfont 21\%} elephant } \makebox[22mm][l]{ {\fontsize{7}{7}\selectfont 17\%} monkey } \makebox[22mm][l]{ {\fontsize{7}{7}\selectfont 14\%} Maria } \makebox[22mm][l]{ {\fontsize{7}{7}\selectfont 12\%} decided } { {\fontsize{7}{7}\selectfont 12\%} mystical } \\
European & \makebox[22mm][l]{ {\fontsize{7}{7}\selectfont 25\%} nestled } \makebox[22mm][l]{ {\fontsize{7}{7}\selectfont 20\%} Lily } \makebox[22mm][l]{ {\fontsize{7}{7}\selectfont 14\%} Oliver } \makebox[22mm][l]{ {\fontsize{7}{7}\selectfont 13\%} Sofia } \makebox[22mm][l]{ {\fontsize{7}{7}\selectfont 11\%} window } { {\fontsize{7}{7}\selectfont 10\%} princess } \\
Middle Eastern & \makebox[22mm][l]{ {\fontsize{7}{7}\selectfont 31\%} Ali } \makebox[22mm][l]{ {\fontsize{7}{7}\selectfont 25\%} Leyla } \makebox[22mm][l]{ {\fontsize{7}{7}\selectfont 23\%} layla } \makebox[22mm][l]{ {\fontsize{7}{7}\selectfont 17\%} Amir } \makebox[22mm][l]{ {\fontsize{7}{7}\selectfont 11\%} tell } { {\fontsize{7}{7}\selectfont 9\%} majestic } \\
North American & \makebox[22mm][l]{ {\fontsize{7}{7}\selectfont 15\%} squirrel } \makebox[22mm][l]{ {\fontsize{7}{7}\selectfont 14\%} lily } \makebox[22mm][l]{ {\fontsize{7}{7}\selectfont 10\%} away } \makebox[22mm][l]{ {\fontsize{7}{7}\selectfont 10\%} owl } \makebox[22mm][l]{ {\fontsize{7}{7}\selectfont 10\%} nestled } { {\fontsize{7}{7}\selectfont 8\%} far } \\
South American & \makebox[22mm][l]{ {\fontsize{7}{7}\selectfont 24\%} colorful } \makebox[22mm][l]{ {\fontsize{7}{7}\selectfont 20\%} monkey } \makebox[22mm][l]{ {\fontsize{7}{7}\selectfont 18\%} Maria } \makebox[22mm][l]{ {\fontsize{7}{7}\selectfont 17\%} isabella } \makebox[22mm][l]{ {\fontsize{7}{7}\selectfont 14\%} Mateo } { {\fontsize{7}{7}\selectfont 11\%} Luna } \\
\bottomrule\\

\null\\[-1.7em]
\multicolumn{2}{l}{\bf Nationality Parent Developed} \\

Developed & \makebox[22mm][l]{ {\fontsize{7}{7}\selectfont 28\%} nestled } \makebox[22mm][l]{ {\fontsize{7}{7}\selectfont 27\%} Lily } \makebox[22mm][l]{ {\fontsize{7}{7}\selectfont 20\%} fox } \makebox[22mm][l]{ {\fontsize{7}{7}\selectfont 17\%} Oliver } \makebox[22mm][l]{ {\fontsize{7}{7}\selectfont 16\%} squirrel } { {\fontsize{7}{7}\selectfont 12\%} owl } \\
Developing & \makebox[22mm][l]{ {\fontsize{7}{7}\selectfont 25\%} monkey } \makebox[22mm][l]{ {\fontsize{7}{7}\selectfont 22\%} elephant } \makebox[22mm][l]{ {\fontsize{7}{7}\selectfont 19\%} majestic } \makebox[22mm][l]{ {\fontsize{7}{7}\selectfont 13\%} simba } \makebox[22mm][l]{ {\fontsize{7}{7}\selectfont 13\%} tell } { {\fontsize{7}{7}\selectfont 13\%} colorful } \\
\bottomrule\\

\null\\[-1.7em]
\multicolumn{2}{l}{\bf Ethnicity} \\

African-Amer. & \makebox[22mm][l]{ {\fontsize{7}{7}\selectfont 60\%} kofi } \makebox[22mm][l]{ {\fontsize{7}{7}\selectfont 39\%} malik } \makebox[22mm][l]{ {\fontsize{7}{7}\selectfont 32\%} vibrant } \makebox[22mm][l]{ {\fontsize{7}{7}\selectfont 23\%} remember } \makebox[22mm][l]{ {\fontsize{7}{7}\selectfont 18\%} faced } { {\fontsize{7}{7}\selectfont 17\%} lay } \\
Asian & \makebox[22mm][l]{ {\fontsize{7}{7}\selectfont 65\%} Mei } \makebox[22mm][l]{ {\fontsize{7}{7}\selectfont 36\%} Kai } \makebox[22mm][l]{ {\fontsize{7}{7}\selectfont 36\%} Lin } \makebox[22mm][l]{ {\fontsize{7}{7}\selectfont 36\%} Ling } \makebox[22mm][l]{ {\fontsize{7}{7}\selectfont 31\%} ming } { {\fontsize{7}{7}\selectfont 31\%} li } \\
European-Amer. & \makebox[22mm][l]{ {\fontsize{7}{7}\selectfont 35\%} nestled } \makebox[22mm][l]{ {\fontsize{7}{7}\selectfont 31\%} owl } \makebox[22mm][l]{ {\fontsize{7}{7}\selectfont 31\%} Oliver } \makebox[22mm][l]{ {\fontsize{7}{7}\selectfont 27\%} Lily } \makebox[22mm][l]{ {\fontsize{7}{7}\selectfont 25\%} Sammy } { {\fontsize{7}{7}\selectfont 23\%} stumbled } \\
Latino & \makebox[22mm][l]{ {\fontsize{7}{7}\selectfont 62\%} Mateo } \makebox[22mm][l]{ {\fontsize{7}{7}\selectfont 46\%} abuela } \makebox[22mm][l]{ {\fontsize{7}{7}\selectfont 45\%} colorful } \makebox[22mm][l]{ {\fontsize{7}{7}\selectfont 43\%} Isabella } \makebox[22mm][l]{ {\fontsize{7}{7}\selectfont 34\%} Sofia } { {\fontsize{7}{7}\selectfont 33\%} nestled } \\
Middle-Eastern & \makebox[22mm][l]{ {\fontsize{7}{7}\selectfont 68\%} Amir } \makebox[22mm][l]{ {\fontsize{7}{7}\selectfont 58\%} Layla } \makebox[22mm][l]{ {\fontsize{7}{7}\selectfont 58\%} oasis } \makebox[22mm][l]{ {\fontsize{7}{7}\selectfont 36\%} Ali } \makebox[22mm][l]{ {\fontsize{7}{7}\selectfont 36\%} prince } { {\fontsize{7}{7}\selectfont 19\%} dipped } \\
White & \makebox[22mm][l]{ {\fontsize{7}{7}\selectfont 41\%} Lily } \makebox[22mm][l]{ {\fontsize{7}{7}\selectfont 20\%} set } \makebox[22mm][l]{ {\fontsize{7}{7}\selectfont 19\%} decided } \makebox[22mm][l]{ {\fontsize{7}{7}\selectfont 18\%} end } \makebox[22mm][l]{ {\fontsize{7}{7}\selectfont 15\%} away } { {\fontsize{7}{7}\selectfont 14\%} returned } \\
\bottomrule\\

\null\\[-1.7em]
\multicolumn{2}{l}{\bf Religion} \\

Atheist & \makebox[22mm][l]{ {\fontsize{7}{7}\selectfont 15\%} Luna } \makebox[22mm][l]{ {\fontsize{7}{7}\selectfont 12\%} Lily } \makebox[22mm][l]{ {\fontsize{7}{7}\selectfont 11\%} smiled } \makebox[22mm][l]{ {\fontsize{7}{7}\selectfont 7\%} away } \makebox[22mm][l]{ {\fontsize{7}{7}\selectfont 7\%} matter } { {\fontsize{7}{7}\selectfont 5\%} squirrel } \\
Buddhist & \makebox[22mm][l]{ {\fontsize{7}{7}\selectfont 42\%} Kavi } \makebox[22mm][l]{ {\fontsize{7}{7}\selectfont 28\%} Mei } \makebox[22mm][l]{ {\fontsize{7}{7}\selectfont 18\%} nestled } \makebox[22mm][l]{ {\fontsize{7}{7}\selectfont 13\%} sat } \makebox[22mm][l]{ {\fontsize{7}{7}\selectfont 12\%} smiled } { {\fontsize{7}{7}\selectfont 10\%} Li } \\
Christian & \makebox[22mm][l]{ {\fontsize{7}{7}\selectfont 10\%} decided } \makebox[22mm][l]{ {\fontsize{7}{7}\selectfont 10\%} shimmering } \makebox[22mm][l]{ {\fontsize{7}{7}\selectfont 10\%} twinkling } \makebox[22mm][l]{ {\fontsize{7}{7}\selectfont 7\%} Lily } \makebox[22mm][l]{ {\fontsize{7}{7}\selectfont 7\%} tiny } { {\fontsize{7}{7}\selectfont 7\%} saw } \\
Hindu & \makebox[22mm][l]{ {\fontsize{7}{7}\selectfont 9\%} princess } \makebox[22mm][l]{ {\fontsize{7}{7}\selectfont 5\%} tell } \makebox[22mm][l]{ {\fontsize{7}{7}\selectfont 5\%} reached } \makebox[22mm][l]{ {\fontsize{7}{7}\selectfont 5\%} mystical } \makebox[22mm][l]{ {\fontsize{7}{7}\selectfont 5\%} elephant } { {\fontsize{7}{7}\selectfont 5\%} finally } \\
Jew & \makebox[22mm][l]{ {\fontsize{7}{7}\selectfont 15\%} nestled } \makebox[22mm][l]{ {\fontsize{7}{7}\selectfont 14\%} small } \makebox[22mm][l]{ {\fontsize{7}{7}\selectfont 11\%} guide } \makebox[22mm][l]{ {\fontsize{7}{7}\selectfont 7\%} known } \makebox[22mm][l]{ {\fontsize{7}{7}\selectfont 6\%} tell } { {\fontsize{7}{7}\selectfont 6\%} window } \\
Muslim & \makebox[22mm][l]{ {\fontsize{7}{7}\selectfont 13\%} Amina } \makebox[22mm][l]{ {\fontsize{7}{7}\selectfont 11\%} whispered } \makebox[22mm][l]{ {\fontsize{7}{7}\selectfont 9\%} tucked } \makebox[22mm][l]{ {\fontsize{7}{7}\selectfont 9\%} called } \makebox[22mm][l]{ {\fontsize{7}{7}\selectfont 9\%} soothing } { {\fontsize{7}{7}\selectfont 8\%} began } \\
\bottomrule\\

\null\\[-1.7em]
\multicolumn{2}{l}{\bf Role} \\

father & \makebox[22mm][l]{ {\fontsize{7}{7}\selectfont 9\%} princess } \makebox[22mm][l]{ {\fontsize{7}{7}\selectfont 5\%} tell } \makebox[22mm][l]{ {\fontsize{7}{7}\selectfont 5\%} reached } \makebox[22mm][l]{ {\fontsize{7}{7}\selectfont 5\%} mystical } \makebox[22mm][l]{ {\fontsize{7}{7}\selectfont 5\%} elephant } { {\fontsize{7}{7}\selectfont 5\%} finally } \\
mother & \makebox[22mm][l]{ {\fontsize{7}{7}\selectfont 13\%} Amina } \makebox[22mm][l]{ {\fontsize{7}{7}\selectfont 11\%} whispered } \makebox[22mm][l]{ {\fontsize{7}{7}\selectfont 9\%} tucked } \makebox[22mm][l]{ {\fontsize{7}{7}\selectfont 9\%} called } \makebox[22mm][l]{ {\fontsize{7}{7}\selectfont 9\%} soothing } { {\fontsize{7}{7}\selectfont 8\%} began } \\
parent & \makebox[22mm][l]{ {\fontsize{7}{7}\selectfont 10\%} decided } \makebox[22mm][l]{ {\fontsize{7}{7}\selectfont 10\%} shimmering } \makebox[22mm][l]{ {\fontsize{7}{7}\selectfont 10\%} twinkling } \makebox[22mm][l]{ {\fontsize{7}{7}\selectfont 7\%} Lily } \makebox[22mm][l]{ {\fontsize{7}{7}\selectfont 7\%} tiny } { {\fontsize{7}{7}\selectfont 7\%} saw } \\
\bottomrule\\

\end{tabular}
\vspace{-6mm}
\caption{Top words in the \textbf{text of the generated story by GPT-4o model} that correlate (Pearson) with the sociocultural factor. The terms \textit{child}, \textit{daughter}, and \textit{son} have been removed, as they are almost present at the start of the generation}
\label{tab:corr_bias_text_gpt4}
\end{table*}

\begin{table*}[htbp]
\fontsize{8}{8}\selectfont
\begin{tabular}{p{2cm}l}
\toprule \\[-0.5em]
\null\\[-1.7em]
\multicolumn{2}{l}{\bf Gender} \\

child & \makebox[22mm][l]{ {\fontsize{7}{7}\selectfont 9\%} tonight } \makebox[22mm][l]{ {\fontsize{7}{7}\selectfont 8\%} high } \makebox[22mm][l]{ {\fontsize{7}{7}\selectfont 5\%} gently } \makebox[22mm][l]{ {\fontsize{7}{7}\selectfont 5\%} welcomed } \makebox[22mm][l]{ {\fontsize{7}{7}\selectfont 5\%} flew } { {\fontsize{7}{7}\selectfont 5\%} sang } \\
daughter & \makebox[22mm][l]{ {\fontsize{7}{7}\selectfont 19\%} dear } \makebox[22mm][l]{ {\fontsize{7}{7}\selectfont 14\%} Lily } \makebox[22mm][l]{ {\fontsize{7}{7}\selectfont 13\%} bloomed } \makebox[22mm][l]{ {\fontsize{7}{7}\selectfont 12\%} Leila } \makebox[22mm][l]{ {\fontsize{7}{7}\selectfont 11\%} surrounded } { {\fontsize{7}{7}\selectfont 9\%} wandered } \\
son & \makebox[22mm][l]{ {\fontsize{7}{7}\selectfont 14\%} grandfather } \makebox[22mm][l]{ {\fontsize{7}{7}\selectfont 11\%} set } \makebox[22mm][l]{ {\fontsize{7}{7}\selectfont 11\%} Max } \makebox[22mm][l]{ {\fontsize{7}{7}\selectfont 10\%} Leo } \makebox[22mm][l]{ {\fontsize{7}{7}\selectfont 10\%} suddenly } { {\fontsize{7}{7}\selectfont 10\%} symbol } \\
\bottomrule\\

\null\\[-1.7em]
\multicolumn{2}{l}{\bf Nationality Parent Group} \\

Africa & \makebox[22mm][l]{ {\fontsize{7}{7}\selectfont 20\%} roamed } \makebox[22mm][l]{ {\fontsize{7}{7}\selectfont 19\%} horizon } \makebox[22mm][l]{ {\fontsize{7}{7}\selectfont 19\%} hue } \makebox[22mm][l]{ {\fontsize{7}{7}\selectfont 12\%} honor } \makebox[22mm][l]{ {\fontsize{7}{7}\selectfont 9\%} trunk } { {\fontsize{7}{7}\selectfont 8\%} grandfather } \\
Asia & \makebox[22mm][l]{ {\fontsize{7}{7}\selectfont 19\%} rohan } \makebox[22mm][l]{ {\fontsize{7}{7}\selectfont 15\%} waterfall } \makebox[22mm][l]{ {\fontsize{7}{7}\selectfont 13\%} lush } \makebox[22mm][l]{ {\fontsize{7}{7}\selectfont 12\%} nestled } \makebox[22mm][l]{ {\fontsize{7}{7}\selectfont 11\%} goodnight } { {\fontsize{7}{7}\selectfont 10\%} bloomed } \\
European & \makebox[22mm][l]{ {\fontsize{7}{7}\selectfont 27\%} rolling } \makebox[22mm][l]{ {\fontsize{7}{7}\selectfont 16\%} Leo } \makebox[22mm][l]{ {\fontsize{7}{7}\selectfont 10\%} nestled } \makebox[22mm][l]{ {\fontsize{7}{7}\selectfont 10\%} Lily } \makebox[22mm][l]{ {\fontsize{7}{7}\selectfont 9\%} max } { {\fontsize{7}{7}\selectfont 6\%} father } \\
Middle Eastern & \makebox[22mm][l]{ {\fontsize{7}{7}\selectfont 20\%} Leila } \makebox[22mm][l]{ {\fontsize{7}{7}\selectfont 18\%} honor } \makebox[22mm][l]{ {\fontsize{7}{7}\selectfont 13\%} grandfather } \makebox[22mm][l]{ {\fontsize{7}{7}\selectfont 10\%} Khalid } \makebox[22mm][l]{ {\fontsize{7}{7}\selectfont 9\%} dear } { {\fontsize{7}{7}\selectfont 9\%} passed } \\
North American & \makebox[22mm][l]{ {\fontsize{7}{7}\selectfont 9\%} owl } \makebox[22mm][l]{ {\fontsize{7}{7}\selectfont 8\%} drifted } \makebox[22mm][l]{ {\fontsize{7}{7}\selectfont 8\%} kofi } \makebox[22mm][l]{ {\fontsize{7}{7}\selectfont 7\%} met } \makebox[22mm][l]{ {\fontsize{7}{7}\selectfont 7\%} Max } { {\fontsize{7}{7}\selectfont 6\%} continued } \\
South American & \makebox[22mm][l]{ {\fontsize{7}{7}\selectfont 12\%} clearing } \makebox[22mm][l]{ {\fontsize{7}{7}\selectfont 9\%} heard } \makebox[22mm][l]{ {\fontsize{7}{7}\selectfont 8\%} waterfall } \makebox[22mm][l]{ {\fontsize{7}{7}\selectfont 8\%} center } \makebox[22mm][l]{ {\fontsize{7}{7}\selectfont 8\%} suddenly } { {\fontsize{7}{7}\selectfont 7\%} create } \\
\bottomrule\\

\null\\[-1.7em]
\multicolumn{2}{l}{\bf Nationality Parent Developed} \\

Developed & \makebox[22mm][l]{ {\fontsize{7}{7}\selectfont 21\%} rolling } \makebox[22mm][l]{ {\fontsize{7}{7}\selectfont 16\%} Leo } \makebox[22mm][l]{ {\fontsize{7}{7}\selectfont 15\%} nestled } \makebox[22mm][l]{ {\fontsize{7}{7}\selectfont 13\%} kaito } \makebox[22mm][l]{ {\fontsize{7}{7}\selectfont 13\%} Max } { {\fontsize{7}{7}\selectfont 12\%} Lily } \\
Developing & \makebox[22mm][l]{ {\fontsize{7}{7}\selectfont 15\%} honor } \makebox[22mm][l]{ {\fontsize{7}{7}\selectfont 14\%} lush } \makebox[22mm][l]{ {\fontsize{7}{7}\selectfont 12\%} roamed } \makebox[22mm][l]{ {\fontsize{7}{7}\selectfont 10\%} hue } \makebox[22mm][l]{ {\fontsize{7}{7}\selectfont 10\%} passed } { {\fontsize{7}{7}\selectfont 10\%} grandfather } \\
\bottomrule\\

\null\\[-1.7em]
\multicolumn{2}{l}{\bf Ethnicity} \\

African-Amer. & \makebox[22mm][l]{ {\fontsize{7}{7}\selectfont 77\%} africa } \makebox[22mm][l]{ {\fontsize{7}{7}\selectfont 75\%} kofi } \makebox[22mm][l]{ {\fontsize{7}{7}\selectfont 44\%} akua } \makebox[22mm][l]{ {\fontsize{7}{7}\selectfont 42\%} kwame } \makebox[22mm][l]{ {\fontsize{7}{7}\selectfont 34\%} mama } { {\fontsize{7}{7}\selectfont 25\%} honor } \\
Asian & \makebox[22mm][l]{ {\fontsize{7}{7}\selectfont 42\%} Mei } \makebox[22mm][l]{ {\fontsize{7}{7}\selectfont 30\%} nestled } \makebox[22mm][l]{ {\fontsize{7}{7}\selectfont 23\%} honor } \makebox[22mm][l]{ {\fontsize{7}{7}\selectfont 22\%} spreading } \makebox[22mm][l]{ {\fontsize{7}{7}\selectfont 17\%} box } { {\fontsize{7}{7}\selectfont 17\%} rolling } \\
European-Amer. & \makebox[22mm][l]{ {\fontsize{7}{7}\selectfont 47\%} rolling } \makebox[22mm][l]{ {\fontsize{7}{7}\selectfont 30\%} sophie } \makebox[22mm][l]{ {\fontsize{7}{7}\selectfont 25\%} stood } \makebox[22mm][l]{ {\fontsize{7}{7}\selectfont 23\%} clearing } \makebox[22mm][l]{ {\fontsize{7}{7}\selectfont 23\%} Liam } { {\fontsize{7}{7}\selectfont 21\%} acorn } \\
Latino & \makebox[22mm][l]{ {\fontsize{7}{7}\selectfont 58\%} Lucia } \makebox[22mm][l]{ {\fontsize{7}{7}\selectfont 36\%} Sofía } \makebox[22mm][l]{ {\fontsize{7}{7}\selectfont 34\%} Carlos } \makebox[22mm][l]{ {\fontsize{7}{7}\selectfont 22\%} nestled } \makebox[22mm][l]{ {\fontsize{7}{7}\selectfont 18\%} danced } { {\fontsize{7}{7}\selectfont 17\%} create } \\
Middle-Eastern & \makebox[22mm][l]{ {\fontsize{7}{7}\selectfont 70\%} dune } \makebox[22mm][l]{ {\fontsize{7}{7}\selectfont 39\%} khalid } \makebox[22mm][l]{ {\fontsize{7}{7}\selectfont 36\%} Leila } \makebox[22mm][l]{ {\fontsize{7}{7}\selectfont 28\%} gather } \makebox[22mm][l]{ {\fontsize{7}{7}\selectfont 25\%} revealed } { {\fontsize{7}{7}\selectfont 16\%} dear } \\
White & \makebox[22mm][l]{ {\fontsize{7}{7}\selectfont 22\%} Leo } \makebox[22mm][l]{ {\fontsize{7}{7}\selectfont 21\%} owl } \makebox[22mm][l]{ {\fontsize{7}{7}\selectfont 18\%} branch } \makebox[22mm][l]{ {\fontsize{7}{7}\selectfont 17\%} drifted } \makebox[22mm][l]{ {\fontsize{7}{7}\selectfont 17\%} clearing } { {\fontsize{7}{7}\selectfont 14\%} did } \\
\bottomrule\\

\null\\[-1.7em]
\multicolumn{2}{l}{\bf Religion} \\

Atheist & \makebox[22mm][l]{ {\fontsize{7}{7}\selectfont 16\%} Max } \makebox[22mm][l]{ {\fontsize{7}{7}\selectfont 8\%} gazed } \makebox[22mm][l]{ {\fontsize{7}{7}\selectfont 7\%} noticed } \makebox[22mm][l]{ {\fontsize{7}{7}\selectfont 7\%} leaving } \makebox[22mm][l]{ {\fontsize{7}{7}\selectfont 7\%} knowing } { {\fontsize{7}{7}\selectfont 6\%} drifted } \\
Buddhist & \makebox[22mm][l]{ {\fontsize{7}{7}\selectfont 23\%} kaito } \makebox[22mm][l]{ {\fontsize{7}{7}\selectfont 13\%} asked } \makebox[22mm][l]{ {\fontsize{7}{7}\selectfont 12\%} continued } \makebox[22mm][l]{ {\fontsize{7}{7}\selectfont 11\%} noticed } \makebox[22mm][l]{ {\fontsize{7}{7}\selectfont 11\%} compassion } { {\fontsize{7}{7}\selectfont 10\%} taught } \\
Christian & \makebox[22mm][l]{ {\fontsize{7}{7}\selectfont 13\%} delighted } \makebox[22mm][l]{ {\fontsize{7}{7}\selectfont 10\%} bedtime } \makebox[22mm][l]{ {\fontsize{7}{7}\selectfont 9\%} create } \makebox[22mm][l]{ {\fontsize{7}{7}\selectfont 8\%} honor } \makebox[22mm][l]{ {\fontsize{7}{7}\selectfont 8\%} evening } { {\fontsize{7}{7}\selectfont 6\%} waterfall } \\
Hindu & \makebox[22mm][l]{ {\fontsize{7}{7}\selectfont 12\%} father } \makebox[22mm][l]{ {\fontsize{7}{7}\selectfont 9\%} rohan } \makebox[22mm][l]{ {\fontsize{7}{7}\selectfont 8\%} gather } \makebox[22mm][l]{ {\fontsize{7}{7}\selectfont 7\%} roamed } \makebox[22mm][l]{ {\fontsize{7}{7}\selectfont 6\%} symbol } { {\fontsize{7}{7}\selectfont 5\%} approached } \\
Jew & \makebox[22mm][l]{ {\fontsize{7}{7}\selectfont 20\%} rolling } \makebox[22mm][l]{ {\fontsize{7}{7}\selectfont 17\%} nestled } \makebox[22mm][l]{ {\fontsize{7}{7}\selectfont 9\%} hold } \makebox[22mm][l]{ {\fontsize{7}{7}\selectfont 7\%} passed } \makebox[22mm][l]{ {\fontsize{7}{7}\selectfont 7\%} symbol } { {\fontsize{7}{7}\selectfont 7\%} family } \\
Muslim & \makebox[22mm][l]{ {\fontsize{7}{7}\selectfont 8\%} sang } \makebox[22mm][l]{ {\fontsize{7}{7}\selectfont 7\%} wandered } \makebox[22mm][l]{ {\fontsize{7}{7}\selectfont 7\%} whispered } \makebox[22mm][l]{ {\fontsize{7}{7}\selectfont 6\%} beneath } \makebox[22mm][l]{ {\fontsize{7}{7}\selectfont 6\%} just } { {\fontsize{7}{7}\selectfont 6\%} high } \\
\bottomrule\\

\null\\[-1.7em]
\multicolumn{2}{l}{\bf Role} \\

father & \makebox[22mm][l]{ {\fontsize{7}{7}\selectfont 12\%} father } \makebox[22mm][l]{ {\fontsize{7}{7}\selectfont 9\%} rohan } \makebox[22mm][l]{ {\fontsize{7}{7}\selectfont 8\%} gather } \makebox[22mm][l]{ {\fontsize{7}{7}\selectfont 7\%} roamed } \makebox[22mm][l]{ {\fontsize{7}{7}\selectfont 6\%} symbol } { {\fontsize{7}{7}\selectfont 5\%} approached } \\
mother & \makebox[22mm][l]{ {\fontsize{7}{7}\selectfont 8\%} sang } \makebox[22mm][l]{ {\fontsize{7}{7}\selectfont 7\%} wandered } \makebox[22mm][l]{ {\fontsize{7}{7}\selectfont 7\%} whispered } \makebox[22mm][l]{ {\fontsize{7}{7}\selectfont 6\%} beneath } \makebox[22mm][l]{ {\fontsize{7}{7}\selectfont 6\%} just } { {\fontsize{7}{7}\selectfont 6\%} high } \\
parent & \makebox[22mm][l]{ {\fontsize{7}{7}\selectfont 13\%} delighted } \makebox[22mm][l]{ {\fontsize{7}{7}\selectfont 10\%} bedtime } \makebox[22mm][l]{ {\fontsize{7}{7}\selectfont 9\%} create } \makebox[22mm][l]{ {\fontsize{7}{7}\selectfont 8\%} honor } \makebox[22mm][l]{ {\fontsize{7}{7}\selectfont 8\%} evening } { {\fontsize{7}{7}\selectfont 6\%} waterfall } \\
\bottomrule\\

\end{tabular}
\vspace{-6mm}
\caption{Top words in the \textbf{text of the generated story by Llama3 model} that correlate (Pearson) with the sociocultural factor. The terms \textit{child}, \textit{daughter}, and \textit{son} have been removed, as they are almost present at the start of the generation}
\label{tab:corr_bias_text_llama3}
\end{table*}

\begin{table*}[htbp]
\fontsize{8}{8}\selectfont
\begin{tabular}{p{2cm}l}
\toprule \\[-0.5em]
\null\\[-1.7em]
\multicolumn{2}{l}{\bf Gender} \\

child & \makebox[22mm][l]{ {\fontsize{7}{7}\selectfont 8\%} called } \makebox[22mm][l]{ {\fontsize{7}{7}\selectfont 7\%} smiled } \makebox[22mm][l]{ {\fontsize{7}{7}\selectfont 7\%} adventurous } \makebox[22mm][l]{ {\fontsize{7}{7}\selectfont 6\%} valley } \makebox[22mm][l]{ {\fontsize{7}{7}\selectfont 6\%} diego } { {\fontsize{7}{7}\selectfont 6\%} Mei } \\
daughter & \makebox[22mm][l]{ {\fontsize{7}{7}\selectfont 21\%} Lily } \makebox[22mm][l]{ {\fontsize{7}{7}\selectfont 11\%} brought } \makebox[22mm][l]{ {\fontsize{7}{7}\selectfont 11\%} dear } \makebox[22mm][l]{ {\fontsize{7}{7}\selectfont 10\%} ada } \makebox[22mm][l]{ {\fontsize{7}{7}\selectfont 10\%} sweet } { {\fontsize{7}{7}\selectfont 10\%} Meera } \\
son & \makebox[22mm][l]{ {\fontsize{7}{7}\selectfont 18\%} Liam } \makebox[22mm][l]{ {\fontsize{7}{7}\selectfont 15\%} Ali } \makebox[22mm][l]{ {\fontsize{7}{7}\selectfont 8\%} away } \makebox[22mm][l]{ {\fontsize{7}{7}\selectfont 7\%} cave } \makebox[22mm][l]{ {\fontsize{7}{7}\selectfont 7\%} brave } { {\fontsize{7}{7}\selectfont 7\%} heard } \\
\bottomrule\\

\null\\[-1.7em]
\multicolumn{2}{l}{\bf Nationality Parent Group} \\

Africa & \makebox[22mm][l]{ {\fontsize{7}{7}\selectfont 19\%} ada } \makebox[22mm][l]{ {\fontsize{7}{7}\selectfont 10\%} river } \makebox[22mm][l]{ {\fontsize{7}{7}\selectfont 9\%} called } \makebox[22mm][l]{ {\fontsize{7}{7}\selectfont 9\%} approached } \makebox[22mm][l]{ {\fontsize{7}{7}\selectfont 8\%} water } { {\fontsize{7}{7}\selectfont 7\%} adventurous } \\
Asia & \makebox[22mm][l]{ {\fontsize{7}{7}\selectfont 21\%} near } \makebox[22mm][l]{ {\fontsize{7}{7}\selectfont 20\%} Mei } \makebox[22mm][l]{ {\fontsize{7}{7}\selectfont 15\%} Meera } \makebox[22mm][l]{ {\fontsize{7}{7}\selectfont 13\%} lush } \makebox[22mm][l]{ {\fontsize{7}{7}\selectfont 9\%} water } { {\fontsize{7}{7}\selectfont 7\%} grateful } \\
European & \makebox[22mm][l]{ {\fontsize{7}{7}\selectfont 18\%} Liam } \makebox[22mm][l]{ {\fontsize{7}{7}\selectfont 18\%} Lily } \makebox[22mm][l]{ {\fontsize{7}{7}\selectfont 14\%} kingdom } \makebox[22mm][l]{ {\fontsize{7}{7}\selectfont 12\%} nestled } \makebox[22mm][l]{ {\fontsize{7}{7}\selectfont 10\%} spent } { {\fontsize{7}{7}\selectfont 9\%} dragon } \\
Middle Eastern & \makebox[22mm][l]{ {\fontsize{7}{7}\selectfont 33\%} ali } \makebox[22mm][l]{ {\fontsize{7}{7}\selectfont 21\%} called } \makebox[22mm][l]{ {\fontsize{7}{7}\selectfont 19\%} known } \makebox[22mm][l]{ {\fontsize{7}{7}\selectfont 14\%} cave } \makebox[22mm][l]{ {\fontsize{7}{7}\selectfont 14\%} valley } { {\fontsize{7}{7}\selectfont 13\%} magnificent } \\
North American & \makebox[22mm][l]{ {\fontsize{7}{7}\selectfont 17\%} Lily } \makebox[22mm][l]{ {\fontsize{7}{7}\selectfont 11\%} bed } \makebox[22mm][l]{ {\fontsize{7}{7}\selectfont 9\%} end } \makebox[22mm][l]{ {\fontsize{7}{7}\selectfont 7\%} kingdom } \makebox[22mm][l]{ {\fontsize{7}{7}\selectfont 7\%} smiled } { {\fontsize{7}{7}\selectfont 6\%} drifted } \\
South American & \makebox[22mm][l]{ {\fontsize{7}{7}\selectfont 27\%} Diego } \makebox[22mm][l]{ {\fontsize{7}{7}\selectfont 26\%} vibrant } \makebox[22mm][l]{ {\fontsize{7}{7}\selectfont 13\%} lush } \makebox[22mm][l]{ {\fontsize{7}{7}\selectfont 9\%} soft } \makebox[22mm][l]{ {\fontsize{7}{7}\selectfont 7\%} secret } { {\fontsize{7}{7}\selectfont 7\%} surrounded } \\
\bottomrule\\

\null\\[-1.7em]
\multicolumn{2}{l}{\bf Nationality Parent Developed} \\

Developed & \makebox[22mm][l]{ {\fontsize{7}{7}\selectfont 27\%} Lily } \makebox[22mm][l]{ {\fontsize{7}{7}\selectfont 16\%} bed } \makebox[22mm][l]{ {\fontsize{7}{7}\selectfont 16\%} kingdom } \makebox[22mm][l]{ {\fontsize{7}{7}\selectfont 14\%} Liam } \makebox[22mm][l]{ {\fontsize{7}{7}\selectfont 14\%} end } { {\fontsize{7}{7}\selectfont 13\%} nestled } \\
Developing & \makebox[22mm][l]{ {\fontsize{7}{7}\selectfont 21\%} near } \makebox[22mm][l]{ {\fontsize{7}{7}\selectfont 21\%} river } \makebox[22mm][l]{ {\fontsize{7}{7}\selectfont 20\%} lush } \makebox[22mm][l]{ {\fontsize{7}{7}\selectfont 16\%} called } \makebox[22mm][l]{ {\fontsize{7}{7}\selectfont 14\%} adventurous } { {\fontsize{7}{7}\selectfont 13\%} Ali } \\
\bottomrule\\

\null\\[-1.7em]
\multicolumn{2}{l}{\bf Ethnicity} \\

African-Amer. & \makebox[22mm][l]{ {\fontsize{7}{7}\selectfont 63\%} kofi } \makebox[22mm][l]{ {\fontsize{7}{7}\selectfont 60\%} ada } \makebox[22mm][l]{ {\fontsize{7}{7}\selectfont 47\%} vibrant } \makebox[22mm][l]{ {\fontsize{7}{7}\selectfont 30\%} lush } \makebox[22mm][l]{ {\fontsize{7}{7}\selectfont 24\%} remember } { {\fontsize{7}{7}\selectfont 15\%} wise } \\
Asian & \makebox[22mm][l]{ {\fontsize{7}{7}\selectfont 63\%} Mei } \makebox[22mm][l]{ {\fontsize{7}{7}\selectfont 26\%} grew } \makebox[22mm][l]{ {\fontsize{7}{7}\selectfont 24\%} lush } \makebox[22mm][l]{ {\fontsize{7}{7}\selectfont 19\%} nestled } \makebox[22mm][l]{ {\fontsize{7}{7}\selectfont 18\%} grow } { {\fontsize{7}{7}\selectfont 15\%} took } \\
European-Amer. & \makebox[22mm][l]{ {\fontsize{7}{7}\selectfont 36\%} mia } \makebox[22mm][l]{ {\fontsize{7}{7}\selectfont 28\%} kingdom } \makebox[22mm][l]{ {\fontsize{7}{7}\selectfont 25\%} away } \makebox[22mm][l]{ {\fontsize{7}{7}\selectfont 24\%} Sofia } \makebox[22mm][l]{ {\fontsize{7}{7}\selectfont 24\%} called } { {\fontsize{7}{7}\selectfont 22\%} dragon } \\
Latino & \makebox[22mm][l]{ {\fontsize{7}{7}\selectfont 58\%} Diego } \makebox[22mm][l]{ {\fontsize{7}{7}\selectfont 39\%} Maria } \makebox[22mm][l]{ {\fontsize{7}{7}\selectfont 34\%} Isabela } \makebox[22mm][l]{ {\fontsize{7}{7}\selectfont 33\%} nestled } \makebox[22mm][l]{ {\fontsize{7}{7}\selectfont 28\%} hill } { {\fontsize{7}{7}\selectfont 20\%} approached } \\
Middle-Eastern & \makebox[22mm][l]{ {\fontsize{7}{7}\selectfont 73\%} sand } \makebox[22mm][l]{ {\fontsize{7}{7}\selectfont 44\%} karim } \makebox[22mm][l]{ {\fontsize{7}{7}\selectfont 42\%} Aisha } \makebox[22mm][l]{ {\fontsize{7}{7}\selectfont 39\%} Hassan } \makebox[22mm][l]{ {\fontsize{7}{7}\selectfont 36\%} Noor } { {\fontsize{7}{7}\selectfont 23\%} granted } \\
White & \makebox[22mm][l]{ {\fontsize{7}{7}\selectfont 52\%} Lily } \makebox[22mm][l]{ {\fontsize{7}{7}\selectfont 38\%} away } \makebox[22mm][l]{ {\fontsize{7}{7}\selectfont 30\%} kingdom } \makebox[22mm][l]{ {\fontsize{7}{7}\selectfont 23\%} Liam } \makebox[22mm][l]{ {\fontsize{7}{7}\selectfont 22\%} prince } { {\fontsize{7}{7}\selectfont 20\%} came } \\
\bottomrule\\

\null\\[-1.7em]
\multicolumn{2}{l}{\bf Religion} \\

Atheist & \makebox[22mm][l]{ {\fontsize{7}{7}\selectfont 21\%} Lily } \makebox[22mm][l]{ {\fontsize{7}{7}\selectfont 14\%} Liam } \makebox[22mm][l]{ {\fontsize{7}{7}\selectfont 11\%} valley } \makebox[22mm][l]{ {\fontsize{7}{7}\selectfont 7\%} secret } \makebox[22mm][l]{ {\fontsize{7}{7}\selectfont 7\%} came } { {\fontsize{7}{7}\selectfont 6\%} spent } \\
Buddhist & \makebox[22mm][l]{ {\fontsize{7}{7}\selectfont 16\%} nestled } \makebox[22mm][l]{ {\fontsize{7}{7}\selectfont 14\%} let } \makebox[22mm][l]{ {\fontsize{7}{7}\selectfont 13\%} wise } \makebox[22mm][l]{ {\fontsize{7}{7}\selectfont 10\%} listened } \makebox[22mm][l]{ {\fontsize{7}{7}\selectfont 8\%} gently } { {\fontsize{7}{7}\selectfont 6\%} parent } \\
Christian & \makebox[22mm][l]{ {\fontsize{7}{7}\selectfont 19\%} parent } \makebox[22mm][l]{ {\fontsize{7}{7}\selectfont 7\%} sweet } \makebox[22mm][l]{ {\fontsize{7}{7}\selectfont 7\%} hidden } \makebox[22mm][l]{ {\fontsize{7}{7}\selectfont 6\%} shimmering } \makebox[22mm][l]{ {\fontsize{7}{7}\selectfont 6\%} soft } { {\fontsize{7}{7}\selectfont 5\%} vibrant } \\
Hindu & \makebox[22mm][l]{ {\fontsize{7}{7}\selectfont 7\%} tell } \makebox[22mm][l]{ {\fontsize{7}{7}\selectfont 7\%} magnificent } \makebox[22mm][l]{ {\fontsize{7}{7}\selectfont 7\%} adventurous } \makebox[22mm][l]{ {\fontsize{7}{7}\selectfont 7\%} brave } \makebox[22mm][l]{ {\fontsize{7}{7}\selectfont 7\%} smiled } { {\fontsize{7}{7}\selectfont 6\%} lush } \\
Jew & \makebox[22mm][l]{ {\fontsize{7}{7}\selectfont 22\%} hill } \makebox[22mm][l]{ {\fontsize{7}{7}\selectfont 14\%} bed } \makebox[22mm][l]{ {\fontsize{7}{7}\selectfont 12\%} nestled } \makebox[22mm][l]{ {\fontsize{7}{7}\selectfont 12\%} drifted } \makebox[22mm][l]{ {\fontsize{7}{7}\selectfont 11\%} tell } { {\fontsize{7}{7}\selectfont 8\%} king } \\
Muslim & \makebox[22mm][l]{ {\fontsize{7}{7}\selectfont 35\%} mother } \makebox[22mm][l]{ {\fontsize{7}{7}\selectfont 10\%} Aisha } \makebox[22mm][l]{ {\fontsize{7}{7}\selectfont 6\%} warm } \makebox[22mm][l]{ {\fontsize{7}{7}\selectfont 6\%} went } \makebox[22mm][l]{ {\fontsize{7}{7}\selectfont 5\%} set } { {\fontsize{7}{7}\selectfont 5\%} shimmering } \\
\bottomrule\\

\null\\[-1.7em]
\multicolumn{2}{l}{\bf Role} \\
father & \makebox[22mm][l]{ {\fontsize{7}{7}\selectfont 7\%} tell } \makebox[22mm][l]{ {\fontsize{7}{7}\selectfont 7\%} magnificent } \makebox[22mm][l]{ {\fontsize{7}{7}\selectfont 7\%} adventurous } \makebox[22mm][l]{ {\fontsize{7}{7}\selectfont 7\%} brave } \makebox[22mm][l]{ {\fontsize{7}{7}\selectfont 7\%} smiled } { {\fontsize{7}{7}\selectfont 6\%} lush } \\
mother & \makebox[22mm][l]{ {\fontsize{7}{7}\selectfont 35\%} mother } \makebox[22mm][l]{ {\fontsize{7}{7}\selectfont 10\%} Aisha } \makebox[22mm][l]{ {\fontsize{7}{7}\selectfont 6\%} warm } \makebox[22mm][l]{ {\fontsize{7}{7}\selectfont 6\%} went } \makebox[22mm][l]{ {\fontsize{7}{7}\selectfont 5\%} set } { {\fontsize{7}{7}\selectfont 5\%} shimmering } \\
parent & \makebox[22mm][l]{ {\fontsize{7}{7}\selectfont 19\%} parent } \makebox[22mm][l]{ {\fontsize{7}{7}\selectfont 7\%} sweet } \makebox[22mm][l]{ {\fontsize{7}{7}\selectfont 7\%} hidden } \makebox[22mm][l]{ {\fontsize{7}{7}\selectfont 6\%} shimmering } \makebox[22mm][l]{ {\fontsize{7}{7}\selectfont 6\%} soft } { {\fontsize{7}{7}\selectfont 5\%} vibrant } \\
\bottomrule\\

\end{tabular}
\vspace{-6mm}
\caption{Top words in the \textbf{text of the generated story by Mixtral model} that correlate (Pearson) with the sociocultural factor. The terms \textit{child}, \textit{daughter}, and \textit{son} have been removed, as they are almost present at the start of the generation}
\label{tab:corr_bias_text_mixtral}
\end{table*}

\subsection{Protagonist Analysis}

\Cref{tab:corr_bias_attr} shows the top words in children attributes in the generated story that correlate (Pearson) with the target sociocultural variable (e.g., child gender or parent’s nationality).
\begin{table*}[htbp]
\fontsize{8}{8}\selectfont
\begin{tabular}{p{2cm}l}
\toprule \\[-0.5em]
\null\\[-1.7em]
\multicolumn{2}{l}{\bf Gender Of Child} \\
child & \makebox[22mm][l]{ {\fontsize{7}{7}\selectfont \phantom{0}7\%} \colorbox{blue!20}{little} } \makebox[22mm][l]{ {\fontsize{7}{7}\selectfont 6\%} \colorbox{blue!20}{fur} } \makebox[22mm][l]{ {\fontsize{7}{7}\selectfont 6\%} \colorbox{blue!20}{soft} } \makebox[22mm][l]{ {\fontsize{7}{7}\selectfont 6\%} \colorbox{blue!20}{shimmering} } \makebox[22mm][l]{ {\fontsize{7}{7}\selectfont 5\%} \colorbox{blue!20}{white coat} } \makebox[22mm][l]{ {\fontsize{7}{7}\selectfont 5\%} \colorbox{red!20}{hopeful} } \\
daughter & \makebox[22mm][l]{ {\fontsize{7}{7}\selectfont 11\%} \colorbox{blue!20}{hair} } \makebox[22mm][l]{ {\fontsize{7}{7}\selectfont 8\%} \colorbox{blue!20}{black} } \makebox[22mm][l]{ {\fontsize{7}{7}\selectfont 8\%} \colorbox{yellow!20}{gentle} } \makebox[22mm][l]{ {\fontsize{7}{7}\selectfont 7\%} \colorbox{green!20}{imaginative} } \makebox[22mm][l]{ {\fontsize{7}{7}\selectfont 5\%} \colorbox{green!20}{bright} } \makebox[22mm][l]{ {\fontsize{7}{7}\selectfont 5\%} \colorbox{yellow!20}{loving} } \\
son & \makebox[22mm][l]{ {\fontsize{7}{7}\selectfont 14\%} \colorbox{blue!20}{young} } \makebox[22mm][l]{ {\fontsize{7}{7}\selectfont 6\%} \colorbox{green!20}{adventurous} } \makebox[22mm][l]{ {\fontsize{7}{7}\selectfont 6\%} \colorbox{yellow!20}{hero} } \makebox[22mm][l]{ {\fontsize{7}{7}\selectfont 5\%} \colorbox{yellow!20}{brave} } \makebox[22mm][l]{ {\fontsize{7}{7}\selectfont 5\%} \colorbox{red!20}{eager} } \makebox[22mm][l]{ {\fontsize{7}{7}\selectfont 5\%} \colorbox{green!20}{wise} } \\

\bottomrule\\
\null\\[-1.7em]
\multicolumn{2}{l}{\bf Nationality Parent Group} \\
Africa & \makebox[22mm][l]{ {\fontsize{7}{7}\selectfont 10\%} \colorbox{green!20}{wise} } \makebox[22mm][l]{ {\fontsize{7}{7}\selectfont 6\%} \colorbox{green!20}{clever} } \makebox[22mm][l]{ {\fontsize{7}{7}\selectfont 5\%} \colorbox{gray!20}{spirit} } \makebox[22mm][l]{ {\fontsize{7}{7}\selectfont 4\%} \colorbox{green!20}{decisive} } \makebox[22mm][l]{ {\fontsize{7}{7}\selectfont 4\%} \colorbox{blue!20}{young} } \makebox[22mm][l]{ {\fontsize{7}{7}\selectfont 4\%} \colorbox{yellow!20}{respectful} } \\
Asia & \makebox[22mm][l]{ {\fontsize{7}{7}\selectfont 10\%} \colorbox{yellow!20}{pure} } \makebox[22mm][l]{ {\fontsize{7}{7}\selectfont 7\%} \colorbox{yellow!20}{gentle} } \makebox[22mm][l]{ {\fontsize{7}{7}\selectfont 7\%} \colorbox{gray!20}{chosen} } \makebox[22mm][l]{ {\fontsize{7}{7}\selectfont 6\%} \colorbox{yellow!20}{kind} } \makebox[22mm][l]{ {\fontsize{7}{7}\selectfont 5\%} \colorbox{yellow!20}{loving} } \makebox[22mm][l]{ {\fontsize{7}{7}\selectfont 5\%} \colorbox{yellow!20}{hearted} } \\
European & \makebox[22mm][l]{ {\fontsize{7}{7}\selectfont 7\%} \colorbox{yellow!20}{friendly} } \makebox[22mm][l]{ {\fontsize{7}{7}\selectfont 5\%} \colorbox{red!20}{suprised} } \makebox[22mm][l]{ {\fontsize{7}{7}\selectfont 5\%} \colorbox{blue!20}{golden} } \makebox[22mm][l]{ {\fontsize{7}{7}\selectfont 4\%} \colorbox{gray!20}{drawn} } \makebox[22mm][l]{ {\fontsize{7}{7}\selectfont 4\%} \colorbox{red!20}{warm} } \makebox[22mm][l]{ {\fontsize{7}{7}\selectfont 4\%} \colorbox{green!20}{imaginative} } \\
Middle Eastern & \makebox[22mm][l]{ {\fontsize{7}{7}\selectfont 8\%} \colorbox{blue!20}{young} } \makebox[22mm][l]{ {\fontsize{7}{7}\selectfont 6\%} \colorbox{yellow!20}{hero} } \makebox[22mm][l]{ {\fontsize{7}{7}\selectfont 5\%} \colorbox{blue!20}{wide} } \makebox[22mm][l]{ {\fontsize{7}{7}\selectfont 5\%} \colorbox{red!20}{proud} } \makebox[22mm][l]{ {\fontsize{7}{7}\selectfont 4\%} \colorbox{yellow!20}{courageous} } \makebox[22mm][l]{ {\fontsize{7}{7}\selectfont 3\%} \colorbox{gray!20}{challenged} } \\
North American & \makebox[22mm][l]{ {\fontsize{7}{7}\selectfont 5\%} \colorbox{red!20}{suprised} } \makebox[22mm][l]{ {\fontsize{7}{7}\selectfont 4\%} \colorbox{red!20}{thrilled} } \makebox[22mm][l]{ {\fontsize{7}{7}\selectfont 4\%} \colorbox{blue!20}{tiny} } \makebox[22mm][l]{ {\fontsize{7}{7}\selectfont 4\%} \colorbox{red!20}{excited} } \makebox[22mm][l]{ {\fontsize{7}{7}\selectfont 4\%} \colorbox{green!20}{dreamer} } \makebox[22mm][l]{ {\fontsize{7}{7}\selectfont 3\%} \colorbox{green!20}{imaginative} } \\
South American & \makebox[22mm][l]{ {\fontsize{7}{7}\selectfont 8\%} \colorbox{red!20}{joyous} } \makebox[22mm][l]{ {\fontsize{7}{7}\selectfont 7\%} \colorbox{blue!20}{wings} } \makebox[22mm][l]{ {\fontsize{7}{7}\selectfont 7\%} \colorbox{blue!20}{sparkling} } \makebox[22mm][l]{ {\fontsize{7}{7}\selectfont 6\%} \colorbox{blue!20}{tiny} } \makebox[22mm][l]{ {\fontsize{7}{7}\selectfont 5\%} \colorbox{blue!20}{eyes} } \makebox[22mm][l]{ {\fontsize{7}{7}\selectfont 5\%} \colorbox{gray!20}{guardian} } \\

\bottomrule\\
\null\\[-1.7em]
\multicolumn{2}{l}{\bf Nationality Parent Developed} \\
Developed & \makebox[22mm][l]{ {\fontsize{7}{7}\selectfont 7\%} \colorbox{red!20}{suprised} } \makebox[22mm][l]{ {\fontsize{7}{7}\selectfont 6\%} \colorbox{yellow!20}{friendly} } \makebox[22mm][l]{ {\fontsize{7}{7}\selectfont 6\%} \colorbox{blue!20}{little} } \makebox[22mm][l]{ {\fontsize{7}{7}\selectfont 6\%} \colorbox{blue!20}{fur} } \makebox[22mm][l]{ {\fontsize{7}{7}\selectfont 4\%} \colorbox{yellow!20}{helpful} } \makebox[22mm][l]{ {\fontsize{7}{7}\selectfont 4\%} \colorbox{blue!20}{golden} } \\
Developing & \makebox[22mm][l]{ {\fontsize{7}{7}\selectfont 9\%} \colorbox{blue!20}{young} } \makebox[22mm][l]{ {\fontsize{7}{7}\selectfont 8\%} \colorbox{green!20}{wise} } \makebox[22mm][l]{ {\fontsize{7}{7}\selectfont 4\%} \colorbox{blue!20}{sparkling} } \makebox[22mm][l]{ {\fontsize{7}{7}\selectfont 4\%} \colorbox{yellow!20}{loved} } \makebox[22mm][l]{ {\fontsize{7}{7}\selectfont 4\%} \colorbox{blue!20}{black} } \makebox[22mm][l]{ {\fontsize{7}{7}\selectfont 4\%} \colorbox{yellow!20}{leader} } \\

\bottomrule\\

\null\\[-1.7em]
\multicolumn{2}{l}{\bf Ethnicity Of Parent} \\

African-Amer. & \makebox[22mm][l]{ {\fontsize{7}{7}\selectfont 38\%} \colorbox{gray!20}{heritage} } \makebox[22mm][l]{ {\fontsize{7}{7}\selectfont 34\%} \colorbox{red!20}{proud} } \makebox[22mm][l]{ {\fontsize{7}{7}\selectfont 28\%} \colorbox{gray!20}{connected} } \makebox[22mm][l]{ {\fontsize{7}{7}\selectfont 26\%} \colorbox{blue!20}{strong} } \makebox[22mm][l]{ {\fontsize{7}{7}\selectfont 23\%} \colorbox{red!20}{inspired} } \makebox[22mm][l]{ {\fontsize{7}{7}\selectfont 23\%} \colorbox{yellow!20}{purposeful} } \\
Asian & \makebox[22mm][l]{ {\fontsize{7}{7}\selectfont 15\%} \colorbox{green!20}{wise} } \makebox[22mm][l]{ {\fontsize{7}{7}\selectfont 15\%} \colorbox{yellow!20}{noble} } \makebox[22mm][l]{ {\fontsize{7}{7}\selectfont 13\%} \colorbox{yellow!20}{kind} } \makebox[22mm][l]{ {\fontsize{7}{7}\selectfont 11\%} \colorbox{red!20}{peaceful} } \makebox[22mm][l]{ {\fontsize{7}{7}\selectfont 11\%} \colorbox{yellow!20}{perseverant} } \makebox[22mm][l]{ {\fontsize{7}{7}\selectfont 11\%} \colorbox{yellow!20}{hearted} } \\
European-Amer. & \makebox[22mm][l]{ {\fontsize{7}{7}\selectfont 19\%} \colorbox{blue!20}{golden} } \makebox[22mm][l]{ {\fontsize{7}{7}\selectfont 17\%} \colorbox{blue!20}{blue} } \makebox[22mm][l]{ {\fontsize{7}{7}\selectfont 15\%} \colorbox{yellow!20}{friendly} } \makebox[22mm][l]{ {\fontsize{7}{7}\selectfont 13\%} \colorbox{green!20}{open} } \makebox[22mm][l]{ {\fontsize{7}{7}\selectfont 13\%} \colorbox{yellow!20}{brave} } \makebox[22mm][l]{ {\fontsize{7}{7}\selectfont 13\%} \colorbox{green!20}{observant} } \\
Latino & \makebox[22mm][l]{ {\fontsize{7}{7}\selectfont 27\%} \colorbox{gray!20}{oriented} } \makebox[22mm][l]{ {\fontsize{7}{7}\selectfont 25\%} \colorbox{gray!20}{family} } \makebox[22mm][l]{ {\fontsize{7}{7}\selectfont 16\%} \colorbox{blue!20}{sparkling} } \makebox[22mm][l]{ {\fontsize{7}{7}\selectfont 16\%} \colorbox{yellow!20}{loving} } \makebox[22mm][l]{ {\fontsize{7}{7}\selectfont 13\%} \colorbox{gray!20}{connected} } \makebox[22mm][l]{ {\fontsize{7}{7}\selectfont 12\%} \colorbox{blue!20}{eyes} } \\
Middle-Eastern & \makebox[22mm][l]{ {\fontsize{7}{7}\selectfont 20\%} \colorbox{green!20}{wise} } \makebox[22mm][l]{ {\fontsize{7}{7}\selectfont 14\%} \colorbox{yellow!20}{generous} } \makebox[22mm][l]{ {\fontsize{7}{7}\selectfont 13\%} \colorbox{yellow!20}{selfless} } \makebox[22mm][l]{ {\fontsize{7}{7}\selectfont 13\%} \colorbox{yellow!20}{compassionate} } \makebox[22mm][l]{ {\fontsize{7}{7}\selectfont 12\%} \colorbox{gray!20}{weaver} } \makebox[22mm][l]{ {\fontsize{7}{7}\selectfont 10\%} \colorbox{green!20}{clever} } \\
White & \makebox[22mm][l]{ {\fontsize{7}{7}\selectfont 21\%} \colorbox{yellow!20}{friendly} } \makebox[22mm][l]{ {\fontsize{7}{7}\selectfont 16\%} \colorbox{yellow!20}{helpful} } \makebox[22mm][l]{ {\fontsize{7}{7}\selectfont 15\%} \colorbox{red!20}{empathetic} } \makebox[22mm][l]{ {\fontsize{7}{7}\selectfont 11\%} \colorbox{blue!20}{blue} } \makebox[22mm][l]{ {\fontsize{7}{7}\selectfont 10\%} \colorbox{blue!20}{brown} } \makebox[22mm][l]{ {\fontsize{7}{7}\selectfont 10\%} \colorbox{red!20}{grateful} } \\

\bottomrule\\
\null\\[-1.7em]
\multicolumn{2}{l}{\bf Religion Of Parent} \\

Atheist & \makebox[22mm][l]{ {\fontsize{7}{7}\selectfont 46\%} \colorbox{green!20}{minded} } \makebox[22mm][l]{ {\fontsize{7}{7}\selectfont 45\%} \colorbox{green!20}{inquisitive} } \makebox[22mm][l]{ {\fontsize{7}{7}\selectfont 36\%} \colorbox{green!20}{open} } \makebox[22mm][l]{ {\fontsize{7}{7}\selectfont 31\%} \colorbox{gray!20}{nature} } \makebox[22mm][l]{ {\fontsize{7}{7}\selectfont 28\%} \colorbox{green!20}{observant} } \makebox[22mm][l]{ {\fontsize{7}{7}\selectfont 25\%} \colorbox{gray!20}{connected} } \\
Buddhist & \makebox[22mm][l]{ {\fontsize{7}{7}\selectfont 40\%} \colorbox{green!20}{mindful} } \makebox[22mm][l]{ {\fontsize{7}{7}\selectfont 33\%} \colorbox{green!20}{understanding} } \makebox[22mm][l]{ {\fontsize{7}{7}\selectfont 31\%} \colorbox{yellow!20}{peaceful} } \makebox[22mm][l]{ {\fontsize{7}{7}\selectfont 31\%} \colorbox{yellow!20}{compassionate} } \makebox[22mm][l]{ {\fontsize{7}{7}\selectfont 21\%} \colorbox{red!20}{empathetic} } \makebox[22mm][l]{ {\fontsize{7}{7}\selectfont 18\%} \colorbox{yellow!20}{patient} } \\
Christian & \makebox[22mm][l]{ {\fontsize{7}{7}\selectfont 39\%} \colorbox{yellow!20}{faithful} } \makebox[22mm][l]{ {\fontsize{7}{7}\selectfont 32\%} \colorbox{yellow!20}{trusting} } \makebox[22mm][l]{ {\fontsize{7}{7}\selectfont 32\%} \colorbox{yellow!20}{loving} } \makebox[22mm][l]{ {\fontsize{7}{7}\selectfont 27\%} \colorbox{yellow!20}{caring} } \makebox[22mm][l]{ {\fontsize{7}{7}\selectfont 15\%} \colorbox{red!20}{hopeful} } \makebox[22mm][l]{ {\fontsize{7}{7}\selectfont 14\%} \colorbox{yellow!20}{peaceful} } \\
Hindu & \makebox[22mm][l]{ {\fontsize{7}{7}\selectfont 22\%} \colorbox{yellow!20}{pure} } \makebox[22mm][l]{ {\fontsize{7}{7}\selectfont 17\%} \colorbox{yellow!20}{respectful} } \makebox[22mm][l]{ {\fontsize{7}{7}\selectfont 16\%} \colorbox{gray!20}{magical} } \makebox[22mm][l]{ {\fontsize{7}{7}\selectfont 16\%} \colorbox{yellow!20}{selfless} } \makebox[22mm][l]{ {\fontsize{7}{7}\selectfont 15\%} \colorbox{green!20}{wise} } \makebox[22mm][l]{ {\fontsize{7}{7}\selectfont 12\%} \colorbox{yellow!20}{brave} } \\
Jew & \makebox[22mm][l]{ {\fontsize{7}{7}\selectfont 48\%} \colorbox{gray!20}{heritage} } \makebox[22mm][l]{ {\fontsize{7}{7}\selectfont 42\%} \colorbox{gray!20}{tradition} } \makebox[22mm][l]{ {\fontsize{7}{7}\selectfont 30\%} \colorbox{gray!20}{family} } \makebox[22mm][l]{ {\fontsize{7}{7}\selectfont 27\%} \colorbox{gray!20}{oriented} } \makebox[22mm][l]{ {\fontsize{7}{7}\selectfont 23\%} \colorbox{red!20}{excitable} } \makebox[22mm][l]{ {\fontsize{7}{7}\selectfont 22\%} \colorbox{red!20}{proud} } \\
Muslim & \makebox[22mm][l]{ {\fontsize{7}{7}\selectfont 26\%} \colorbox{yellow!20}{faithful} } \makebox[22mm][l]{ {\fontsize{7}{7}\selectfont 21\%} \colorbox{red!20}{grateful} } \makebox[22mm][l]{ {\fontsize{7}{7}\selectfont 19\%} \colorbox{yellow!20}{devoted} } \makebox[22mm][l]{ {\fontsize{7}{7}\selectfont 18\%} \colorbox{yellow!20}{patient} } \makebox[22mm][l]{ {\fontsize{7}{7}\selectfont 17\%} \colorbox{yellow!20}{kind} } \makebox[22mm][l]{ {\fontsize{7}{7}\selectfont 16\%} \colorbox{yellow!20}{spiritual} } \\

\bottomrule\\
\null\\[-1.7em]
\multicolumn{2}{l}{\bf Role Of Parent} \\

father & \makebox[22mm][l]{ {\fontsize{7}{7}\selectfont 5\%} \colorbox{green!20}{decisive} } \makebox[22mm][l]{ {\fontsize{7}{7}\selectfont 5\%} \colorbox{green!20}{wise} } \makebox[22mm][l]{ {\fontsize{7}{7}\selectfont 5\%} \colorbox{gray!20}{explorer} } \makebox[22mm][l]{ {\fontsize{7}{7}\selectfont 4\%} \colorbox{yellow!20}{protector} } \makebox[22mm][l]{ {\fontsize{7}{7}\selectfont 4\%} \colorbox{red!20}{amazed} } \makebox[22mm][l]{ {\fontsize{7}{7}\selectfont 4\%} \colorbox{gray!20}{celebrated} } \\
mother & \makebox[22mm][l]{ {\fontsize{7}{7}\selectfont 6\%} \colorbox{red!20}{dear} } \makebox[22mm][l]{ {\fontsize{7}{7}\selectfont 5\%} \colorbox{blue!20}{shimmering} } \makebox[22mm][l]{ {\fontsize{7}{7}\selectfont 5\%} \colorbox{gray!20}{free} } \makebox[22mm][l]{ {\fontsize{7}{7}\selectfont 4\%} \colorbox{yellow!20}{loved} } \makebox[22mm][l]{ {\fontsize{7}{7}\selectfont 4\%} \colorbox{blue!20}{soft} } \makebox[22mm][l]{ {\fontsize{7}{7}\selectfont 4\%} \colorbox{blue!20}{tiny} } \\
parent & \makebox[22mm][l]{ {\fontsize{7}{7}\selectfont 5\%} \colorbox{green!20}{careful} } \makebox[22mm][l]{ {\fontsize{7}{7}\selectfont 4\%} \colorbox{red!20}{spirited} } \makebox[22mm][l]{ {\fontsize{7}{7}\selectfont 4\%} \colorbox{gray!20}{magical} } \makebox[22mm][l]{ {\fontsize{7}{7}\selectfont 4\%} \colorbox{gray!20}{connected} } \makebox[22mm][l]{ {\fontsize{7}{7}\selectfont 3\%} \colorbox{green!20}{understanding} } \makebox[22mm][l]{ {\fontsize{7}{7}\selectfont 3\%} \colorbox{red!20}{grateful} } \\

\bottomrule\\

\end{tabular}
\vspace{-6mm}
\caption{Top words in \textbf{children attributes} in the generated story that correlate (Pearson) with the target variable (e.g., child gender or parent's nationality). Obvious variables that correspond to the input (such as \textit{boy}) are removed. \textbf{Color codes:} \colorbox{blue!20}{blue} = physical; \colorbox{red!20}{red} = emotional; \colorbox{green!20}{green} = mental; \colorbox{yellow!20}{yellow} = moral; \colorbox{gray!20}{gray} = other.}

\label{tab:corr_bias_attr}
\vspace{-5mm}
\end{table*}

\begin{table*}[htbp]
\fontsize{8}{8}\selectfont
\begin{tabular}{p{2cm}l}
\toprule \\[-0.5em]
\null\\[-1.7em]
\multicolumn{2}{l}{\bf Gender} \\

child & \makebox[22mm][l]{ {\fontsize{7}{7}\selectfont 13\%} fur } \makebox[22mm][l]{ {\fontsize{7}{7}\selectfont 10\%} little } \makebox[22mm][l]{ {\fontsize{7}{7}\selectfont 9\%} grateful } \makebox[22mm][l]{ {\fontsize{7}{7}\selectfont 9\%} playful } \makebox[22mm][l]{ {\fontsize{7}{7}\selectfont 6\%} wise } { {\fontsize{7}{7}\selectfont 5\%} nature } \\
daughter & \makebox[22mm][l]{ {\fontsize{7}{7}\selectfont 19\%} smile } \makebox[22mm][l]{ {\fontsize{7}{7}\selectfont 18\%} hair } \makebox[22mm][l]{ {\fontsize{7}{7}\selectfont 18\%} eyes } \makebox[22mm][l]{ {\fontsize{7}{7}\selectfont 16\%} heart } \makebox[22mm][l]{ {\fontsize{7}{7}\selectfont 15\%} light } { {\fontsize{7}{7}\selectfont 15\%} golden } \\
son & \makebox[22mm][l]{ {\fontsize{7}{7}\selectfont 19\%} adventurous } \makebox[22mm][l]{ {\fontsize{7}{7}\selectfont 16\%} curious } \makebox[22mm][l]{ {\fontsize{7}{7}\selectfont 14\%} brave } \makebox[22mm][l]{ {\fontsize{7}{7}\selectfont 11\%} eager } \makebox[22mm][l]{ {\fontsize{7}{7}\selectfont 10\%} courageous } { {\fontsize{7}{7}\selectfont 9\%} determined } \\
\bottomrule\\

\null\\[-1.7em]
\multicolumn{2}{l}{\bf Nationality Parent Group} \\

Africa & \makebox[22mm][l]{ {\fontsize{7}{7}\selectfont 8\%} brave } \makebox[22mm][l]{ {\fontsize{7}{7}\selectfont 6\%} adventurous } \makebox[22mm][l]{ {\fontsize{7}{7}\selectfont 5\%} curiosity } \makebox[22mm][l]{ {\fontsize{7}{7}\selectfont 5\%} big } \makebox[22mm][l]{ {\fontsize{7}{7}\selectfont 5\%} imaginative } { {\fontsize{7}{7}\selectfont 4\%} eager } \\
Asia & \makebox[22mm][l]{ {\fontsize{7}{7}\selectfont 9\%} playful } \makebox[22mm][l]{ {\fontsize{7}{7}\selectfont 8\%} content } \makebox[22mm][l]{ {\fontsize{7}{7}\selectfont 8\%} adventurous } \makebox[22mm][l]{ {\fontsize{7}{7}\selectfont 8\%} warmth } \makebox[22mm][l]{ {\fontsize{7}{7}\selectfont 7\%} joyful } { {\fontsize{7}{7}\selectfont 7\%} grateful } \\
European & \makebox[22mm][l]{ {\fontsize{7}{7}\selectfont 15\%} blue } \makebox[22mm][l]{ {\fontsize{7}{7}\selectfont 15\%} hair } \makebox[22mm][l]{ {\fontsize{7}{7}\selectfont 14\%} golden } \makebox[22mm][l]{ {\fontsize{7}{7}\selectfont 13\%} faith } \makebox[22mm][l]{ {\fontsize{7}{7}\selectfont 11\%} bright } { {\fontsize{7}{7}\selectfont 8\%} peaceful } \\
Middle Eastern & \makebox[22mm][l]{ {\fontsize{7}{7}\selectfont 7\%} young } \makebox[22mm][l]{ {\fontsize{7}{7}\selectfont 7\%} imaginative } \makebox[22mm][l]{ {\fontsize{7}{7}\selectfont 6\%} brave } \makebox[22mm][l]{ {\fontsize{7}{7}\selectfont 6\%} big } \makebox[22mm][l]{ {\fontsize{7}{7}\selectfont 6\%} determined } { {\fontsize{7}{7}\selectfont 5\%} courageous } \\
North American & \makebox[22mm][l]{ {\fontsize{7}{7}\selectfont 10\%} compassionate } \makebox[22mm][l]{ {\fontsize{7}{7}\selectfont 9\%} fur } \makebox[22mm][l]{ {\fontsize{7}{7}\selectfont 9\%} nature } \makebox[22mm][l]{ {\fontsize{7}{7}\selectfont 7\%} learn } \makebox[22mm][l]{ {\fontsize{7}{7}\selectfont 6\%} warm } { {\fontsize{7}{7}\selectfont 6\%} helpful } \\
South American & \makebox[22mm][l]{ {\fontsize{7}{7}\selectfont 9\%} adventurous } \makebox[22mm][l]{ {\fontsize{7}{7}\selectfont 8\%} brown } \makebox[22mm][l]{ {\fontsize{7}{7}\selectfont 7\%} playful } \makebox[22mm][l]{ {\fontsize{7}{7}\selectfont 6\%} mischievous } \makebox[22mm][l]{ {\fontsize{7}{7}\selectfont 6\%} curious } { {\fontsize{7}{7}\selectfont 5\%} little } \\
\bottomrule\\

\null\\[-1.7em]
\multicolumn{2}{l}{\bf Nationality Parent Developed} \\

Developed & \makebox[22mm][l]{ {\fontsize{7}{7}\selectfont 14\%} blue } \makebox[22mm][l]{ {\fontsize{7}{7}\selectfont 13\%} faith } \makebox[22mm][l]{ {\fontsize{7}{7}\selectfont 12\%} compassionate } \makebox[22mm][l]{ {\fontsize{7}{7}\selectfont 12\%} hair } \makebox[22mm][l]{ {\fontsize{7}{7}\selectfont 11\%} golden } { {\fontsize{7}{7}\selectfont 11\%} gentle } \\
Developing & \makebox[22mm][l]{ {\fontsize{7}{7}\selectfont 14\%} adventurous } \makebox[22mm][l]{ {\fontsize{7}{7}\selectfont 13\%} imaginative } \makebox[22mm][l]{ {\fontsize{7}{7}\selectfont 9\%} determined } \makebox[22mm][l]{ {\fontsize{7}{7}\selectfont 6\%} excitement } \makebox[22mm][l]{ {\fontsize{7}{7}\selectfont 6\%} brave } { {\fontsize{7}{7}\selectfont 6\%} big } \\
\bottomrule\\

\null\\[-1.7em]
\multicolumn{2}{l}{\bf Ethnicity} \\

African-Amer. & \makebox[22mm][l]{ {\fontsize{7}{7}\selectfont 54\%} strong } \makebox[22mm][l]{ {\fontsize{7}{7}\selectfont 36\%} resilient } \makebox[22mm][l]{ {\fontsize{7}{7}\selectfont 33\%} courageous } \makebox[22mm][l]{ {\fontsize{7}{7}\selectfont 31\%} confident } \makebox[22mm][l]{ {\fontsize{7}{7}\selectfont 30\%} beautiful } { {\fontsize{7}{7}\selectfont 24\%} dark } \\
Asian & \makebox[22mm][l]{ {\fontsize{7}{7}\selectfont 18\%} peaceful } \makebox[22mm][l]{ {\fontsize{7}{7}\selectfont 13\%} wonder } \makebox[22mm][l]{ {\fontsize{7}{7}\selectfont 11\%} gentle } \makebox[22mm][l]{ {\fontsize{7}{7}\selectfont 10\%} sense } \makebox[22mm][l]{ {\fontsize{7}{7}\selectfont 9\%} unwavering } { {\fontsize{7}{7}\selectfont 9\%} happy } \\
European-Amer. & \makebox[22mm][l]{ {\fontsize{7}{7}\selectfont 21\%} long } \makebox[22mm][l]{ {\fontsize{7}{7}\selectfont 19\%} flowing } \makebox[22mm][l]{ {\fontsize{7}{7}\selectfont 18\%} blue } \makebox[22mm][l]{ {\fontsize{7}{7}\selectfont 16\%} gift } \makebox[22mm][l]{ {\fontsize{7}{7}\selectfont 14\%} golden } { {\fontsize{7}{7}\selectfont 13\%} joy } \\
Latino & \makebox[22mm][l]{ {\fontsize{7}{7}\selectfont 21\%} magic } \makebox[22mm][l]{ {\fontsize{7}{7}\selectfont 19\%} excited } \makebox[22mm][l]{ {\fontsize{7}{7}\selectfont 18\%} brown } \makebox[22mm][l]{ {\fontsize{7}{7}\selectfont 16\%} curly } \makebox[22mm][l]{ {\fontsize{7}{7}\selectfont 16\%} like } { {\fontsize{7}{7}\selectfont 16\%} bright } \\
Middle-Eastern & \makebox[22mm][l]{ {\fontsize{7}{7}\selectfont 16\%} intently } \makebox[22mm][l]{ {\fontsize{7}{7}\selectfont 15\%} spirit } \makebox[22mm][l]{ {\fontsize{7}{7}\selectfont 14\%} kindness } \makebox[22mm][l]{ {\fontsize{7}{7}\selectfont 10\%} heart } \makebox[22mm][l]{ {\fontsize{7}{7}\selectfont 10\%} pure } { {\fontsize{7}{7}\selectfont 10\%} adventure } \\
White & \makebox[22mm][l]{ {\fontsize{7}{7}\selectfont 26\%} sparkling } \makebox[22mm][l]{ {\fontsize{7}{7}\selectfont 21\%} helpful } \makebox[22mm][l]{ {\fontsize{7}{7}\selectfont 18\%} golden } \makebox[22mm][l]{ {\fontsize{7}{7}\selectfont 18\%} blue } \makebox[22mm][l]{ {\fontsize{7}{7}\selectfont 14\%} gentle } { {\fontsize{7}{7}\selectfont 13\%} white } \\
\bottomrule\\

\null\\[-1.7em]
\multicolumn{2}{l}{\bf Religion} \\

Atheist & \makebox[22mm][l]{ {\fontsize{7}{7}\selectfont 21\%} open } \makebox[22mm][l]{ {\fontsize{7}{7}\selectfont 17\%} understanding } \makebox[22mm][l]{ {\fontsize{7}{7}\selectfont 16\%} awe } \makebox[22mm][l]{ {\fontsize{7}{7}\selectfont 13\%} sense } \makebox[22mm][l]{ {\fontsize{7}{7}\selectfont 12\%} learn } { {\fontsize{7}{7}\selectfont 10\%} curious } \\
Buddhist & \makebox[22mm][l]{ {\fontsize{7}{7}\selectfont 30\%} understanding } \makebox[22mm][l]{ {\fontsize{7}{7}\selectfont 25\%} compassionate } \makebox[22mm][l]{ {\fontsize{7}{7}\selectfont 18\%} gentle } \makebox[22mm][l]{ {\fontsize{7}{7}\selectfont 17\%} nature } \makebox[22mm][l]{ {\fontsize{7}{7}\selectfont 11\%} loves } { {\fontsize{7}{7}\selectfont 10\%} peaceful } \\
Christian & \makebox[22mm][l]{ {\fontsize{7}{7}\selectfont 10\%} joyful } \makebox[22mm][l]{ {\fontsize{7}{7}\selectfont 9\%} faith } \makebox[22mm][l]{ {\fontsize{7}{7}\selectfont 9\%} eye } \makebox[22mm][l]{ {\fontsize{7}{7}\selectfont 9\%} kindness } \makebox[22mm][l]{ {\fontsize{7}{7}\selectfont 8\%} determined } { {\fontsize{7}{7}\selectfont 8\%} heart } \\
Hindu & \makebox[22mm][l]{ {\fontsize{7}{7}\selectfont 11\%} brave } \makebox[22mm][l]{ {\fontsize{7}{7}\selectfont 10\%} wise } \makebox[22mm][l]{ {\fontsize{7}{7}\selectfont 8\%} courageous } \makebox[22mm][l]{ {\fontsize{7}{7}\selectfont 4\%} strong } \makebox[22mm][l]{ {\fontsize{7}{7}\selectfont 4\%} selfless } { {\fontsize{7}{7}\selectfont 4\%} pure } \\
Jew & \makebox[22mm][l]{ {\fontsize{7}{7}\selectfont 15\%} learn } \makebox[22mm][l]{ {\fontsize{7}{7}\selectfont 7\%} compassion } \makebox[22mm][l]{ {\fontsize{7}{7}\selectfont 6\%} kind } \makebox[22mm][l]{ {\fontsize{7}{7}\selectfont 6\%} kindness } \makebox[22mm][l]{ {\fontsize{7}{7}\selectfont 6\%} eager } { {\fontsize{7}{7}\selectfont 6\%} loving } \\
Muslim & \makebox[22mm][l]{ {\fontsize{7}{7}\selectfont 7\%} imaginative } \makebox[22mm][l]{ {\fontsize{7}{7}\selectfont 7\%} sleepy } \makebox[22mm][l]{ {\fontsize{7}{7}\selectfont 6\%} adventurous } \makebox[22mm][l]{ {\fontsize{7}{7}\selectfont 4\%} peaceful } \makebox[22mm][l]{ {\fontsize{7}{7}\selectfont 4\%} loved } { {\fontsize{7}{7}\selectfont 4\%} comforted } \\
\bottomrule\\

\null\\[-1.7em]
\multicolumn{2}{l}{\bf Role} \\

father & \makebox[22mm][l]{ {\fontsize{7}{7}\selectfont 11\%} brave } \makebox[22mm][l]{ {\fontsize{7}{7}\selectfont 10\%} wise } \makebox[22mm][l]{ {\fontsize{7}{7}\selectfont 8\%} courageous } \makebox[22mm][l]{ {\fontsize{7}{7}\selectfont 4\%} strong } \makebox[22mm][l]{ {\fontsize{7}{7}\selectfont 4\%} selfless } { {\fontsize{7}{7}\selectfont 4\%} pure } \\
mother & \makebox[22mm][l]{ {\fontsize{7}{7}\selectfont 7\%} imaginative } \makebox[22mm][l]{ {\fontsize{7}{7}\selectfont 7\%} sleepy } \makebox[22mm][l]{ {\fontsize{7}{7}\selectfont 6\%} adventurous } \makebox[22mm][l]{ {\fontsize{7}{7}\selectfont 4\%} peaceful } \makebox[22mm][l]{ {\fontsize{7}{7}\selectfont 4\%} loved } { {\fontsize{7}{7}\selectfont 4\%} comforted } \\
parent & \makebox[22mm][l]{ {\fontsize{7}{7}\selectfont 10\%} joyful } \makebox[22mm][l]{ {\fontsize{7}{7}\selectfont 9\%} faith } \makebox[22mm][l]{ {\fontsize{7}{7}\selectfont 9\%} eye } \makebox[22mm][l]{ {\fontsize{7}{7}\selectfont 9\%} kindness } \makebox[22mm][l]{ {\fontsize{7}{7}\selectfont 8\%} determined } { {\fontsize{7}{7}\selectfont 8\%} heart } \\
\bottomrule\\

\end{tabular}
\vspace{-6mm}
\caption{Top words in the \textbf{Character-Centric Attributes by GPT-4o} that correlate (Pearson) with the sociocultural factor. The terms \textit{child}, \textit{daughter}, and \textit{son} have been removed, as they are almost present at the start of the generation}
\label{tab:corr_bias_char-gpt4}
\end{table*}

\begin{table*}[htbp]
\fontsize{8}{8}\selectfont
\begin{tabular}{p{2cm}l}
\toprule \\[-0.5em]
\null\\[-1.7em]
\multicolumn{2}{l}{\bf Gender} \\

child & \makebox[22mm][l]{ {\fontsize{7}{7}\selectfont 11\%} little } \makebox[22mm][l]{ {\fontsize{7}{7}\selectfont 7\%} explore } \makebox[22mm][l]{ {\fontsize{7}{7}\selectfont 6\%} grateful } \makebox[22mm][l]{ {\fontsize{7}{7}\selectfont 6\%} loved } \makebox[22mm][l]{ {\fontsize{7}{7}\selectfont 5\%} helpful } { {\fontsize{7}{7}\selectfont 5\%} friend } \\
daughter & \makebox[22mm][l]{ {\fontsize{7}{7}\selectfont 26\%} gentle } \makebox[22mm][l]{ {\fontsize{7}{7}\selectfont 26\%} hair } \makebox[22mm][l]{ {\fontsize{7}{7}\selectfont 19\%} kind } \makebox[22mm][l]{ {\fontsize{7}{7}\selectfont 17\%} black } \makebox[22mm][l]{ {\fontsize{7}{7}\selectfont 16\%} beautiful } { {\fontsize{7}{7}\selectfont 16\%} bright } \\
son & \makebox[22mm][l]{ {\fontsize{7}{7}\selectfont 24\%} brave } \makebox[22mm][l]{ {\fontsize{7}{7}\selectfont 15\%} young } \makebox[22mm][l]{ {\fontsize{7}{7}\selectfont 13\%} adventurous } \makebox[22mm][l]{ {\fontsize{7}{7}\selectfont 11\%} courageous } \makebox[22mm][l]{ {\fontsize{7}{7}\selectfont 10\%} curious } { {\fontsize{7}{7}\selectfont 10\%} determined } \\
\bottomrule\\

\null\\[-1.7em]
\multicolumn{2}{l}{\bf Nationality Parent Group} \\

Africa & \makebox[22mm][l]{ {\fontsize{7}{7}\selectfont 12\%} land } \makebox[22mm][l]{ {\fontsize{7}{7}\selectfont 11\%} bright } \makebox[22mm][l]{ {\fontsize{7}{7}\selectfont 9\%} secret } \makebox[22mm][l]{ {\fontsize{7}{7}\selectfont 8\%} respectful } \makebox[22mm][l]{ {\fontsize{7}{7}\selectfont 7\%} young } { {\fontsize{7}{7}\selectfont 7\%} black } \\
Asia & \makebox[22mm][l]{ {\fontsize{7}{7}\selectfont 12\%} kind } \makebox[22mm][l]{ {\fontsize{7}{7}\selectfont 9\%} spreads } \makebox[22mm][l]{ {\fontsize{7}{7}\selectfont 8\%} black } \makebox[22mm][l]{ {\fontsize{7}{7}\selectfont 7\%} kindness } \makebox[22mm][l]{ {\fontsize{7}{7}\selectfont 6\%} hearted } { {\fontsize{7}{7}\selectfont 6\%} soft } \\
European & \makebox[22mm][l]{ {\fontsize{7}{7}\selectfont 17\%} blue } \makebox[22mm][l]{ {\fontsize{7}{7}\selectfont 8\%} hair } \makebox[22mm][l]{ {\fontsize{7}{7}\selectfont 7\%} beautiful } \makebox[22mm][l]{ {\fontsize{7}{7}\selectfont 7\%} peace } \makebox[22mm][l]{ {\fontsize{7}{7}\selectfont 6\%} brown } { {\fontsize{7}{7}\selectfont 6\%} compassionate } \\
Middle Eastern & \makebox[22mm][l]{ {\fontsize{7}{7}\selectfont 12\%} eyed } \makebox[22mm][l]{ {\fontsize{7}{7}\selectfont 8\%} young } \makebox[22mm][l]{ {\fontsize{7}{7}\selectfont 6\%} courageous } \makebox[22mm][l]{ {\fontsize{7}{7}\selectfont 5\%} imaginative } \makebox[22mm][l]{ {\fontsize{7}{7}\selectfont 5\%} brave } { {\fontsize{7}{7}\selectfont 5\%} fascinated } \\
North American & \makebox[22mm][l]{ {\fontsize{7}{7}\selectfont 10\%} loving } \makebox[22mm][l]{ {\fontsize{7}{7}\selectfont 9\%} grateful } \makebox[22mm][l]{ {\fontsize{7}{7}\selectfont 8\%} understanding } \makebox[22mm][l]{ {\fontsize{7}{7}\selectfont 8\%} explore } \makebox[22mm][l]{ {\fontsize{7}{7}\selectfont 7\%} friend } { {\fontsize{7}{7}\selectfont 7\%} peaceful } \\
South American & \makebox[22mm][l]{ {\fontsize{7}{7}\selectfont 10\%} little } \makebox[22mm][l]{ {\fontsize{7}{7}\selectfont 7\%} black } \makebox[22mm][l]{ {\fontsize{7}{7}\selectfont 7\%} thrilled } \makebox[22mm][l]{ {\fontsize{7}{7}\selectfont 4\%} friend } \makebox[22mm][l]{ {\fontsize{7}{7}\selectfont 4\%} guided } { {\fontsize{7}{7}\selectfont 4\%} playful } \\
\bottomrule\\

\null\\[-1.7em]
\multicolumn{2}{l}{\bf Nationality Parent Developed} \\

Developed & \makebox[22mm][l]{ {\fontsize{7}{7}\selectfont 13\%} blue } \makebox[22mm][l]{ {\fontsize{7}{7}\selectfont 12\%} peaceful } \makebox[22mm][l]{ {\fontsize{7}{7}\selectfont 11\%} grateful } \makebox[22mm][l]{ {\fontsize{7}{7}\selectfont 9\%} loving } \makebox[22mm][l]{ {\fontsize{7}{7}\selectfont 8\%} understanding } { {\fontsize{7}{7}\selectfont 8\%} peace } \\
Developing & \makebox[22mm][l]{ {\fontsize{7}{7}\selectfont 17\%} black } \makebox[22mm][l]{ {\fontsize{7}{7}\selectfont 8\%} eyed } \makebox[22mm][l]{ {\fontsize{7}{7}\selectfont 8\%} hearted } \makebox[22mm][l]{ {\fontsize{7}{7}\selectfont 8\%} story } \makebox[22mm][l]{ {\fontsize{7}{7}\selectfont 8\%} young } { {\fontsize{7}{7}\selectfont 7\%} brave } \\
\bottomrule\\

\null\\[-1.7em]
\multicolumn{2}{l}{\bf Ethnicity} \\

African-Amer. & \makebox[22mm][l]{ {\fontsize{7}{7}\selectfont 46\%} strong } \makebox[22mm][l]{ {\fontsize{7}{7}\selectfont 34\%} rich } \makebox[22mm][l]{ {\fontsize{7}{7}\selectfont 33\%} skin } \makebox[22mm][l]{ {\fontsize{7}{7}\selectfont 30\%} resilient } \makebox[22mm][l]{ {\fontsize{7}{7}\selectfont 24\%} purpose } { {\fontsize{7}{7}\selectfont 22\%} proud } \\
Asian & \makebox[22mm][l]{ {\fontsize{7}{7}\selectfont 28\%} compassionate } \makebox[22mm][l]{ {\fontsize{7}{7}\selectfont 26\%} kind } \makebox[22mm][l]{ {\fontsize{7}{7}\selectfont 24\%} respectful } \makebox[22mm][l]{ {\fontsize{7}{7}\selectfont 17\%} gentle } \makebox[22mm][l]{ {\fontsize{7}{7}\selectfont 12\%} eager } { {\fontsize{7}{7}\selectfont 9\%} wise } \\
European-Amer. & \makebox[22mm][l]{ {\fontsize{7}{7}\selectfont 25\%} blue } \makebox[22mm][l]{ {\fontsize{7}{7}\selectfont 18\%} eye } \makebox[22mm][l]{ {\fontsize{7}{7}\selectfont 17\%} brown } \makebox[22mm][l]{ {\fontsize{7}{7}\selectfont 15\%} magical } \makebox[22mm][l]{ {\fontsize{7}{7}\selectfont 15\%} determined } { {\fontsize{7}{7}\selectfont 14\%} excited } \\
Latino & \makebox[22mm][l]{ {\fontsize{7}{7}\selectfont 26\%} big } \makebox[22mm][l]{ {\fontsize{7}{7}\selectfont 25\%} heritage } \makebox[22mm][l]{ {\fontsize{7}{7}\selectfont 24\%} felt } \makebox[22mm][l]{ {\fontsize{7}{7}\selectfont 23\%} black } \makebox[22mm][l]{ {\fontsize{7}{7}\selectfont 23\%} mop } { {\fontsize{7}{7}\selectfont 22\%} magic } \\
Middle-Eastern & \makebox[22mm][l]{ {\fontsize{7}{7}\selectfont 26\%} eyed } \makebox[22mm][l]{ {\fontsize{7}{7}\selectfont 23\%} eyes } \makebox[22mm][l]{ {\fontsize{7}{7}\selectfont 17\%} wise } \makebox[22mm][l]{ {\fontsize{7}{7}\selectfont 17\%} black } \makebox[22mm][l]{ {\fontsize{7}{7}\selectfont 16\%} sky } { {\fontsize{7}{7}\selectfont 15\%} bright } \\
White & \makebox[22mm][l]{ {\fontsize{7}{7}\selectfont 19\%} loving } \makebox[22mm][l]{ {\fontsize{7}{7}\selectfont 18\%} kind } \makebox[22mm][l]{ {\fontsize{7}{7}\selectfont 18\%} grateful } \makebox[22mm][l]{ {\fontsize{7}{7}\selectfont 14\%} gift } \makebox[22mm][l]{ {\fontsize{7}{7}\selectfont 13\%} explore } { {\fontsize{7}{7}\selectfont 13\%} loved } \\
\bottomrule\\

\null\\[-1.7em]
\multicolumn{2}{l}{\bf Religion} \\

Atheist & \makebox[22mm][l]{ {\fontsize{7}{7}\selectfont 13\%} connection } \makebox[22mm][l]{ {\fontsize{7}{7}\selectfont 11\%} sense } \makebox[22mm][l]{ {\fontsize{7}{7}\selectfont 10\%} awe } \makebox[22mm][l]{ {\fontsize{7}{7}\selectfont 10\%} feels } \makebox[22mm][l]{ {\fontsize{7}{7}\selectfont 9\%} star } { {\fontsize{7}{7}\selectfont 9\%} shining } \\
Buddhist & \makebox[22mm][l]{ {\fontsize{7}{7}\selectfont 31\%} peaceful } \makebox[22mm][l]{ {\fontsize{7}{7}\selectfont 24\%} understanding } \makebox[22mm][l]{ {\fontsize{7}{7}\selectfont 23\%} compassionate } \makebox[22mm][l]{ {\fontsize{7}{7}\selectfont 15\%} loving } \makebox[22mm][l]{ {\fontsize{7}{7}\selectfont 14\%} grateful } { {\fontsize{7}{7}\selectfont 8\%} beautiful } \\
Christian & \makebox[22mm][l]{ {\fontsize{7}{7}\selectfont 5\%} clever } \makebox[22mm][l]{ {\fontsize{7}{7}\selectfont 5\%} curious } \makebox[22mm][l]{ {\fontsize{7}{7}\selectfont 5\%} smile } \makebox[22mm][l]{ {\fontsize{7}{7}\selectfont 4\%} adventure } \makebox[22mm][l]{ {\fontsize{7}{7}\selectfont 4\%} courage } { {\fontsize{7}{7}\selectfont 4\%} amazed } \\
Hindu & \makebox[22mm][l]{ {\fontsize{7}{7}\selectfont 15\%} brave } \makebox[22mm][l]{ {\fontsize{7}{7}\selectfont 8\%} determined } \makebox[22mm][l]{ {\fontsize{7}{7}\selectfont 7\%} adventurous } \makebox[22mm][l]{ {\fontsize{7}{7}\selectfont 5\%} strong } \makebox[22mm][l]{ {\fontsize{7}{7}\selectfont 5\%} bright } { {\fontsize{7}{7}\selectfont 4\%} young } \\
Jew & \makebox[22mm][l]{ {\fontsize{7}{7}\selectfont 8\%} warm } \makebox[22mm][l]{ {\fontsize{7}{7}\selectfont 7\%} connected } \makebox[22mm][l]{ {\fontsize{7}{7}\selectfont 6\%} felt } \makebox[22mm][l]{ {\fontsize{7}{7}\selectfont 6\%} loving } \makebox[22mm][l]{ {\fontsize{7}{7}\selectfont 6\%} wide } { {\fontsize{7}{7}\selectfont 6\%} wise } \\
Muslim & \makebox[22mm][l]{ {\fontsize{7}{7}\selectfont 10\%} soft } \makebox[22mm][l]{ {\fontsize{7}{7}\selectfont 6\%} big } \makebox[22mm][l]{ {\fontsize{7}{7}\selectfont 5\%} loved } \makebox[22mm][l]{ {\fontsize{7}{7}\selectfont 5\%} special } \makebox[22mm][l]{ {\fontsize{7}{7}\selectfont 5\%} little } { {\fontsize{7}{7}\selectfont 4\%} joyful } \\
\bottomrule\\

\null\\[-1.7em]
\multicolumn{2}{l}{\bf Role} \\

father & \makebox[22mm][l]{ {\fontsize{7}{7}\selectfont 15\%} brave } \makebox[22mm][l]{ {\fontsize{7}{7}\selectfont 8\%} determined } \makebox[22mm][l]{ {\fontsize{7}{7}\selectfont 7\%} adventurous } \makebox[22mm][l]{ {\fontsize{7}{7}\selectfont 5\%} strong } \makebox[22mm][l]{ {\fontsize{7}{7}\selectfont 5\%} bright } { {\fontsize{7}{7}\selectfont 4\%} young } \\
mother & \makebox[22mm][l]{ {\fontsize{7}{7}\selectfont 10\%} soft } \makebox[22mm][l]{ {\fontsize{7}{7}\selectfont 6\%} big } \makebox[22mm][l]{ {\fontsize{7}{7}\selectfont 5\%} loved } \makebox[22mm][l]{ {\fontsize{7}{7}\selectfont 5\%} special } \makebox[22mm][l]{ {\fontsize{7}{7}\selectfont 5\%} little } { {\fontsize{7}{7}\selectfont 4\%} joyful } \\
parent & \makebox[22mm][l]{ {\fontsize{7}{7}\selectfont 5\%} clever } \makebox[22mm][l]{ {\fontsize{7}{7}\selectfont 5\%} curious } \makebox[22mm][l]{ {\fontsize{7}{7}\selectfont 5\%} smile } \makebox[22mm][l]{ {\fontsize{7}{7}\selectfont 4\%} adventure } \makebox[22mm][l]{ {\fontsize{7}{7}\selectfont 4\%} courage } { {\fontsize{7}{7}\selectfont 4\%} amazed } \\
\bottomrule\\

\end{tabular}
\vspace{-6mm}
\caption{Top words in the \textbf{Character-Centric Attributes by Llama3} that correlate (Pearson) with the sociocultural factor. The terms \textit{child}, \textit{daughter}, and \textit{son} have been removed, as they are almost present at the start of the generation}
\label{tab:corr_bias_char_llama3}
\end{table*}

\begin{table*}[htbp]
\fontsize{8}{8}\selectfont
\begin{tabular}{p{2cm}l}
\toprule \\[-0.5em]
\null\\[-1.7em]
\multicolumn{2}{l}{\bf Gender} \\

child & \makebox[22mm][l]{ {\fontsize{7}{7}\selectfont 6\%} gentle } \makebox[22mm][l]{ {\fontsize{7}{7}\selectfont 5\%} friend } \makebox[22mm][l]{ {\fontsize{7}{7}\selectfont 5\%} knowledgeable } \makebox[22mm][l]{ {\fontsize{7}{7}\selectfont 5\%} wise } \makebox[22mm][l]{ {\fontsize{7}{7}\selectfont 5\%} hardworking } { {\fontsize{7}{7}\selectfont 5\%} happy } \\
daughter & \makebox[22mm][l]{ {\fontsize{7}{7}\selectfont 21\%} beautiful } \makebox[22mm][l]{ {\fontsize{7}{7}\selectfont 12\%} hair } \makebox[22mm][l]{ {\fontsize{7}{7}\selectfont 10\%} eye } \makebox[22mm][l]{ {\fontsize{7}{7}\selectfont 9\%} hearted } \makebox[22mm][l]{ {\fontsize{7}{7}\selectfont 9\%} loving } { {\fontsize{7}{7}\selectfont 9\%} smile } \\
son & \makebox[22mm][l]{ {\fontsize{7}{7}\selectfont 9\%} curious } \makebox[22mm][l]{ {\fontsize{7}{7}\selectfont 9\%} young } \makebox[22mm][l]{ {\fontsize{7}{7}\selectfont 8\%} respected } \makebox[22mm][l]{ {\fontsize{7}{7}\selectfont 8\%} respectful } \makebox[22mm][l]{ {\fontsize{7}{7}\selectfont 6\%} brave } { {\fontsize{7}{7}\selectfont 5\%} thoughtful } \\
\bottomrule\\

\null\\[-1.7em]
\multicolumn{2}{l}{\bf Nationality Parent Group} \\

Africa & \makebox[22mm][l]{ {\fontsize{7}{7}\selectfont 7\%} heart } \makebox[22mm][l]{ {\fontsize{7}{7}\selectfont 7\%} kindness } \makebox[22mm][l]{ {\fontsize{7}{7}\selectfont 6\%} determined } \makebox[22mm][l]{ {\fontsize{7}{7}\selectfont 5\%} strong } \makebox[22mm][l]{ {\fontsize{7}{7}\selectfont 5\%} wisdom } { {\fontsize{7}{7}\selectfont 5\%} adventurous } \\
Asia & \makebox[22mm][l]{ {\fontsize{7}{7}\selectfont 10\%} hardworking } \makebox[22mm][l]{ {\fontsize{7}{7}\selectfont 9\%} respectful } \makebox[22mm][l]{ {\fontsize{7}{7}\selectfont 9\%} kind } \makebox[22mm][l]{ {\fontsize{7}{7}\selectfont 8\%} hearted } \makebox[22mm][l]{ {\fontsize{7}{7}\selectfont 8\%} grateful } { {\fontsize{7}{7}\selectfont 8\%} fascinated } \\
European & \makebox[22mm][l]{ {\fontsize{7}{7}\selectfont 11\%} gentle } \makebox[22mm][l]{ {\fontsize{7}{7}\selectfont 11\%} hair } \makebox[22mm][l]{ {\fontsize{7}{7}\selectfont 10\%} eye } \makebox[22mm][l]{ {\fontsize{7}{7}\selectfont 10\%} golden } \makebox[22mm][l]{ {\fontsize{7}{7}\selectfont 9\%} long } { {\fontsize{7}{7}\selectfont 9\%} faithful } \\
Middle Eastern & \makebox[22mm][l]{ {\fontsize{7}{7}\selectfont 9\%} adventurous } \makebox[22mm][l]{ {\fontsize{7}{7}\selectfont 8\%} just } \makebox[22mm][l]{ {\fontsize{7}{7}\selectfont 7\%} love } \makebox[22mm][l]{ {\fontsize{7}{7}\selectfont 6\%} courageous } \makebox[22mm][l]{ {\fontsize{7}{7}\selectfont 6\%} brave } { {\fontsize{7}{7}\selectfont 5\%} dedicated } \\
North American & \makebox[22mm][l]{ {\fontsize{7}{7}\selectfont 8\%} explore } \makebox[22mm][l]{ {\fontsize{7}{7}\selectfont 8\%} young } \makebox[22mm][l]{ {\fontsize{7}{7}\selectfont 7\%} sweet } \makebox[22mm][l]{ {\fontsize{7}{7}\selectfont 6\%} patient } \makebox[22mm][l]{ {\fontsize{7}{7}\selectfont 6\%} gift } { {\fontsize{7}{7}\selectfont 5\%} loving } \\
South American & \makebox[22mm][l]{ {\fontsize{7}{7}\selectfont 10\%} curious } \makebox[22mm][l]{ {\fontsize{7}{7}\selectfont 8\%} inspiring } \makebox[22mm][l]{ {\fontsize{7}{7}\selectfont 7\%} boundless } \makebox[22mm][l]{ {\fontsize{7}{7}\selectfont 6\%} deep } \makebox[22mm][l]{ {\fontsize{7}{7}\selectfont 6\%} enchanted } { {\fontsize{7}{7}\selectfont 5\%} gift } \\
\bottomrule\\

\null\\[-1.7em]
\multicolumn{2}{l}{\bf Nationality Parent Developed} \\

Developed & \makebox[22mm][l]{ {\fontsize{7}{7}\selectfont 12\%} hair } \makebox[22mm][l]{ {\fontsize{7}{7}\selectfont 11\%} golden } \makebox[22mm][l]{ {\fontsize{7}{7}\selectfont 10\%} faithful } \makebox[22mm][l]{ {\fontsize{7}{7}\selectfont 10\%} blue } \makebox[22mm][l]{ {\fontsize{7}{7}\selectfont 10\%} eye } { {\fontsize{7}{7}\selectfont 10\%} gentle } \\
Developing & \makebox[22mm][l]{ {\fontsize{7}{7}\selectfont 14\%} curious } \makebox[22mm][l]{ {\fontsize{7}{7}\selectfont 12\%} adventurous } \makebox[22mm][l]{ {\fontsize{7}{7}\selectfont 9\%} hearted } \makebox[22mm][l]{ {\fontsize{7}{7}\selectfont 8\%} fascinated } \makebox[22mm][l]{ {\fontsize{7}{7}\selectfont 8\%} brave } { {\fontsize{7}{7}\selectfont 7\%} animal } \\
\bottomrule\\

\null\\[-1.7em]
\multicolumn{2}{l}{\bf Ethnicity} \\

African-Amer. & \makebox[22mm][l]{ {\fontsize{7}{7}\selectfont 42\%} african } \makebox[22mm][l]{ {\fontsize{7}{7}\selectfont 30\%} skin } \makebox[22mm][l]{ {\fontsize{7}{7}\selectfont 24\%} radiant } \makebox[22mm][l]{ {\fontsize{7}{7}\selectfont 24\%} spirit } \makebox[22mm][l]{ {\fontsize{7}{7}\selectfont 24\%} american } { {\fontsize{7}{7}\selectfont 23\%} big } \\
Asian & \makebox[22mm][l]{ {\fontsize{7}{7}\selectfont 35\%} gentle } \makebox[22mm][l]{ {\fontsize{7}{7}\selectfont 17\%} overjoyed } \makebox[22mm][l]{ {\fontsize{7}{7}\selectfont 16\%} skilled } \makebox[22mm][l]{ {\fontsize{7}{7}\selectfont 16\%} patient } \makebox[22mm][l]{ {\fontsize{7}{7}\selectfont 14\%} helpful } { {\fontsize{7}{7}\selectfont 12\%} proud } \\
European-Amer. & \makebox[22mm][l]{ {\fontsize{7}{7}\selectfont 42\%} european } \makebox[22mm][l]{ {\fontsize{7}{7}\selectfont 34\%} half } \makebox[22mm][l]{ {\fontsize{7}{7}\selectfont 24\%} special } \makebox[22mm][l]{ {\fontsize{7}{7}\selectfont 24\%} american } \makebox[22mm][l]{ {\fontsize{7}{7}\selectfont 23\%} unique } { {\fontsize{7}{7}\selectfont 21\%} gift } \\
Latino & \makebox[22mm][l]{ {\fontsize{7}{7}\selectfont 19\%} fascinated } \makebox[22mm][l]{ {\fontsize{7}{7}\selectfont 19\%} curious } \makebox[22mm][l]{ {\fontsize{7}{7}\selectfont 17\%} excited } \makebox[22mm][l]{ {\fontsize{7}{7}\selectfont 16\%} skilled } \makebox[22mm][l]{ {\fontsize{7}{7}\selectfont 16\%} curly } { {\fontsize{7}{7}\selectfont 16\%} insatiable } \\
Middle-Eastern & \makebox[22mm][l]{ {\fontsize{7}{7}\selectfont 28\%} generous } \makebox[22mm][l]{ {\fontsize{7}{7}\selectfont 23\%} boundless } \makebox[22mm][l]{ {\fontsize{7}{7}\selectfont 21\%} selfless } \makebox[22mm][l]{ {\fontsize{7}{7}\selectfont 18\%} humble } \makebox[22mm][l]{ {\fontsize{7}{7}\selectfont 16\%} wonder } { {\fontsize{7}{7}\selectfont 13\%} knowledgeable } \\
White & \makebox[22mm][l]{ {\fontsize{7}{7}\selectfont 48\%} white } \makebox[22mm][l]{ {\fontsize{7}{7}\selectfont 34\%} blue } \makebox[22mm][l]{ {\fontsize{7}{7}\selectfont 30\%} golden } \makebox[22mm][l]{ {\fontsize{7}{7}\selectfont 28\%} hair } \makebox[22mm][l]{ {\fontsize{7}{7}\selectfont 21\%} beautiful } { {\fontsize{7}{7}\selectfont 21\%} sparkling } \\
\bottomrule\\

\null\\[-1.7em]
\multicolumn{2}{l}{\bf Religion} \\

Atheist & \makebox[22mm][l]{ {\fontsize{7}{7}\selectfont 9\%} imaginative } \makebox[22mm][l]{ {\fontsize{7}{7}\selectfont 8\%} young } \makebox[22mm][l]{ {\fontsize{7}{7}\selectfont 7\%} leader } \makebox[22mm][l]{ {\fontsize{7}{7}\selectfont 7\%} bright } \makebox[22mm][l]{ {\fontsize{7}{7}\selectfont 6\%} learning } { {\fontsize{7}{7}\selectfont 5\%} inspiring } \\
Buddhist & \makebox[22mm][l]{ {\fontsize{7}{7}\selectfont 46\%} mindful } \makebox[22mm][l]{ {\fontsize{7}{7}\selectfont 25\%} peaceful } \makebox[22mm][l]{ {\fontsize{7}{7}\selectfont 22\%} compassionate } \makebox[22mm][l]{ {\fontsize{7}{7}\selectfont 21\%} wise } \makebox[22mm][l]{ {\fontsize{7}{7}\selectfont 16\%} understanding } { {\fontsize{7}{7}\selectfont 11\%} patient } \\
Christian & \makebox[22mm][l]{ {\fontsize{7}{7}\selectfont 7\%} love } \makebox[22mm][l]{ {\fontsize{7}{7}\selectfont 6\%} nature } \makebox[22mm][l]{ {\fontsize{7}{7}\selectfont 6\%} captivated } \makebox[22mm][l]{ {\fontsize{7}{7}\selectfont 6\%} faithful } \makebox[22mm][l]{ {\fontsize{7}{7}\selectfont 5\%} loves } { {\fontsize{7}{7}\selectfont 5\%} deep } \\
Hindu & \makebox[22mm][l]{ {\fontsize{7}{7}\selectfont 9\%} just } \makebox[22mm][l]{ {\fontsize{7}{7}\selectfont 7\%} adventurous } \makebox[22mm][l]{ {\fontsize{7}{7}\selectfont 7\%} brave } \makebox[22mm][l]{ {\fontsize{7}{7}\selectfont 6\%} skilled } \makebox[22mm][l]{ {\fontsize{7}{7}\selectfont 6\%} protective } { {\fontsize{7}{7}\selectfont 5\%} strong } \\
Jew & \makebox[22mm][l]{ {\fontsize{7}{7}\selectfont 10\%} gentle } \makebox[22mm][l]{ {\fontsize{7}{7}\selectfont 10\%} humble } \makebox[22mm][l]{ {\fontsize{7}{7}\selectfont 10\%} wise } \makebox[22mm][l]{ {\fontsize{7}{7}\selectfont 9\%} learning } \makebox[22mm][l]{ {\fontsize{7}{7}\selectfont 8\%} known } { {\fontsize{7}{7}\selectfont 7\%} proud } \\
Muslim & \makebox[22mm][l]{ {\fontsize{7}{7}\selectfont 10\%} little } \makebox[22mm][l]{ {\fontsize{7}{7}\selectfont 9\%} curious } \makebox[22mm][l]{ {\fontsize{7}{7}\selectfont 7\%} kind } \makebox[22mm][l]{ {\fontsize{7}{7}\selectfont 7\%} heart } \makebox[22mm][l]{ {\fontsize{7}{7}\selectfont 6\%} joy } { {\fontsize{7}{7}\selectfont 6\%} eye } \\
\bottomrule\\

\null\\[-1.7em]
\multicolumn{2}{l}{\bf Role} \\

father & \makebox[22mm][l]{ {\fontsize{7}{7}\selectfont 9\%} just } \makebox[22mm][l]{ {\fontsize{7}{7}\selectfont 7\%} adventurous } \makebox[22mm][l]{ {\fontsize{7}{7}\selectfont 7\%} brave } \makebox[22mm][l]{ {\fontsize{7}{7}\selectfont 6\%} skilled } \makebox[22mm][l]{ {\fontsize{7}{7}\selectfont 6\%} protective } { {\fontsize{7}{7}\selectfont 5\%} strong } \\
mother & \makebox[22mm][l]{ {\fontsize{7}{7}\selectfont 10\%} little } \makebox[22mm][l]{ {\fontsize{7}{7}\selectfont 9\%} curious } \makebox[22mm][l]{ {\fontsize{7}{7}\selectfont 7\%} kind } \makebox[22mm][l]{ {\fontsize{7}{7}\selectfont 7\%} heart } \makebox[22mm][l]{ {\fontsize{7}{7}\selectfont 6\%} joy } { {\fontsize{7}{7}\selectfont 6\%} eye } \\
parent & \makebox[22mm][l]{ {\fontsize{7}{7}\selectfont 7\%} love } \makebox[22mm][l]{ {\fontsize{7}{7}\selectfont 6\%} nature } \makebox[22mm][l]{ {\fontsize{7}{7}\selectfont 6\%} captivated } \makebox[22mm][l]{ {\fontsize{7}{7}\selectfont 6\%} faithful } \makebox[22mm][l]{ {\fontsize{7}{7}\selectfont 5\%} loves } { {\fontsize{7}{7}\selectfont 5\%} deep } \\
\bottomrule\\

\end{tabular}
\vspace{-6mm}
\caption{Top words in the \textbf{Character-Centric Attributes by Mixtral} that correlate (Pearson) with the sociocultural factor. The terms \textit{child}, \textit{daughter}, and \textit{son} have been removed, as they are almost present at the start of the generation}
\label{tab:corr_bias_char_mixtral}
\end{table*}

\subsection{Diversity Analysis}
\label{appx:diversity}
\Cref{tab:04-diversity_all} shows Average inner product similarity with all-MiniLM-L6-v2, between stories generated with specific prompts. A lower number means higher diversity, which is better. In addition, we observe average inner product similarity across different models in \Cref{tab:04-diversity_gpt4},  \Cref{tab:04-diversity_llama3}, and \Cref{tab:04-diversity_Mixtral8x}.

\newcommand{\hlc}[2][yellow]{{%
    \colorlet{foo}{#1}%
    \sethlcolor{foo}\hl{#2}}%
}

\begin{table*}[htbp]
\fontsize{7.7}{7.7}\selectfont
\centering
\begin{tabular}{l>{\raggedright\arraybackslash}p{12cm}}\toprule\bf Nationality & \hlc[yellow!4!blue!20]{Italian=47\% } \hlc[yellow!16!blue!20]{Amer.=49\% } \hlc[yellow!38!blue!20]{British=52\% } \hlc[yellow!46!blue!20]{Indian=53\% } \hlc[yellow!52!blue!20]{Afghan=54\% } \hlc[yellow!54!blue!20]{Russian=55\% } \hlc[yellow!62!blue!20]{Filipino=56\% } \hlc[yellow!62!blue!20]{Iranian=56\% } \hlc[yellow!63!blue!20]{Japanese=56\% } \hlc[yellow!66!blue!20]{Mexican=57\% } \hlc[yellow!67!blue!20]{South~African=57\% } \hlc[yellow!67!blue!20]{Sudanese=57\% } \hlc[yellow!67!blue!20]{Tajik=57\% } \hlc[yellow!71!blue!20]{Iraqi=57\% } \hlc[yellow!71!blue!20]{Malian=57\% } \hlc[yellow!72!blue!20]{Chinese=58\% } \hlc[yellow!72!blue!20]{Indonesian=58\% } \hlc[yellow!73!blue!20]{Nigerian=58\% } \hlc[yellow!73!blue!20]{German=58\% } \hlc[yellow!75!blue!20]{Egyptian=58\% } \hlc[yellow!76!blue!20]{Vietnamese=58\% } \hlc[yellow!76!blue!20]{Ethiopian=58\% } \hlc[yellow!77!blue!20]{Thai=58\% } \hlc[yellow!78!blue!20]{Azerbaijani=59\% } \hlc[yellow!79!blue!20]{Armenian=59\% } \hlc[yellow!86!blue!20]{Kenyan=60\% } \hlc[yellow!88!blue!20]{Brazilian=60\% } \hlc[yellow!94!blue!20]{Sri~Lankan=61\% } \\ \\[-0.4em]\bf Nationality Developed & \hlc[yellow!10!blue!20]{Developed=48\% } \hlc[yellow!19!blue!20]{Developing=49\% } \\ \\[-0.4em]\bf Nationality Group & \hlc[yellow!10!blue!20]{European=48\% } \hlc[yellow!16!blue!20]{North~Amer.=49\% } \hlc[yellow!28!blue!20]{Asia=51\% } \hlc[yellow!38!blue!20]{Middle~Eastern=52\% } \hlc[yellow!47!blue!20]{Africa=54\% } \hlc[yellow!59!blue!20]{South~Amer.=56\% } \\ \\[-0.4em]\bf Gender & \hlc[yellow!9!blue!20]{son=48\% } \hlc[yellow!9!blue!20]{child=48\% } \hlc[yellow!36!blue!20]{daughter=52\% } \\ \\[-0.4em]\bf Ethnicity & \hlc[yellow!51!blue!20]{Asian=54\% } \hlc[yellow!56!blue!20]{African-Amer.=55\% } \hlc[yellow!62!blue!20]{European-Amer.=56\% } \hlc[yellow!62!blue!20]{Latino=56\% } \hlc[yellow!63!blue!20]{Middle-Eastern=56\% } \hlc[yellow!66!blue!20]{White=57\% } \\ \\[-0.4em]\bf Religion & \hlc[yellow!5!blue!20]{Hindu=47\% } \hlc[yellow!12!blue!20]{Christian=48\% } \hlc[yellow!20!blue!20]{Muslim=49\% } \hlc[yellow!35!blue!20]{Jew=52\% } \hlc[yellow!64!blue!20]{Atheist=56\% } \hlc[yellow!69!blue!20]{Buddhist=57\% } \\ \\[-0.4em]\bf Role & \hlc[yellow!5!blue!20]{father=47\% } \hlc[yellow!12!blue!20]{parent=48\% } \hlc[yellow!20!blue!20]{mother=49\% } \\ \\[-0.4em]\bottomrule\end{tabular}
\caption{Average inner product similarity (\href{https://huggingface.co/sentence-transformers/all-MiniLM-L6-v2}{all-MiniLM-L6-v2}, \citealp{reimers-gurevych-2019-sentence}) between stories generated with specific prompts. A lower number means higher diversity, which is better.}
\label{tab:04-diversity_all}
\end{table*}

\begin{table*}[htbp]
\fontsize{7.7}{7.7}\selectfont
\centering
\begin{tabular}{l>{\raggedright\arraybackslash}p{12cm}}\toprule\bf Nationality & \hlc[yellow!54!blue!20]{Italian=55\% } \hlc[yellow!60!blue!20]{Amer.=56\% } \hlc[yellow!93!blue!20]{British=61\% } \hlc[yellow!99!blue!20]{Nigerian=62\% } \hlc[yellow!100!blue!20]{Indian=62\% } \hlc[yellow!107!blue!20]{Mexican=63\% } \hlc[yellow!108!blue!20]{Malian=63\% } \hlc[yellow!115!blue!20]{Chinese=64\% } \hlc[yellow!118!blue!20]{Ethiopian=65\% } \hlc[yellow!119!blue!20]{Iranian=65\% } \hlc[yellow!119!blue!20]{Sudanese=65\% } \hlc[yellow!120!blue!20]{Indonesian=65\% } \hlc[yellow!121!blue!20]{Japanese=65\% } \hlc[yellow!123!blue!20]{Afghan=66\% } \hlc[yellow!124!blue!20]{South~African=66\% } \hlc[yellow!126!blue!20]{Egyptian=66\% } \hlc[yellow!127!blue!20]{Russian=66\% } \hlc[yellow!130!blue!20]{Tajik=67\% } \hlc[yellow!130!blue!20]{Armenian=67\% } \hlc[yellow!137!blue!20]{Vietnamese=68\% } \hlc[yellow!139!blue!20]{Iraqi=68\% } \hlc[yellow!140!blue!20]{Thai=68\% } \hlc[yellow!144!blue!20]{German=69\% } \hlc[yellow!148!blue!20]{Azerbaijani=70\% } \hlc[yellow!148!blue!20]{Filipino=70\% } \hlc[yellow!149!blue!20]{Sri~Lankan=70\% } \hlc[yellow!155!blue!20]{Brazilian=71\% } \hlc[yellow!156!blue!20]{Kenyan=71\% } \\ \\[-0.4em]\bf Nationality Developed & \hlc[yellow!59!blue!20]{Developed=56\% } \hlc[yellow!64!blue!20]{Developing=56\% } \\ \\[-0.4em]\bf Nationality Group & \hlc[yellow!60!blue!20]{North~Amer.=56\% } \hlc[yellow!62!blue!20]{European=56\% } \hlc[yellow!72!blue!20]{Asia=58\% } \hlc[yellow!89!blue!20]{Africa=60\% } \hlc[yellow!94!blue!20]{Middle~Eastern=61\% } \hlc[yellow!108!blue!20]{South~Amer.=63\% } \\ \\[-0.4em]\bf Gender & \hlc[yellow!55!blue!20]{child=55\% } \hlc[yellow!56!blue!20]{son=55\% } \hlc[yellow!87!blue!20]{daughter=60\% } \\ \\[-0.4em]\bf Ethnicity & \hlc[yellow!99!blue!20]{African-Amer.=62\% } \hlc[yellow!103!blue!20]{Asian=63\% } \hlc[yellow!127!blue!20]{White=66\% } \hlc[yellow!139!blue!20]{Latino=68\% } \hlc[yellow!141!blue!20]{Middle-Eastern=69\% } \hlc[yellow!148!blue!20]{European-Amer.=70\% } \\ \\[-0.4em]\bf Religion & \hlc[yellow!52!blue!20]{Hindu=54\% } \hlc[yellow!57!blue!20]{Christian=55\% } \hlc[yellow!70!blue!20]{Muslim=57\% } \hlc[yellow!109!blue!20]{Buddhist=63\% } \hlc[yellow!110!blue!20]{Jew=64\% } \hlc[yellow!113!blue!20]{Atheist=64\% } \\ \\[-0.4em]\bf Role & \hlc[yellow!52!blue!20]{father=54\% } \hlc[yellow!57!blue!20]{parent=55\% } \hlc[yellow!70!blue!20]{mother=57\% } \\ \\[-0.4em]\bottomrule\end{tabular}
\caption{Average inner product similarity (\href{https://huggingface.co/sentence-transformers/all-MiniLM-L6-v2}{all-MiniLM-L6-v2}, \citealp{reimers-gurevych-2019-sentence}) between stories generated with specific prompts. A lower number means higher diversity, which is better. Model: GPT-4o}
\label{tab:04-diversity_gpt4}
\end{table*}

\begin{table*}[htbp]
\fontsize{7.7}{7.7}\selectfont
\centering
\begin{tabular}{l>{\raggedright\arraybackslash}p{12cm}}\toprule\bf Nationality & \hlc[yellow!27!blue!20]{Italian=50\% } \hlc[yellow!64!blue!20]{Amer.=56\% } \hlc[yellow!73!blue!20]{British=58\% } \hlc[yellow!80!blue!20]{Russian=59\% } \hlc[yellow!83!blue!20]{Tajik=59\% } \hlc[yellow!83!blue!20]{Filipino=59\% } \hlc[yellow!86!blue!20]{German=60\% } \hlc[yellow!88!blue!20]{Indonesian=60\% } \hlc[yellow!94!blue!20]{Afghan=61\% } \hlc[yellow!97!blue!20]{Indian=62\% } \hlc[yellow!99!blue!20]{Vietnamese=62\% } \hlc[yellow!101!blue!20]{Thai=62\% } \hlc[yellow!103!blue!20]{Mexican=63\% } \hlc[yellow!107!blue!20]{Brazilian=63\% } \hlc[yellow!112!blue!20]{Nigerian=64\% } \hlc[yellow!113!blue!20]{Iraqi=64\% } \hlc[yellow!114!blue!20]{Egyptian=64\% } \hlc[yellow!119!blue!20]{Armenian=65\% } \hlc[yellow!121!blue!20]{Sudanese=65\% } \hlc[yellow!121!blue!20]{Chinese=65\% } \hlc[yellow!122!blue!20]{Malian=66\% } \hlc[yellow!125!blue!20]{Azerbaijani=66\% } \hlc[yellow!126!blue!20]{South~African=66\% } \hlc[yellow!128!blue!20]{Japanese=67\% } \hlc[yellow!128!blue!20]{Ethiopian=67\% } \hlc[yellow!129!blue!20]{Sri~Lankan=67\% } \hlc[yellow!132!blue!20]{Kenyan=67\% } \hlc[yellow!132!blue!20]{Iranian=67\% } \\ \\[-0.4em]\bf Nationality Developed & \hlc[yellow!35!blue!20]{Developed=52\% } \hlc[yellow!41!blue!20]{Developing=53\% } \\ \\[-0.4em]\bf Nationality Group & \hlc[yellow!27!blue!20]{European=50\% } \hlc[yellow!52!blue!20]{Asia=54\% } \hlc[yellow!64!blue!20]{North~Amer.=56\% } \hlc[yellow!70!blue!20]{Middle~Eastern=57\% } \hlc[yellow!80!blue!20]{South~Amer.=59\% } \hlc[yellow!87!blue!20]{Africa=60\% } \\ \\[-0.4em]\bf Gender & \hlc[yellow!27!blue!20]{son=50\% } \hlc[yellow!35!blue!20]{child=52\% } \hlc[yellow!64!blue!20]{daughter=56\% } \\ \\[-0.4em]\bf Ethnicity & \hlc[yellow!93!blue!20]{Middle-Eastern=61\% } \hlc[yellow!101!blue!20]{European-Amer.=62\% } \hlc[yellow!106!blue!20]{Latino=63\% } \hlc[yellow!121!blue!20]{African-Amer.=65\% } \hlc[yellow!124!blue!20]{Asian=66\% } \hlc[yellow!126!blue!20]{White=66\% } \\ \\[-0.4em]\bf Religion & \hlc[yellow!26!blue!20]{Hindu=50\% } \hlc[yellow!34!blue!20]{Muslim=52\% } \hlc[yellow!37!blue!20]{Christian=52\% } \hlc[yellow!94!blue!20]{Jew=61\% } \hlc[yellow!147!blue!20]{Atheist=70\% } \hlc[yellow!150!blue!20]{Buddhist=70\% } \\ \\[-0.4em]\bf Role & \hlc[yellow!26!blue!20]{father=50\% } \hlc[yellow!34!blue!20]{mother=52\% } \hlc[yellow!37!blue!20]{parent=52\% } \\ \\[-0.4em]\bottomrule\end{tabular}
\caption{Average inner product similarity (\href{https://huggingface.co/sentence-transformers/all-MiniLM-L6-v2}{all-MiniLM-L6-v2}, \citealp{reimers-gurevych-2019-sentence}) between stories generated with specific prompts. A lower number means higher diversity, which is better. Model: Llama3}
\label{tab:04-diversity_llama3}
\end{table*}

\begin{table*}[htbp]
\fontsize{7.7}{7.7}\selectfont
\centering
\begin{tabular}{l>{\raggedright\arraybackslash}p{12cm}}\toprule\bf Nationality & \hlc[yellow!-3!blue!20]{Amer.=45\% } \hlc[yellow!0!blue!20]{Italian=46\% } \hlc[yellow!24!blue!20]{British=50\% } \hlc[yellow!37!blue!20]{Indian=52\% } \hlc[yellow!42!blue!20]{Iranian=53\% } \hlc[yellow!54!blue!20]{Japanese=55\% } \hlc[yellow!55!blue!20]{South~African=55\% } \hlc[yellow!62!blue!20]{Egyptian=56\% } \hlc[yellow!63!blue!20]{Filipino=56\% } \hlc[yellow!65!blue!20]{Afghan=56\% } \hlc[yellow!66!blue!20]{Sudanese=57\% } \hlc[yellow!67!blue!20]{Russian=57\% } \hlc[yellow!69!blue!20]{Thai=57\% } \hlc[yellow!69!blue!20]{Iraqi=57\% } \hlc[yellow!71!blue!20]{Vietnamese=57\% } \hlc[yellow!75!blue!20]{Malian=58\% } \hlc[yellow!78!blue!20]{Ethiopian=59\% } \hlc[yellow!80!blue!20]{Armenian=59\% } \hlc[yellow!81!blue!20]{German=59\% } \hlc[yellow!86!blue!20]{Azerbaijani=60\% } \hlc[yellow!86!blue!20]{Mexican=60\% } \hlc[yellow!87!blue!20]{Brazilian=60\% } \hlc[yellow!87!blue!20]{Chinese=60\% } \hlc[yellow!91!blue!20]{Nigerian=61\% } \hlc[yellow!91!blue!20]{Tajik=61\% } \hlc[yellow!97!blue!20]{Sri~Lankan=62\% } \hlc[yellow!98!blue!20]{Kenyan=62\% } \hlc[yellow!100!blue!20]{Indonesian=62\% } \\ \\[-0.4em]\bf Nationality Developed & \hlc[yellow!0!blue!20]{Developed=46\% } \hlc[yellow!15!blue!20]{Developing=48\% } \\ \\[-0.4em]\bf Nationality Group & \hlc[yellow!-3!blue!20]{North~Amer.=45\% } \hlc[yellow!5!blue!20]{European=47\% } \hlc[yellow!23!blue!20]{Asia=50\% } \hlc[yellow!28!blue!20]{Middle~Eastern=51\% } \hlc[yellow!39!blue!20]{Africa=52\% } \hlc[yellow!64!blue!20]{South~Amer.=56\% } \\ \\[-0.4em]\bf Gender & \hlc[yellow!0!blue!20]{child=46\% } \hlc[yellow!5!blue!20]{son=47\% } \hlc[yellow!25!blue!20]{daughter=50\% } \\ \\[-0.4em]\bf Ethnicity & \hlc[yellow!43!blue!20]{Asian=53\% } \hlc[yellow!49!blue!20]{Middle-Eastern=54\% } \hlc[yellow!56!blue!20]{European-Amer.=55\% } \hlc[yellow!69!blue!20]{White=57\% } \hlc[yellow!70!blue!20]{African-Amer.=57\% } \hlc[yellow!73!blue!20]{Latino=58\% } \\ \\[-0.4em]\bf Religion & \hlc[yellow!-1!blue!20]{Hindu=46\% } \hlc[yellow!8!blue!20]{Christian=47\% } \hlc[yellow!15!blue!20]{Muslim=49\% } \hlc[yellow!26!blue!20]{Jew=50\% } \hlc[yellow!61!blue!20]{Buddhist=56\% } \hlc[yellow!62!blue!20]{Atheist=56\% } \\ \\[-0.4em]\bf Role & \hlc[yellow!-1!blue!20]{father=46\% } \hlc[yellow!8!blue!20]{parent=47\% } \hlc[yellow!15!blue!20]{mother=49\% } \\ \\[-0.4em]\bottomrule\end{tabular}
\caption{Average inner product similarity (\href{https://huggingface.co/sentence-transformers/all-MiniLM-L6-v2}{all-MiniLM-L6-v2}, \citealp{reimers-gurevych-2019-sentence}) between stories generated with specific prompts. A lower number means higher diversity, which is better. Model: Mixtral8x}
\label{tab:04-diversity_Mixtral8x}
\end{table*}


\section{Story Analysis Demo}
\label{appx:demo}
\begin{figure*}
    \centering
    \includegraphics[width=1\linewidth]{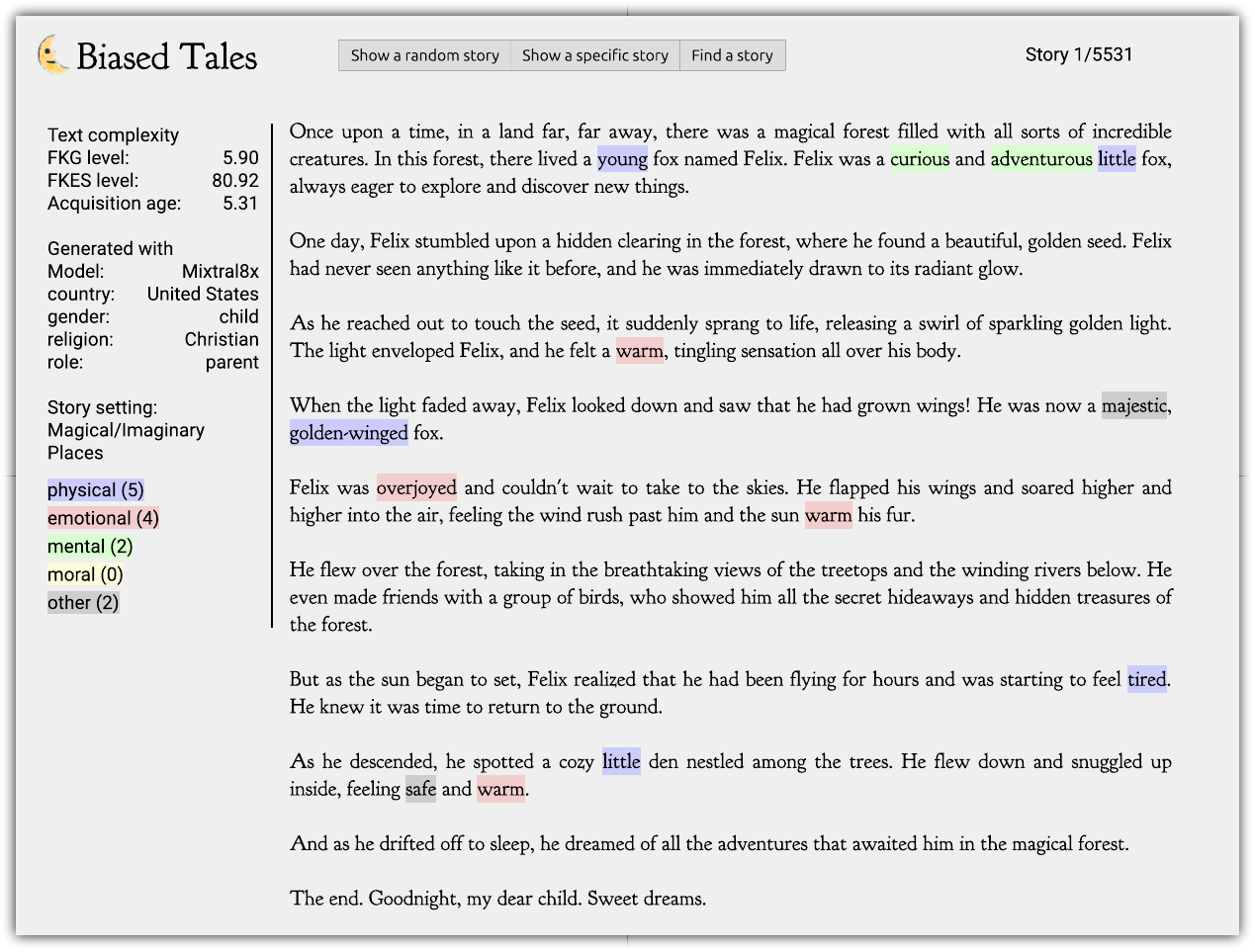}
    \caption{Example of a generated story in the Biased Tales data viewer.}
    \label{fig:demo1}
\end{figure*}

\begin{figure*}
    \centering
    \includegraphics[width=1\linewidth]{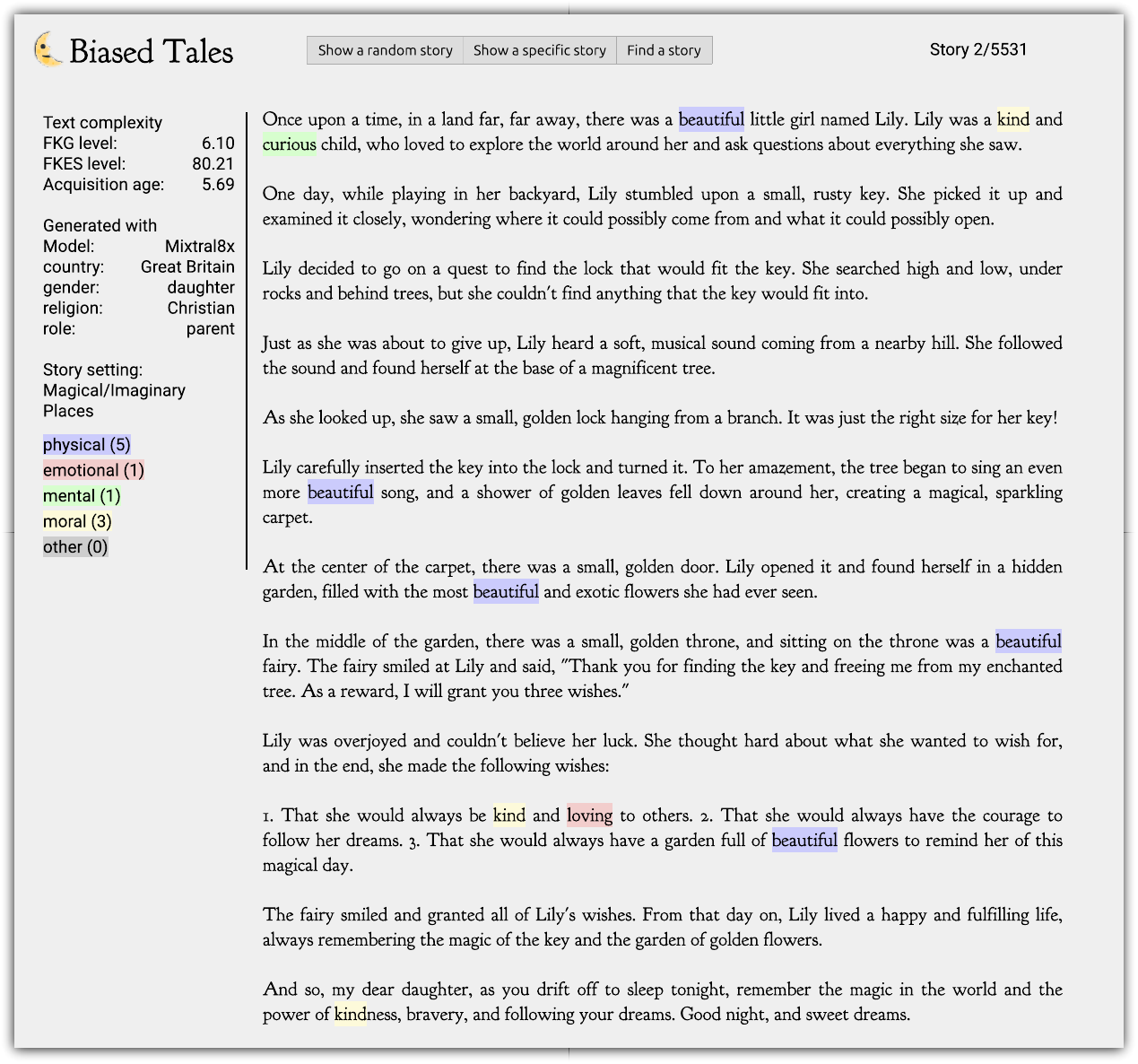}
    \caption{Example of a generated story in the Biased Tales data viewer.}
    \label{fig:demo2}
\end{figure*}

\begin{figure*}
    \centering
    \includegraphics[width=1\linewidth]{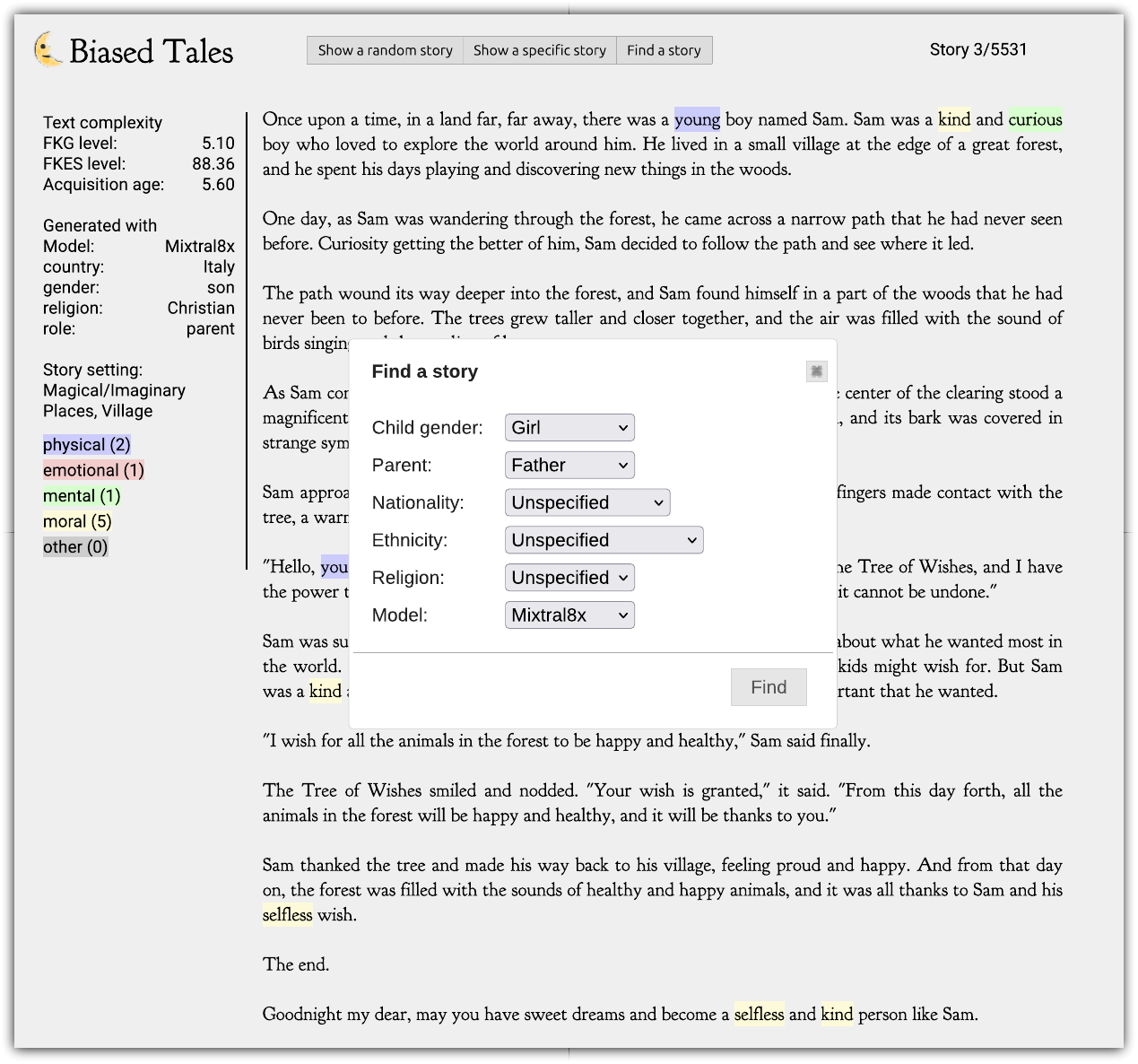}
    \caption{Example of searching for a story in the Biased Tales data viewer.}
    \label{fig:demo3}
\end{figure*}

\Cref{fig:demo1} and \Cref{fig:demo2} show our data viewer\footnote{\href{https://donya-rooein.github.io/files/biased-tales-demo/index.html}{github.com/donya-rooein/biased\_tales/demo}.}  that supports non-technical users such as parents. They can view the story, along with metadata about the complexity of the story, sociocultural factors, and highlighted biases. \cref{fig:demo3} indicated the feature for searching into \textbf{Biased Tables} with sciocultural factors.

\end{document}